\documentclass{ieeetj}
\usepackage{cite}
\usepackage{amsmath,amssymb,amsfonts}
\usepackage{graphicx,color}
\usepackage{textcomp} 
\usepackage{xcolor}
\usepackage{hyperref}
\hypersetup{hidelinks=true}

\usepackage{algorithm,algorithmicx}
\def\BibTeX{{\rm B\kern-.05em{\sc i\kern-.025em b}\kern-.08em
    T\kern-.1667em\lower.7ex\hbox{E}\kern-.125emX}}

\newcommand{\fig}[1]{Fig.~\ref{#1}}

\def\epsgaiji#1{\leavevmode\kern-0.025zw\raise-.37zh\hbox{%
  \epsfile{file=#1,width=1.05zw}}\kern-0.025zw}
\newcommand{\MARU}[1]{{\ooalign{\hfil#1\/\hfil\crcr\raise.167ex\hbox{\mathhexbox20D}}}}

\usepackage{array}

\newcommand{\subcaption}[2]{\small (#1)~#2}

\newcommand{\MS}{Motion~Stack\xspace}
\newcommand{\MB}{MoonBot\xspace}
\newcommand{\MBS}{\MB{} \MS{}\xspace}

\newcommand{\MBM}{\MB~Minimal\xspace}

\newcommand{\MBD}{\MB~Dragon\xspace}
\newcommand{\MBT}{\MB~Tricycle\xspace}

\newcommand{\moit}{Moveit!\xspace}
\newcommand{\ro}{ROS1\xspace}
\newcommand{\rt}{ROS2\xspace}

\newcommand{\PCID}{\texttt{PC\_ID}\xspace}
\newcommand{\PY}{Python\xspace}

\newcommand{\mocap}{Motion~Capture\xspace}

\usepackage{flushend} % to level the end of the page
\usepackage{siunitx} % to use \SI command to visualize the unit
\usepackage{xspace} % for smart spaces

\AtBeginDocument{\definecolor{tmlcncolor}{cmyk}{0.93,0.59,0.15,0.02}\definecolor{NavyBlue}{RGB}{0,86,125}}

\def\OJlogo{\vspace{-4pt}$<$Society logo(s) and publication title will appear here.$>$}
\def\seclogo{\vspace{10pt}$<$Society logo(s) and publication title will appear here.$>$}

\def\authorrefmark#1{\ensuremath{^{\textbf{#1}}}}

\begin{document}
\receiveddate{XX Month, XXXX}
\reviseddate{XX Month, XXXX}
\accepteddate{XX Month, XXXX}
\publisheddate{XX Month, XXXX}
\currentdate{XX Month, XXXX}
\doiinfo{XXXX.2022.1234567}

\markboth{}{Author {et al.}}

\title{MoonBot: Modular and On-demand Reconfigurable Robot \\Toward Moon Base Construction}

%%%%%%%%%%%%%%%%%%%%%%%%%%%%%%%%%%%%%%%%%%%%%%%%%%%%%%%%%%%%%%%%%%%%%%%%%%%%%%%%%%%%%
%%% Due to Double-Anonymous Review Process, we will Not show the authors/affiliation/funding acknowledgement until the paper accepted.
%%%%%%%%%%%%%%%%%%%%%%%%%%%%%%%%%%%%%%%%%%%%%%%%%%%%%%%%%%%%%%%%%%%%%%%%%%%%%%%%%%%%%
\author{Kentaro Uno\authorrefmark{1*}, 
Elian Neppel\authorrefmark{1*}, 
Gustavo H. Diaz\authorrefmark{1*}, 
Ashutosh Mishra\authorrefmark{1*},
Shamistan Karimov\authorrefmark{1*}, 
A. Sejal Jain\authorrefmark{1}, 
Ayesha Habib\authorrefmark{1}, 
Pascal Pama\authorrefmark{1}, 
Hazal Gozbasi\authorrefmark{1},
Shreya Santra\authorrefmark{1}, 
Kazuya Yoshida\authorrefmark{1}}
\affil{Space Robotics Laboratory (SRL), Department of Aerospace Engineering, Graduate School of Engineering, \\Tohoku University, Sendai 980-8579, Japan.}
\corresp{*These authors contributed equally. Corresponding author: Kentaro Uno (email: unoken@tohoku.ac.jp).}
\authornote{This work was supported by Japan Science and Technology Agency's Moonshot R\&D Program, Grant Number JPMJMS223B.}

\begin{abstract}
The allure of lunar surface exploration and development has recently captured widespread global attention. Robots have proved to be indispensable for exploring uncharted terrains, uncovering and leveraging local resources, and facilitating the construction of future human habitats. In this paper, we introduce MoonBot, a modular and reconfigurable robotic system engineered to maximize functionality while operating within the stringent mass constraints of lunar payloads and adapting to varying environmental conditions and task requirements.
This paper details the design and development of MoonBot and presents a preliminary field demonstration that validates the proof of concept through the execution of milestone tasks simulating the establishment of lunar infrastructure. These tasks include essential civil engineering operations, the transportation and deployment of infrastructural components, and assistive operations with inflatable modules. Furthermore, we systematically summarize the lessons learned during testing, focusing on the connector design providing valuable insights for the advancement of modular robotic systems in future lunar missions. 
\end{abstract}

\begin{IEEEkeywords}
% Enter key words or phrases in alphabetical order, separated by commas. Using the \textit{IEEE Thesaurus} can help you find the best standardized keywords to fit your article. Use the \underline{\href{https://www.ieee.org/publications/services/thesaurus.html}{thesaurus access request form}} for free access to the \textit{IEEE Thesaurus}.
Modular robot, Moon base construction, Reconfigurable robot, Robotic assembly.
% Space exploration, Space vehicles, Telerobotics
\end{IEEEkeywords}


\maketitle

\section{INTRODUCTION}
% --------------------------------------------------------------------------
\begin{figure*}[t]
\centerline{\includegraphics[width=\linewidth]{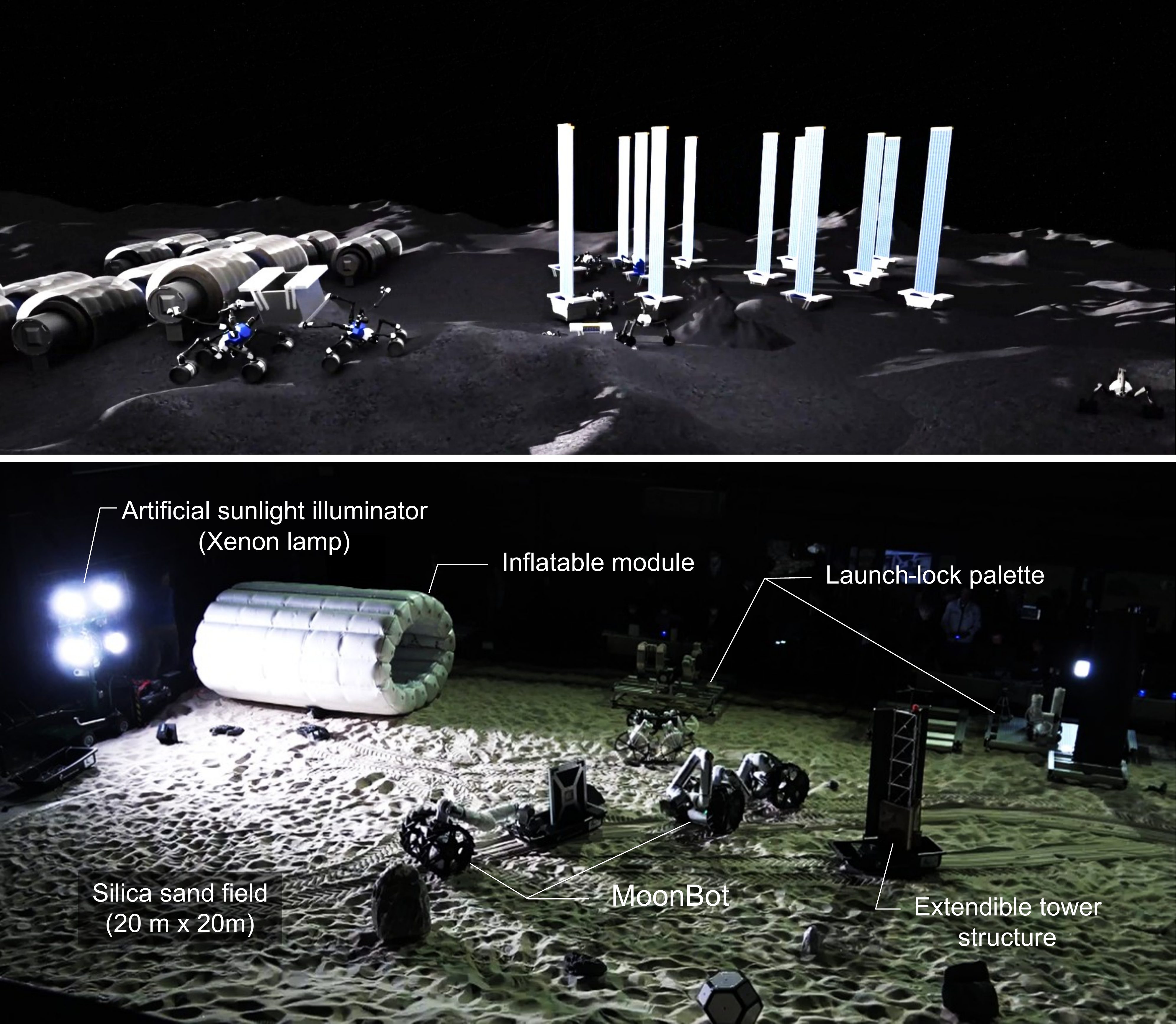}}
\caption{Conceptual illustration of the robotic construction of the base for long-term exploration and resource utilization around the pole on the Moon (top) and MoonBot's first field demonstration in a large sand field (bottom).
In the initial phase of human outpost construction, robots will be solely responsible for autonomously establishing essential infrastructure, including energy stations, shelters, and communication systems. A modular robotic system provides an efficient solution, optimizing performance while adhering to strict mass constraints during launch.
In an unexplored lunar environment, modular robots will dynamically reconfigure their morphology, optimizing locomotion and manipulation capabilities to adapt to unforeseen challenges and execute diverse range of tasks.
At the lunar poles, solar power modules are designed to stand upright like pillars, allowing them to efficiently capture sunlight from low-angle horizontal directions. Cylindrical modules, which will serve as robot garages and human habitable spaces, feature an inflatable, modular structure. These modules, along with the robots, will be transported to the lunar surface, where the robots will assist in their assembly, ensuring a scalable and efficient construction process.
\label{moonBaseConcept}}
\end{figure*}
% These modules are transported to the moon mounted on launch palletes and will commence the self-assembly of mobile robots upon arrival.
% --------------------------------------------------------------------------
\IEEEPARstart{S}{ince} the Apollo missions over half a century ago sparked humanity’s quest for lunar exploration, there has been a resurgence of interest in returning to the Moon, driven by significant advancements in robotics, precise landing technologies, and the potential for resource utilization. 
In recent years, a number of successful robotic missions have been launched to the Moon, including Indian Space Research Organization (ISRO)'s Pragyan rover exploring the lunar south polar region~\cite{Chandrayaan-3}, Japan Aerospace Exploration Agency (JAXA)'s SLIM lander~\cite{slim} demonstrating precision landing, and China National Space Administration's (CNSA) Chang’e6 sample return from the far-side of the Moon~\cite{chang'e6}. These missions have underscored the Moon's critical role in future space exploration and sustainable human presence beyond Earth. Establishing a lunar base offers significant scientific and economic benefits, from facilitating low-gravity manufacturing of innovative pharmaceuticals~\cite{Yamada2020,Braddock2020Drug} and materials~\cite{warzecha2020olanzapine}, to enhancing our understanding of lunar science and providing a launchpad for deep-space exploration missions~\cite{JOHNSON2003727}.

The Artemis program, led by the National Aeronautics and Space Administration (NASA), is a comprehensive initiative focused on establishing a long-term human presence on the Moon as a precursor to Mars exploration~\cite{artemis}. One of the core objectives is the development of lunar infrastructure, with a strong emphasis on leveraging robotic technologies for construction, assembly, and resource utilization. The southern polar region of the Moon, with its permanently shadowed areas rich in resources, is considered the most promising location for establishing such a base. The resources of this region are the key to human settlement by using in-situ resources, including water-ice for life support and propellant production~\cite{FLAHAUT2020104750, Paige2010}. 

However, unlike the equatorial regions previously explored, 
the polar regions present additional challenging conditions, including significant variations in terrain elevation, extreme temperature fluctuations between day and night, and areas with limited natural illumination, as the Sun remains just above the horizon for extended periods~\cite{GLASER2018170,AGU2008}. These harsh environmental factors necessitate the development of highly adaptable, efficient, and autonomous robotic systems for effective operation. Such robotic platforms must be capable of navigating challenging terrain, handling unpredictable situations, and performing a diverse range of manipulation tasks under these extreme conditions. 

Space exploration is severely limited by the immense fuel requirements needed to transport materials into the lunar orbit, with costs often exceeding millions of dollars per kilogram~\cite{lunar_payload}. In current Moon and Mars exploration missions, a single robot is typically developed and deployed to meet specific mission requirements. This approach prioritizes mission-specific reliability rather than versatility and reusability in robot design. However, once the mission is completed, the robot is usually not reusable for other applications or future missions. To enhance mission efficiency and ensure the sustainability of long-term space exploration, significant technological advancements are urgently needed. % As a rule, the scale and complexity of the robot depends on the number of expected tasks.
One promising solution lies in the use of modular robotic systems, which offer a clear advantage in this context. By employing a core set of interchangeable modules, these systems can be customized and reconfigured to meet specific mission demands, providing flexibility and scalability~\cite{reconfigurable_modular_robots_survey}. Moreover, modular robots enhance serviceability such that if a module fails, it can simply be replaced, minimizing downtime and maintenance costs~\cite{paik_modrob, Whitman-2020-126627}. 
Establishing foundational infrastructure in extraterrestrial environments---such as deployable components like solar power units, communication systems, and inflatable habitats---can be effectively achieved through the deployment and maintenance of autonomous modular robots~\cite{Moon_base}. These robots offer the flexibility and adaptability essential for efficient setup and long-term operation. The modular design allows for rapid reconfiguration, improves robustness against failures, and addresses the limitations of mass for space transportation~\cite{modular_robot_extraterrestrial, polybot2}. Instead of designing a new robot for each task, a preconfigured set of modules can be deployed on the lunar surface. The vast array of possible module combinations allows flexible design, enabling the creation of modular robots capable of handling a wide range of tasks on demand~\cite{POST2021530}.

% --------------------------------------------------------------------------
\begin{figure*}[t]
    % \begin{minipage}[t]{.55\linewidth}
    %   \centering 
    %   \centerline{\includegraphics[height=50mm]{./fig/launch_lock_palette.png}}
    %   (a)
    % \end{minipage}
    % % \hspace{.03\linewidth}
    % \begin{minipage}[t]{.45\linewidth} 
    %   \centering 
    %   \centerline{\includegraphics[height=50mm]{./fig/modular_robots_possible_modes.png}}
    %   (b)
    % \end{minipage}
    \centerline{\includegraphics[width=\linewidth]{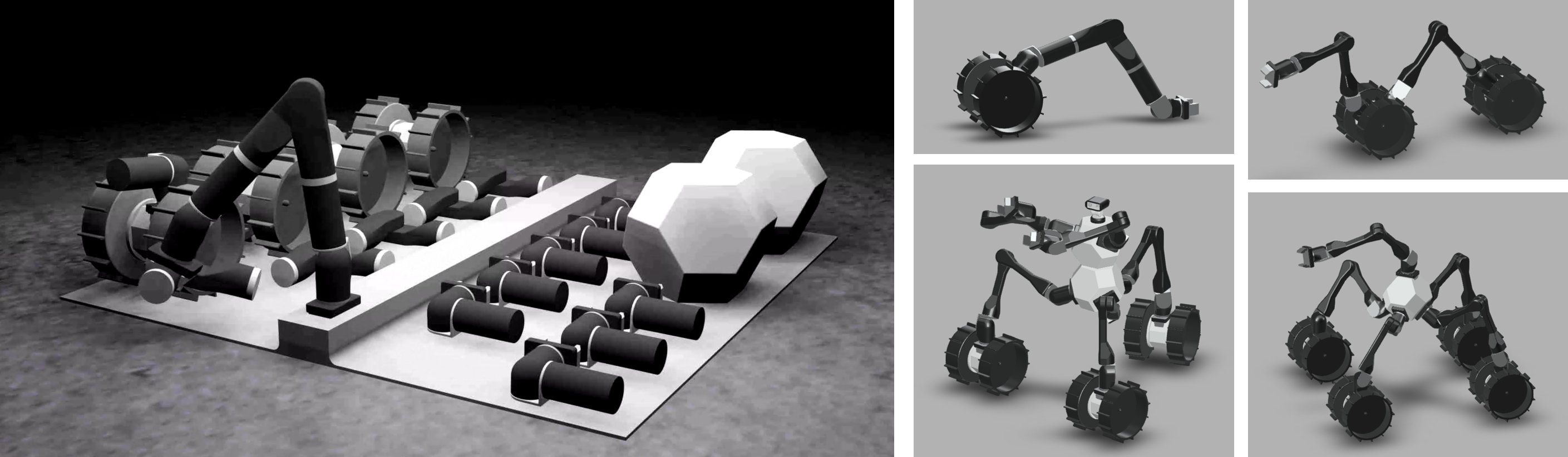}}
\caption{Self-assembly of the modular robots concept on the Moon. The robots are transported to the lunar surface as compact modules affixed to the specialized plate called a launch-lock palette, which is housed in the payload space of the landing vehicle (left). Each module is categorized based on its functionality: Bases, Limbs, Wheels, and tool-changing End Effectors. Upon arrival, the robot modules can autonomously reconfigure themselves into various morphologies, adapting to different tasks and mobility requirements (right). \label{MoonBot_concept}}
\end{figure*}
% These modules are transported to the moon mounted on launch palletes and will commence the self-assembly of mobile robots upon arrival.
% --------------------------------------------------------------------------
\subsection{RELATED WORKS}
Modularity in engineering refers to the division of systems into smaller, independent components, which streamline processes such as manufacturing, assembly, maintenance, and repair~\cite{modularRobotReview2007}. In the late 1980s, Fukuda et al. introduced the ``Dynamically Reconfigurable Robotic System (DRRS),'' which could reconfigure its structure based on a given task and situation~\cite{DDRS_Fukuda}. This approach enables the construction of complex systems from modular units, enhancing flexibility, allowing reconfiguration, and improving adaptability while optimizing design and operational efficiency~\cite{Yim2009, paik_modrob}. Modularity accelerates the design process, enabling rapid prototyping of near-final systems, allowing for quicker adjustments and iterative improvements to the robotic system as mission needs evolve~\cite{Whitman-2020-126627, SMORES2}. {It is also important to mention that potential space applications, such as building and servicing large-scale structures in orbit, have been a recurring topic of research in modular robotics for several years~\cite{victor2007self, Goeller2012}.}

There are two primary design philosophies for robotic modules: homogeneous and heterogeneous modular systems.
Homogeneous modules consist of identical, cellular units, typically featuring one or two degrees of freedom. These modules are designed to connect and disconnect from one another, allowing the system to reconfigure itself based on the task requirement. Building on DRRS, Fukuda et al. developed the Cellular Robotic System (CEBOT)~\cite{CEBOT_FUKUDA}, leading to the design of various reconfigurable modular robot prototypes based on the CEBOT concept as seen in PolyBots~\cite{polybot}, SMORES~\cite{SMORES1}, and Roombots~\cite{roombots}. This self-reconfigurability enables the robot to morph into various configurations. It can transform into a manipulator to move objects, modify its form for efficient locomotion, or assemble into infrastructure, demonstrating the versatility of modular design~\cite{Superbot, atron, M-TRAN, armadas}. Research on homogeneous modular robotics has advanced through swarm robotics, with Kilobot being a prominent example, demonstrating large-scale coordination with a thousand units, where each module was designed as a low-cost miniaturized functionality, demonstrated by Rubenstein et al.~\cite{kilobot, kilobotSelfassembly}. 
% : two-dimensional mobility and communication. 
Conventionally, the advantage of a homogeneous modular system has been confirmed by scoping on the structure construction by themselves. For this, the module is typically designed in repeatable shapes (e.g. cubic or polyhedral), which allows the formation of lattice-based structures by connecting to each other~\cite{Terada_Murata,NASAAMESModularStructure,termite-inspired,MIT_asm_robot,toriiICM}. 

On the other hand, heterogeneous modular robots have been actively studied since the 2010s. This type of modular system employs interchangeable modules that are not necessarily identical; each module is designed with a specific function or task in mind, allowing for a more specialized approach to problem-solving. These systems enhance efficiency in executing complex tasks by utilizing modules tailored for specific functions, as seen in robots like Snapbot~\cite{snapbot, snapbotv2}, Eigenbot~\cite{Eigenbot} and SABER ~\cite{Romanov2, Romanov1}. The combination of different specialized modules enables greater adaptability in dynamic environments, making it possible to create robots tailored to specific missions or challenges~\cite{BACA2012522, control_multi_leg_appli, GAN2022, Matsuno_modular}. 

%added about modular robot control software 
Modular robots offer significant advantages over traditional designs, but they also introduce several challenges. The complexity of a modular robot is influenced by its degree of decentralization. Fully independent mobile modules require each module to be a self-contained system, equipped with its own power, computing, sensors, motors, and communication~\cite{MSR_Vu}. To achieve modularity and fulfill the self-contained requirements of a reconfigurable robot, it is crucial to distribute control tasks across each module---including torque/force control, position control, friction compensation, and limit detection~\cite{AHMADZADEH201527}. While this decentralized approach provides maximum redundancy and reconfigurability, it also increases the computational demands~\cite{heuss02419268}. A reconfigurable robot's motion is primarily controlled by its kinematic model, even for dynamic tasks. In~\cite{modular_control}, Xuemei et al., proposed a hybrid control architecture that integrates both centralized and distributed control mechanisms. In this framework, a centralized supervisory controller handles high-level tasks, including global path planning, decision-making, and overall mission coordination. Simultaneously, distributed module controllers are responsible for lower-level operations such as torque and current regulation, real-time position limit response, and processing of sensor feedback signals. This architecture enhances system adaptability, ensuring efficient coordination between modules while maintaining robustness in dynamic and uncertain environments. MiniQuad developed in~\cite{control_multi_leg_appli} implements a distributed hierarchical control system with the flexibility to quickly replace the hardware modules. In~\cite{Terada_Murata}, Terada and Murata presented homogeneous robotic modules that are capable of fully automated construction under a distributed control system. In~\cite{modular_robot_extraterrestrial}, Cordie et al. proposed a framework to identify failures and efficient methods to repair wheel-legged modular robots. Such robots can adapt and fix themselves, making them highly effective for operations in unpredictable and challenging terrains, critical for assembly and construction in extreme environments. In the case of modular legged robots, it is essential to understand and analyze the gait characteristics. In~\cite{Ning_2019}, Ning et al., developed comprehensive kinematic and dynamic models by analyzing the gait characteristics of robots equipped with various modular leg configurations. Through this analysis, they established the relationships between leg motion parameters and the corresponding driving torques required for locomotion. These derivations serve as a fundamental basis for motion control strategies, enabling precise actuation and adaptive movement. By understanding these relationships, the control system can optimize torque distribution, improve energy efficiency, and enhance the robot’s stability across different terrains and operational conditions. In~\cite{Eigenbot}, Whitman et al., developed methods for the automatic generation of gaits in modular legged robots, utilizing trajectory optimization to design quasi-static gaits. These techniques take into account the robot’s geometric configuration and foot contact patterns to determine efficient and stable locomotion strategies. These approaches were tested on the multi-limbed Eigenbot, enabling dynamic limb reassignment for locomotion, manipulation, and inspection, along with real-time gait adaptation to hardware failures. In~\cite{RCS_ALBUS200587}, Albus and Barbera introduced a 4-level Real-time Control System (RCS), a cognitive architecture with high-level nodes handling strategic decisions like route planning and mission prioritization, while low-level nodes manage tactical operations like obstacle avoidance. This delegation reduces the computational load on individual units, enabling them to focus on specific tasks. Therefore, a modular robot software framework should be such that it can handle a wide range of configurations and skills as per the task requirements reliably and safely. In~\cite{Model_based2010}, Arney et al. introduce a framework for model-based programming of modular robot control, with code generation and verification. It incorporates a programming model based on extended finite state machines, supporting compositionality across both hardware and software levels, along with formal verification to detect issues prior to deployment. In~\cite{Claudio}, Amatucci et al. enhanced Model Predictive Control (MPC) for legged robots using distributed optimization. By decomposing dynamics into modular subsystems and parallel computation, it reduces computational time significantly and can be scaled to the individual modules. These studies have significantly advanced the development of modular robot software, aligning it with evolving hardware requirements.

{For space missions, the constraints on launchable mass demands considerable trade-off between the number of modules and the extent of functionality per module. While dividing the system into smaller modules increases the flexibility and adaptability of modular robots, but also raises the number of inter-module connections and the complexity of necessary operations when reconfiguring, thereby introducing additional potential points of failure.
In our project, we adopt a heterogeneous, functionality-based modular design to balance sufficient modularity with operational simplicity during reconfiguration. This approach reduces both the risk of failure and the time required in missions where the robot’s operable period is severely limited on the Moon, similar to the methods described in ~\cite{BACA2012522,GAN2022}.}
%In our project, to balance sufficient modularity with operational simplicity during reconfiguration---and thereby reduce not only the risk of failure but also the time required in missions where the robot’s operable period is severely limited on the Moon---we adopt a heterogeneous, functionality-based modular design, similar to the approaches as described in~\cite{BACA2012522,GAN2022}.}

Our work in this paper highlights the potential of modular robotic systems for space applications, with a focus on lunar exploration and human base development.

\subsection{MISSION SCENARIO}
% --------------------------------------------------------------------------
\begin{figure*}[t]
\centerline{\includegraphics[width=\linewidth]{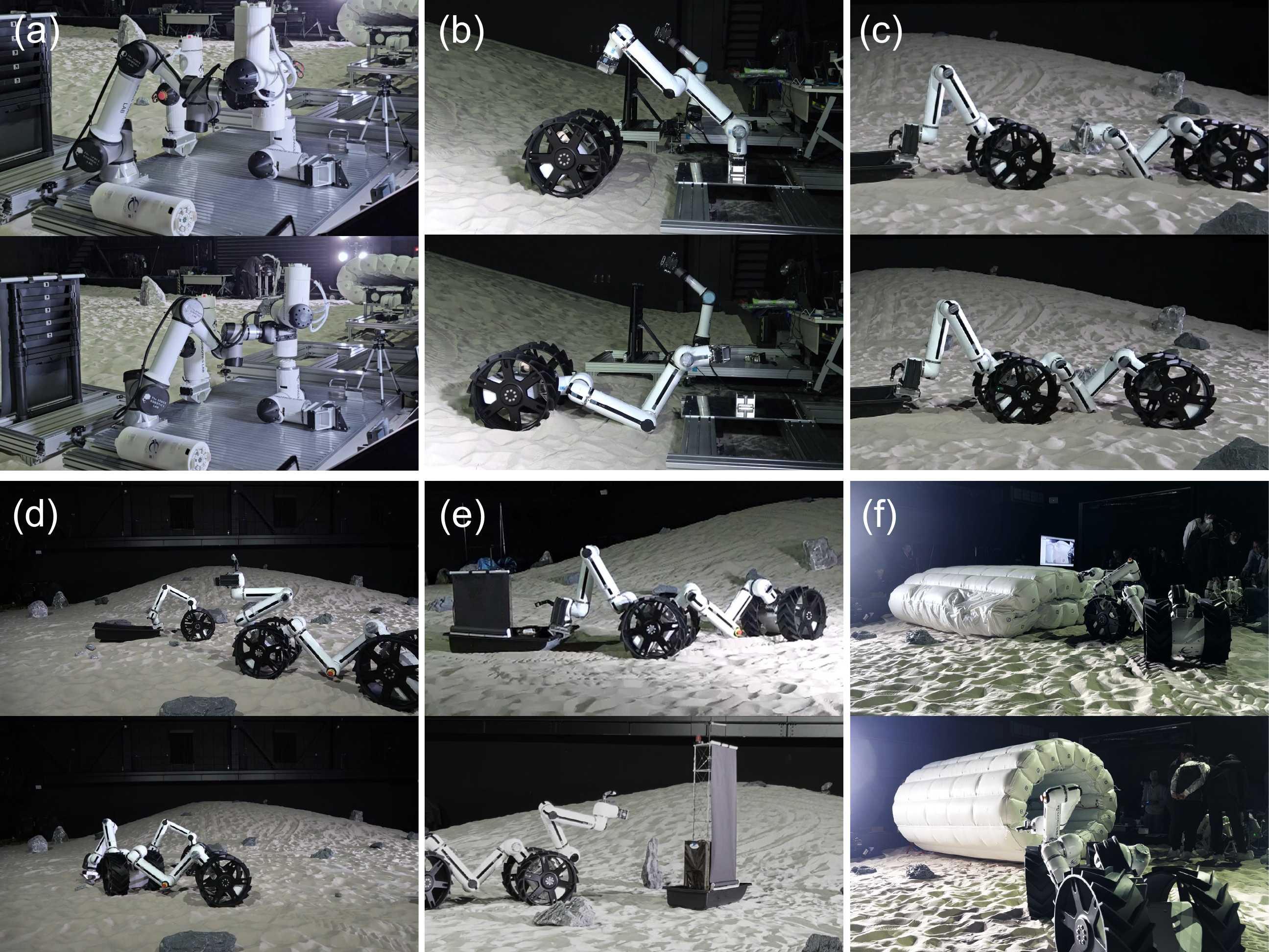}}
\caption{Preliminary demonstrations with MoonBot (Modular \& On-demand reconfigurable roBot) in the lunar analogue test field describing the lunar base construction mission scenario. (a) The initial production of the modular robot is executed on the launch-lock palette, a platform that is directly deployed onto the lunar surface from the landing vehicle. (b) Once the assembly of the articulated Limb module is completed, it grasps the rotational Wheel module, forming a 1 $\times$ Limb + 1 $\times$ Wheel configuration, referred to as the \textit{Minimal} configuration. This represents the minimum functional unit with sufficient mobility and manipulation capability. (c) MoonBot's modularity allows the system to split into smaller, minimal robots or combine into larger, more complex configurations based on the required traction and dexterity. For instance, two Minimal units can be connected in series to form a \textit{Dragon} configuration, enhancing overall capability. (d) A team consisting of a Minimal and a Dragon configuration clears rocks and rakes the field as a civil engineering preparation to ensure the stable placement of infrastructural equipment. (e) A Dragon configuration can tow a deployable solar tower structure weighing over 30\;kg by utilizing its rear Limb as a fixed manipulator, securely gripping a sledge for transport across the sandy terrain. (f) MoonBot assists in the deployment of inflatable modules by monitoring their inflation status and inserting rolling-proof stoppers to prevent unintended movement after full inflation. \label{mission_scenario_scene_by_scene}}
\end{figure*}
% These modules are transported to the moon mounted on launch palletes and will commence the self-assembly of mobile robots upon arrival.
% --------------------------------------------------------------------------
\label{Mission_Scenario} 
Our project aims to establish a permanent lunar base for human activities by the 2050s. This long-term outpost is envisioned to pioneer a new era of human creativity and exploration since the Apollo program. Towards this goal, we primarily focus on the fully robotic establishment of critical infrastructures---such as solar power generators, {pressurized and temperature-stable shelters for future human habitats}, and local/global communication bases---on the Moon during the 2040s, as conceptualized in \fig{moonBaseConcept}. Core technologies are being validated through ground-based testing and will be leveraged to develop preliminary flight models of the robots and associated components, with the initial deployment of selected modules anticipated in the 2030s. 
% effectively utilize the limited amount of components deployed on the Moon, enabling modules to be reconfigurable according to the lunar terrain conditions and mission tasks. The developed technologies can also be applicable on the Earth, such as disaster response and infrastructure settlement in harsh and unstructured environment.

To efficiently construct the lunar base, launching components as modules and assembling them in-situ, similar to previous large-scale spacecraft constructions like the International Space Station (ISS)~\cite{ISS}, is a practical approach. Our proposal involves the development of a group of diverse robots with modular designs. {The combination of such heterogeneous modules provides a balanced trade-off between adaptability and simplicity, enabling rapid responses to situational demands and task complexity. This heterogeneous modular robotic system is particularly well-suited for lunar missions, where frequent delivery of new hardware is impractical and the operational time of robots is severely constrained.}

{We have developed our robotic system, named {MoonBot} (Modular \& On-demand reconfigurable roBot), which is composed of multiple different types of functionally minimal modular segments, drawing inspiration from prior promising demonstrations in modular robotics studies~\cite{snapbot,BACA2012522,GAN2022}.} Our design includes the following core modules: {Limb: An articulated module that can function either a leg or an arm; Hand: An end-effector module designed for gripping and manipulation tasks, intended for use in conjunction with the Limb module; Wheel: A rotational module designed for speedy locomotion; and Body: A central base module equipped with a high-capacity battery and a high-performance on-board computer for extended and advanced operations}. This self-reconfigurable modular robot system is capable of varying its structure and degree of freedom to adapt to diverse tasks, such as carrying, assembling, or manipulating components, as well as to varying situational complexities. Possible configuration options for the robotic system are illustrated in \fig{MoonBot_concept}. Given that continuous improvement and adaptability are central to the hardware concept, the associated control solutions must evolve in parallel. As MoonBot encounters unseen environments and challenges, its controllers must adapt in sync with changing hardware configurations to meet up-to-date task requirements.

The representative phases of the proposed mission scenario are showcased scene by scene with the real MoonBot hardware in \fig{mission_scenario_scene_by_scene}. First, the robotic modules are self-assembled on a launch-lock palette, which is deployed directly onto the Moon's surface. The self-organization process begins with a single limb module, which functions as a base-mounted robot arm to sequentially connect the articulated modules on the palette. 
% (\fig{mission_scenario_scene_by_scene}(a)). 
After initially assembling several articulated modules (i.e. limbs), the system utilizes one limb module to initiate the formation of various robotic configurations that achieve minimal mobility and manipulation capabilities, exemplified by a configuration consisting of 1 $\times$ Limb + 1 $\times$ Wheel. 
% (\fig{mission_scenario_scene_by_scene}(b)). 
Depending on the task and the surrounding terrain to travel on, the robots can be mutually connected to form more complex morphologies. As a representative example, two minimal configurations can be serially connected to create a dragon-like configuration, comprising 2 $\times$ Limbs + 2 $\times$ Wheels. This configuration achieves improved traction because the combined modules increase the ground reactive forces, a critical factor under lunar gravity (1/6~G). 
% Dragon's last limb serves as a mobile manipulator for more stable dexterity. 
% (\fig{mission_scenario_scene_by_scene}(c)). 
An inverse operation is also available, allowing the system to separate into two independent modules for efficient task parallelization. This reconfiguration process comprising both coupling and separation is managed by the higher-level controller of the MoonBot software, including the overall mission sequential planner and task allocation algorithm. 
Towards the lunar outpost establishment, we have identified the following three milestone tasks as particularly essential.

\begin{itemize}
    
    \item {\textbf{Fundamental civil engineering:}} \\
    Basic terrain preparation is essential for improving robotic mobility and ensuring the stable placement of large-scale structures. This process involves the removal of rocks and the leveling of sandy surfaces.
    Collaborative efforts by multiple robotic units can expedite these tasks; for instance, a Minimal robot can be used to transport rock gravel that has been collected by a Dragon robot. \\
    % (\fig{mission_scenario_scene_by_scene}(d)).\\
    
    \item {\textbf{Infrastructural hardware transport and deployment:}} \\
    By connecting additional wheel modules in configuration such as Dragon configuration, the robot shows a greater capability for transporting heavier structure. Using these efficient configurations, we plan to deploy extendible tower structures that primarily serve two functions
    % (\fig{mission_scenario_scene_by_scene}(e))
    :
    \begin{itemize}
        \item {Photovoltaic station:} \\
        The proposed location for the lunar base is in the Moon's polar region, where sunlight strikes horizontally. Unlike terrestrial solar panels, which are typically installed horizontally against the ground, solar panels in this environment must be positioned vertically on the side of tower structure to maximize solar irradiance.
        \item {Local communication base:} \\
        While Earth--Moon communication will rely on the lander's communication system, a secure and stable local network is essential for coordinating a modular robotic team composed of multiple units. For this, the tower will incorporate a communication mast with pole antennas mounted at its apex.
    \end{itemize}
    % \begin{itemize}
    %     \item Connecting additional wheel modules, such as the Dragon configuration, which enhances the robot's capability to carry heavier structures.
    %     \item Using these efficient configurations, we plan to transport extendible tower structures from the lander. These tower structures primarily serve two functions:
    % \end{itemize}

    \item {\textbf{Robotic assistance to deploy the pressurized module:}} \\
    Deploying an inflatable structure as a primary shelter is fundamental to supporting human activity in the lunar base. This structure is designed to protect humans from the severe temperature fluctuations characteristic 
    % (exceeding 200\;$^\circ $C) 
    on the Moon. 
    % (\fig{mission_scenario_scene_by_scene}(f)). 
    
\end{itemize}
% The ground-based demonstration will include multiple robots cooperatively transporting large objects over rough terrain with obstacles, and then deploying and assembling the structures. 
% --------------------------------------------------------------------------
\begin{figure*}[t]
    % \centerline{\includegraphics[width=\linewidth]{./fig/moonbot_hardware_overview.pdf}}
    \centerline{\includegraphics[width=\linewidth]{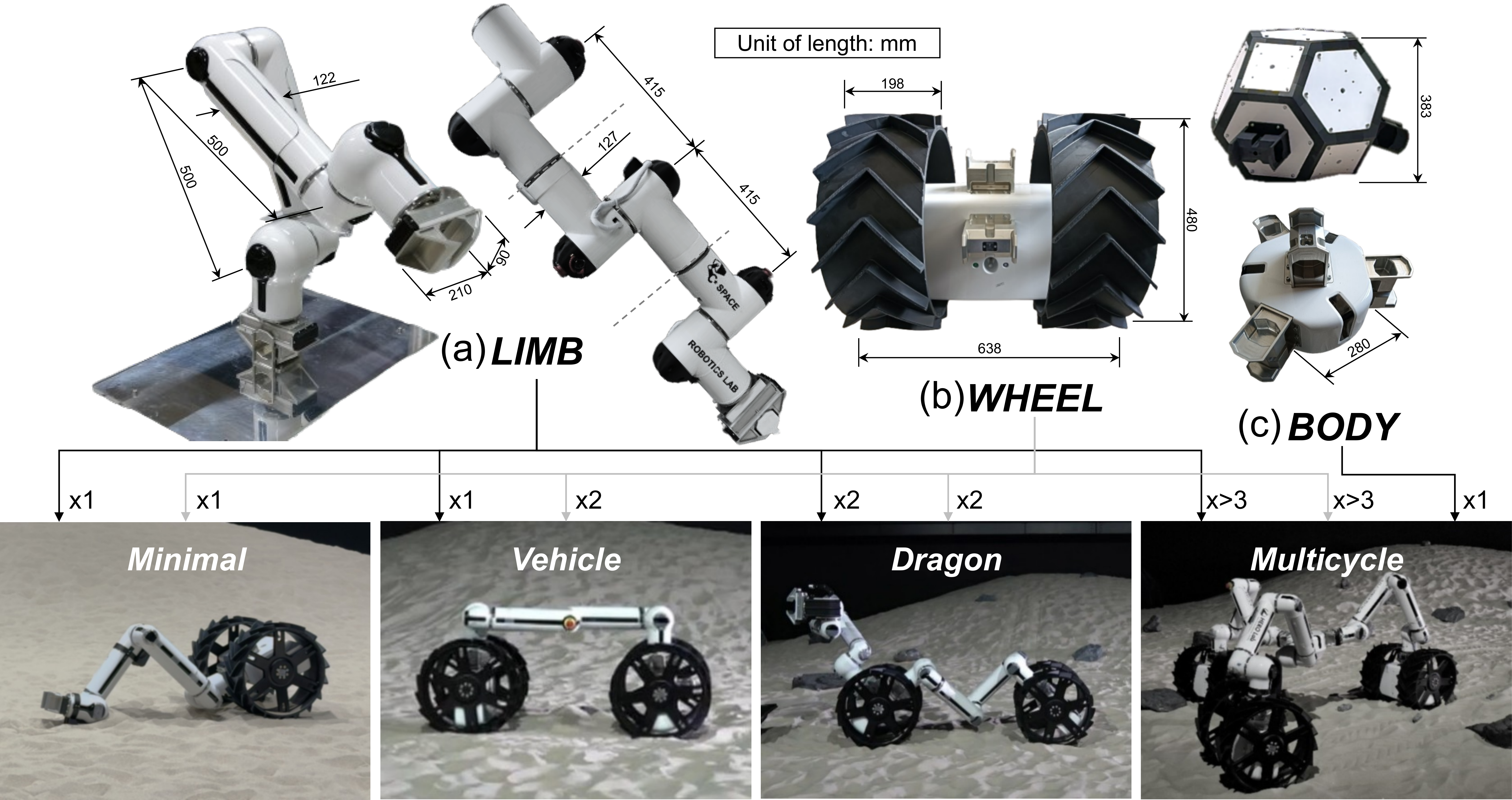}}
    \caption{MoonBot modules developed. (a) The Limb module is a symmetrical seven-degree-of-freedom (7-DOF) articulated system capable of equipping a gripper at both ends, allowing it to function as either a leg or an arm. The parallel jaw gripper has a maximum opening distance of 80\;mm. (b) The Wheel module serves as a mobile base equipped with two independently rotating grousered wheels for high-speed locomotion. It features grapple fixtures that allow Limb modules to connect, enabling serial configurations of limb-wheel couplings. (c) The Body module serves as a central hub for connecting multiple Limb modules. To explore optimal designs, we prototyped various 3D shapes, including a polyhedron (upper) and a cylinder (lower). These body modules can be equipped with a high-capacity battery and a high-performance on-board computer, enabling extended operation and advanced functionality. {Tested morphologies composed of these modules include \textit{Minimal} (1 × Limb + 1 × Wheel), providing the minimum ability for locomotion and manipulation; \textit{Vehicle} (1 × Limb + 2 × Wheels), offering enhanced traversability as a scouting rover; \textit{Dragon} (2 × Minimals connected serially), combining Vehicle-level mobility with a manipulation capabilities; and \textit{Multicycle} (multiple Minimals connected in parallel to a Body module), providing a stable mobile base capable of accommodating additional structures---in the form of parallel actuator or heavier payload.}
    % ; and the \textit{Cargo} (4 × Limbs + 4 × Wheels + 1 × Body), representing a configuration to lift and transport a massive and heavy payload.
\label{moonbot_hardware_overview}}
\end{figure*}
% --------------------------------------------------------------------------

\subsection{CONTRIBUTION}
To authors' best knowledge, while numerous studies on modular robotic systems have discussed their potential application in space construction, most of them are limited within a conceptual study rather than being experimentally demonstrated in realistic and practical scenarios. Our approach is highly mission-oriented, incorporating rapid prototyping and repeated field testing. This paper details the development of a functionality-based modular space robot: MoonBot, and serves as the first field testing report to demonstrate the potential of its heterogeneous modular system to accomplish various milestone tasks toward establishing lunar base infrastructure. 

The field test was conducted in 20\;m $\times$ 20\;m lunar analogue test site, fully covered with silica sand. In-depth discussions of the essential lessons learned throughout this real-world field testing are presented to address the current gap in knowledge regarding the practical utilization of modular robotic systems for space missions. The following sections present the hardware design, the control software stack for teleoperation, and the results from the lunar analogue field demonstrations. {All the lessons learned throughout the field tests will be detailed to address the current knowledge gaps in the utilization of the heterogeneous modular robots for lunar mission.}

%%%%%%%%%%%%%%%%%%%%%%%%%%%%%%%%%%%%%%%%%%%%%%%%%%%%%%%%%%%%%%%%%%%%%%%%%%%%%%%%%%%
%%% ROBOT HARDWARE DESIGN
%%%%%%%%%%%%%%%%%%%%%%%%%%%%%%%%%%%%%%%%%%%%%%%%%%%%%%%%%%%%%%%%%%%%%%%%%%%%%%%%%%%
\section{ROBOT HARDWARE DESIGN}
% This project investigates a modular robotic system designed to adapt to diverse tasks and environmental conditions through the reconfiguration of its components.
{MoonBot is a heterogeneous modular robotic system composed of different functionality-based modules, enabling the robot to modify its physical structure and locomotion modes in response to specific operational requirements.}
% The system's modularity allows individual modules to attach, detach, and be substituted as needed

% This modular approach is exemplified by the MoonBot platform, which features inter-module connection interfaces that facilitate the assembly of various configurations tailored to different tasks. 
% For instance, a heavy-lifting robot can be constructed by integrating stronger manipulators, while a snakelike unit for cave exploration can be achieved by combining multiple arm modules. 
% This flexibility is evaluated at two levels of modularity.

% Firstly, a fully operational 7-degree-of-freedom (7-DOF) robotic arm
% % , referred to as MoonBot limbs and developed by HERO Lab, 
% are used. Additional components, such as wheel units and a central body, were incorporated to facilitate the assembly and assessment of multiple robotic configurations. This setup allows for the evaluation of the system’s ability to perform a variety range of tasks with a modular framework.

% Secondly, smaller-scale modular components, termed reduced MoonBot modules, were developed to explore configurations with different degrees of freedom and limb lengths. These modules, including 3-DOF and 1-DOF units, provide a platform for testing and optimizing the system's adaptability.

% This dual-level approach highlights the potential of modular systems to meet various task requirements with a reduced need for specialized robots. 

The following sections {detail the hardware design and system integration of the MoonBot platform, which has been preliminarily developed as a ground-testing prototype to demonstrate the capability of a heterogeneous modular robot as an executor of lunar mission tasks. Therefore, at this stage, our primary focus was on demonstrating the functionality and feasibility of the modular reconfigurable robotic system in a lunar analogue environment. Consequently, environmental qualification factors such as resistance to launching and landing vibration, dust contamination, thermal cycling and radiation were not fully addressed in the present study. Nevertheless, these aspects remain critical for advancing MoonBot toward actual space missions. As part of our future work to develop a “space-ready” MoonBot, we plan to conduct dedicated qualification tests, including vibration experiments, dust-proof evaluations, thermal vacuum testing in environmental chambers, and radiation exposure assessments, in collaboration with specialized facilities. Following these efforts, the first lunar mission of MoonBot is envisioned, in which a minimal number of modules will be launched to the Moon to validate in-situ self-assembly, reconfiguration, locomotion, and manipulation. Once these fundamental capabilities are verified, additional MoonBot modules will be deployed to progressively undertake the actual construction of lunar infrastructure.}

\subsection{LIMB MODULE}
\subsubsection{OVERVIEW}
\begin{figure}[t]
    \centerline{\includegraphics[width=\linewidth]{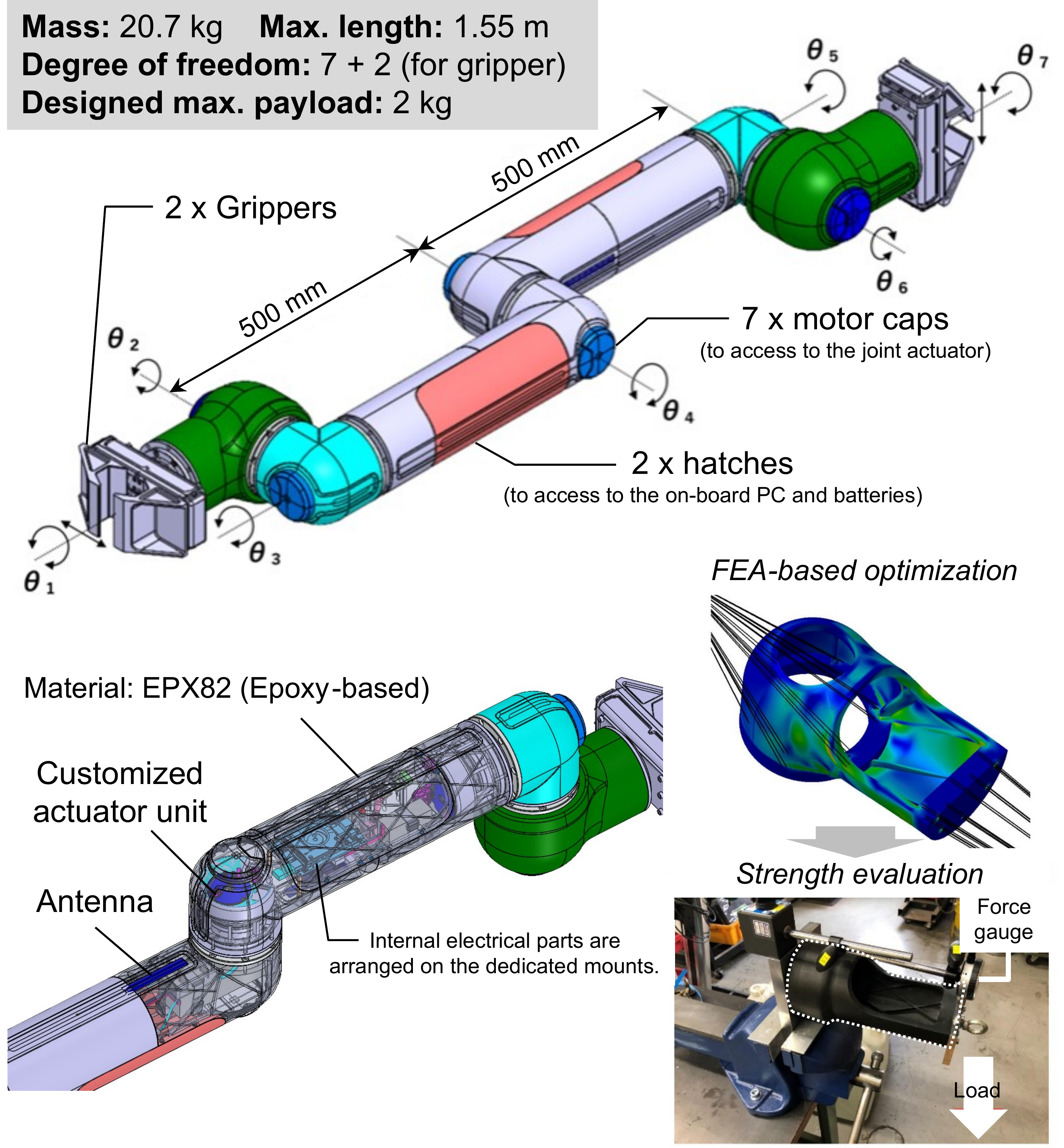}}
    \caption{\label{fig:hero_limb_design} 7-DOF limb module design. Lightweight yet high-strength housing is produced by 3D printing with stereolithography (SLA) material.}
\end{figure}

The MoonBot limb system is engineered to be compact and lightweight, measuring 1.55\;m in length and weighing 20.7\;kg on Earth (\fig{fig:hero_limb_design}). It can transport objects weighing up to 2\;kg, suitable for lifting small payloads. The robot's arm possesses seven degrees of freedom (7-DOF), enabling a diverse range of motion. It is equipped with one 1-DOF gripper at each end for efficient object handling and manipulation. As the Limb module is longitudinally symmetrical, these points are interchangeable depending on which end acts as the end effector, and which is attached to the body module. The system employs brushless direct current (BLDC) motors with the gearbox of the high reduction ratio (1:960). %for seamless and economical operation. %accurate speed and torque regulation under various load conditions. BLDC motors are compact and provide high power density, vital for the weight constraints in space robotic missions. Importantly, they are low-maintenance and have an extended lifespan, contributing to the longevity of the MoonBot platform.

Power is provided by two separate lithium polymer (LiPo) battery lines: a 11.1\;V for the on-board computer and sensors, and a 22.2\;V for the actuators, allowing untethered operation for a couple of hours. The batteries and on-board computer are stored in the compartment space located inside of the longest two links, quickly accessible by opening hatches. The small hatches open to the actuator modules of their corresponding links. The hatches are held shut by snap-fit lids and magnets to aid in alignment.
% When the batteries run out of power, operators can exchange them for fresh batteries through quick-access hatches. The robot is powered by a rechargeable battery, enabling it to function independently of a power source, therefore making it suitable for diverse environments. 
%The robot employs Wi-Fi modules for communication with the control station and utilizes infrared LEDs for inter-module connection monitoring.
% proving particularly advantageous in situations laden with dust or particulate matter such as regolith. 
The primary control unit is a LattePanda Alpha DFR0547~\cite{LattePanda} board installed with Ubuntu 20.04 LTS, which oversees all the robot's operations. An internal USB hub is included, allowing communication, sensory, and other peripherals to be added as more functions are added to the MoonBot.
% Access to the internal components of the Limb module is available through quick-access hatches found on four points on the limb: two large hatches on the side of the longer limb link sections, and two small circular hatches on points that would correspond to the ``wrist'' and ``shoulder'' of a typical robot manipulator. 
% The large hatches give access to the battery compartment and computer compartments, and the small hatches open to the actuator modules of their corresponding links. The hatches are held shut by snap-fit lids and magnets to aid in alignment. This allows quick battery exchange and maintenance, reducing downtime between tests to just a few minutes. --> This is merged into the above text.
% --------------------------------------------------------------------------
% --------------------------------------------------------------------------
% % --------------------------------------------------------------------------
% \begin{figure}[t]
% 	\centerline{\includegraphics[width=\linewidth]{fig/strength_eval.png}}
% 	\caption{\label{fig:strength_eval}Lightweight, High-Strength Housing: SLA 3D Printed Design for Optimal Performance }
% \end{figure}
% % --------------------------------------------------------------------------

The robot is fitted with multiple sensors: motor current sensors, joint angle sensors employing Hall effect to ascertain the arm's position, photo-reflective sensors for defining each joint's zero position, an Inertial Measurement Unit (IMU) for sensing the robot's posture, and a battery level voltage gauge for tracking power consumption. Teleoperation of the MoonBot is accomplished via wireless communication using a Wi-Fi antenna mounted on the outer surface of the link casing. {The end-effector of the Limb module utilizes infrared LEDs for inter-module connection monitoring.}

% The robot incorporates a lightweight design utilizing resin casings and duralumin, thereby minimizing its overall weight while maintaining structural integrity. 
% and possesses robust communication capabilities among its units, even in adverse situations

The Limb unit was designed to balance lightweight construction with high structural strength. The housing is manufactured using stereolithography (SLA) 3D printing, which enables the creation of complex, high-precision geometries while the internal mechanical structure was made of milled duralumin. This manufacturing method allows for significant weight reduction while maintaining durability. The primary material used in the limbs is EPX82~\cite{Carbon_EPX_82}, an epoxy-based resin known for its superior mechanical properties, including high tensile strength and resilience, further contributing to the limb's overall robustness.

An integrated design approach combines the structural framework with electrical component fixtures, eliminating the need for separate mounting hardware. This monolithic structure not only simplifies assembly but also enhances stability, reducing the impact of mechanical vibrations and external shocks during operation. To ensure optimal structural performance, the arm unit’s geometry has been refined through finite element analysis (FEA)~\cite{FEA}. The FEA process identified areas of high stress and allowed for shape optimization, maximizing strength while minimizing unnecessary material usage. The prototyped housing was then assessed through manual loading tests.

% \todo{redundant section, needs revision}
% The effectiveness of these design optimizations was validated through experimental strength evaluations, where the arm unit was subjected to various load conditions. These tests confirmed that the design meets the required performance standards, maintaining structural integrity under maximum load conditions. The successful integration of lightweight materials, advanced manufacturing techniques, and shape optimization ensures that the robotic arm delivers efficient and reliable performance while remaining lightweight and adaptable.

% Overall, the combination of cutting-edge materials, advanced structural optimization, and experimental validation further enhances the robotic system’s capabilities. The arm unit’s design allows it to withstand rigorous operational demands, making the system a versatile and dependable solution for applications requiring precise manipulation and operation in diverse and challenging environments.

\subsubsection{JOINT ACTUATOR UNIT}
% --------------------------------------------------------------------------
\begin{figure}[t]
	\centerline{\includegraphics[width=\linewidth]{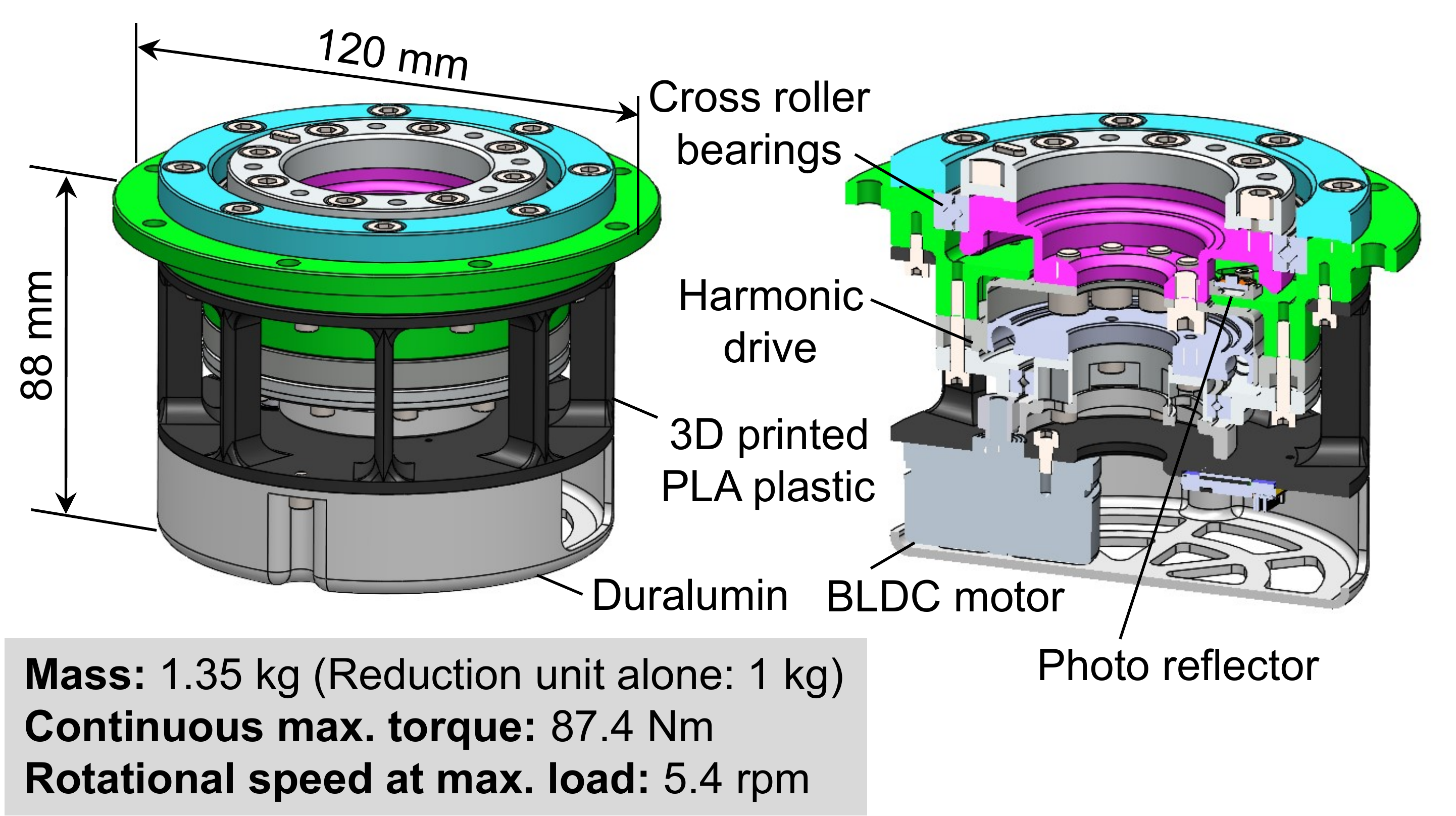}}
	\caption{\label{fig:joint_actuator_unit}Joint actuator unit design. Lightweight yet robust design was achieved using super duralumin and carbon-infused resin.}
\end{figure}
% --------------------------------------------------------------------------
The joint actuator unit for the limb module was designed to use a lightweight structure, employing super duralumin and carbon-infused resin components (see \fig{fig:joint_actuator_unit}). The total unit weight is 1.35\;kg, with the reduction unit weighing 1\;kg. For the actuators, BLDC motors: Maxon EC45 flat were chosen; the 50\;W model was used at roll joints ($\theta_1$, $\theta_3$, $\theta_5$, and $\theta_7$) and 80\;W was used at pitch joints ($\theta_2$, $\theta_4$, and $\theta_6$), which require high torque. Notably, this miniaturized actuator can be attached outside of the reduction gear mechanism, enabling all the harnesses to go through the central hollow. Total reduction ratio of $1:960$ was achieved by the combination of the harmonic drive with the reduction ratio of $1:160$ and regular gear with the reduction of $1:6$. Despite its low weight, the actuator delivers a constant maximum torque of 87.4\;Nm, allowing it to maintain substantial forces during operation. The actuator can attain a maximum rotational speed of 5.4\;rpm (revolutions per minute) under full load, ensuring both strength and precision. The actuator unit features cross-roller bearings, which provide substantial load capacity and seamless transmission, guaranteeing efficient and stable joint action. 

A harmonic drive is incorporated into the system to deliver exact gear reduction, improving the actuator's capacity for precise motion control. The BLDC motor functions as the primary drive component, providing high torque-to-weight ratio and precise speed control. Additionally, a photo-reflector is used to create a reference position when calibrating joint states, preventing position drift.

% \todo{this paragraph of the summary of the joint actuator would be too redundant.}
% The joint actuator unit integrates innovative materials and high-performance components to deliver a lightweight and powerful solution for robotic applications. Its durable construction and effective operation under strict weight and performance requirements in the simulated lunar construction scenarios.

\subsubsection{GRIPPER}
% --------------------------------------------------------------------------
\begin{figure}[t]
	\centerline{\includegraphics[width=\linewidth]{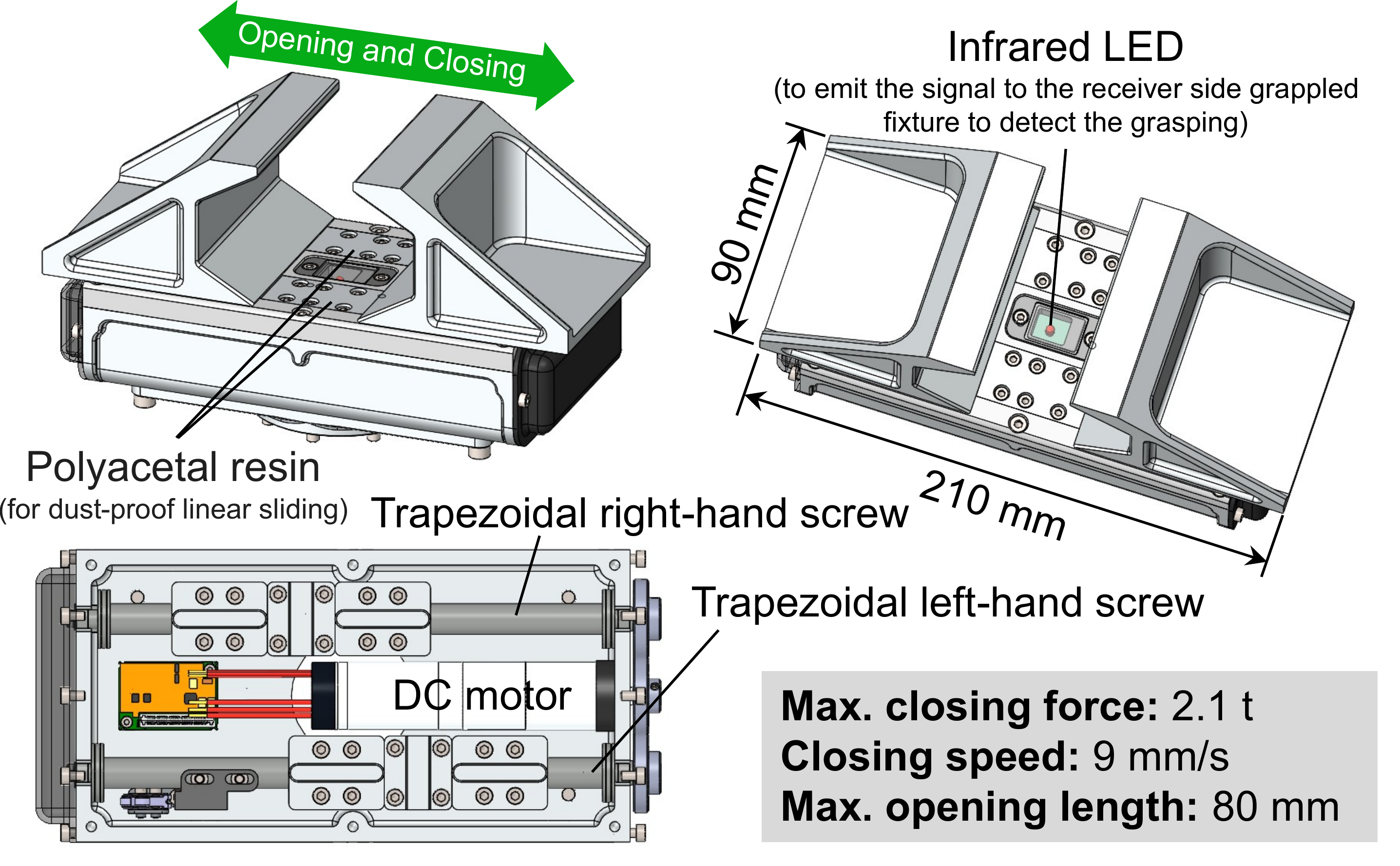}}
	\caption{1-DOF pinching gripper design. The minimized clearance achieved using polyacetal resin with sufficiently low friction prevents dust contamination of the mechanism during operation in lunar sand. \label{fig:Ggripper_Mechanism}}
\end{figure}
% --------------------------------------------------------------------------
The gripper mechanism of the robotic arm is a key because it is used to connect the other modules (body and wheel modules) to reconfigure the mode of the modular robotic system as well as purely used for the manipulation tasks. {For this purpose, MoonBot's gripper is dedicated to stable and strong connection with the other module when it is fully closed.} Essential to its operation is the capacity to open and close with high accuracy and strong pinching force. The mechanism attains a maximum gripping force of 2.1\;tons, enabling it to securely grasp and operate heavy objects while preserving stability and control (see \fig{fig:Ggripper_Mechanism}).

Additionaly, this end-effector is a part that is most exposed to the regolith dust while performing tasks on the Moon's sandy environment. To mitigate the problems presented by regolith, including abrasive dust and fine particles commonly encountered in interplanetary environments, the gripper integrates many countermeasures. A notable characteristic is the dustproof linear sliding mechanism, employing Polyoxymethylene (POM) with minimal tolerances. This design mitigates dust infiltration and diminishes wear on moving parts, hence ensuring prolonged performance and durability of the gripper in dusty conditions. {POM was specifically employed for this sliding component due to its low-friction performance, whereas most of the other parts were prototyped using polylactic acid (PLA).} 

Detection of the gripping of the grapple fixture is enabled via infrared sensors without physical connectors. This wireless communication technique is especially beneficial in settings where dust and debris may disrupt conventional wired connections.
% guaranteeing seamless functioning and coordination among various components of the robotic system. % --> would be too redundant
The gripper mechanism functions at a velocity of 9\;mm/s, ensuring an optimal balance between speed and precision. This facilitates rapid yet exact movements, crucial for activities demanding both efficiency and precision. The gripper's actuation is driven by a DC motor: Maxon EC45 flat, selected for its dependability and controllability, enhancing the mechanism's overall responsiveness. The trapezoidal screws on both the left and right sides of the gripper are essential to convert the rotational motion of the DC motor into linear motion for the gripper's opening and closing operations. The trapezoidal configuration of the screws guarantees smooth and uniform motion, minimizing backlash and improving the accuracy of the gripping mechanism. 

\subsection{MODULAR LIMB}
% Secondly, smaller-scale modular limbs were developed to explore configurations with different degrees of freedom and limb lengths. These modules, including 3-DOF and 1-DOF units, provide a platform for testing and optimizing the system's adaptability.
In addition to the full 7-DOF limb module, we have pursued the development of a modular limb system composed of lower-DOF articulated units. This approach examines modularity not only at the functional system level but also through the development of smaller-scale modular units. These units are designed to provide flexibility in configuring various degrees of freedom, thereby enhancing the overall reachability of the robotic limb. A representative of our complete modular robotic limb features a 7-DOF configuration achieved by integrating three modular units: two 3-DOF modules and one 1-DOF module. The 3-DOF modules are arranged in a roll-pitch-roll configuration, while the 1-DOF module provides a pitch movement, collectively supporting a wide range of configurations.

% {In MoonBot design, we consider it essential to include a reduced-DOF limb option in order to avoid an over-specified, overweight, and complex morphology. While this paper presents one possible example of a modular limb as a starting point, there remains a significant design space to explore in developing reduced-DOF limb modules. Such modules could be employed (instead of using full 7-DOF limbs) in certain parts such as a bridging limb of Vehicle and Dragon.}

{
Although this study highlights meter-scale 7-DOF limbs, it is essential to pursue heterogeneity within the MoonBot platform. 
Lower-DOF limbs offer advantages in reducing weight and mechanical complexity, particularly in assemblies requiring only basic control. In many cases, rotational actuation of the wheel's vertical axis is sufficient, making additional degrees of freedom redundant. Future iterations of MoonBot are therefore actively investigating simplified limb designs, for simplified assemblies.
}

\subsubsection{REDUCED DOF LIMB MODULES}
% --------------------------------------------------------------------------
\begin{figure}[t]
    \centerline{\includegraphics[width=\linewidth]{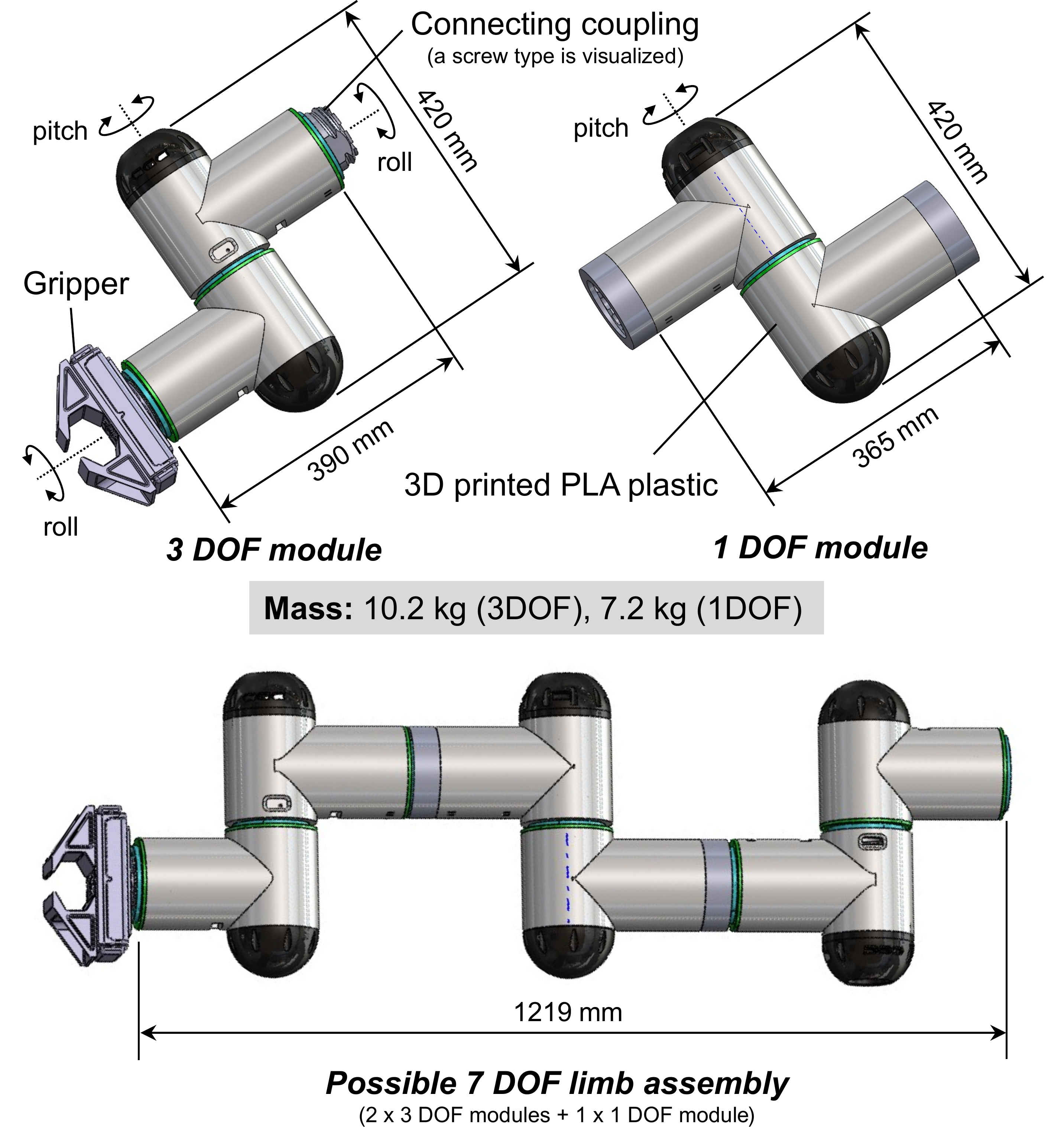}}
    \caption{Modular limb design, prototyped to study in to the lower level of modularity. This first prototype was designed to maintain the scale of a full 7-DOF limb module while accommodating intermodule separability. \label{fig:moonbot_g_modular_limb_design}}
\end{figure}
% --------------------------------------------------------------------------
The mechanical design of the reduced DOF modules emphasizes lightweight construction, compact form factor, and modularity. The prototyping of this modular limb was performed using in-house additive manufacturing and commercial off-the-shelf (COTS) electronics. PLA was used to print the body. Each 3-DOF unit measures 450\;mm in length and weighs 10.2\;kg, and each 1-DOF unit measures 450\;mm in length and weighs 7.2\;kg (\fig{fig:moonbot_g_modular_limb_design}), which gives a scale consistency (1.5\;m in length) with the full 7-DOF module.

Each module incorporates BLDC motors (model: MN8017) attached to the same harmonic drive used in the full 7-DOF limb module for the joint actuators. In one of the roll joints of the 3-DOF units, the same gripper can be mounted, while a coupling for module connection is attached to the other roll joint, providing an interface for linking with subsequent modules.
% This strategic arrangement of actuators improves the versatility of the module, making it suitable for various tasks. 
The BLDC motors are controlled by ODrive ODv44-ST motor controllers~\cite{Odrive} that provide current feedback and allow for position, speed, and current control. The actuators also include 14-bit resolution encoders, enabling closed-loop control through field-oriented control (FOC) algorithms in the Odrive. The robot is powered by two 6-cell LiPo batteries, each with a voltage rating of 22.2\;V, which are connected in series. These batteries are securely housed within a 3D-printed case that is integrated into the module's battery. The module includes components such as a battery level indicator, which monitors the battery voltage, and an emergency stop switch for safety.
% , ensuring fail-safe functionality by immediately halting operations in the event of a malfunction.
The primary control unit is a Latte Panda 3 Delta board, which communicates with the Odrives via CAN bus.
% and allows the control of the units through the ROS framework. 
This on-board computer is equipped with Wi-Fi connectivity, enabling remote control and real-time command reception.

\subsubsection{INTERMODULE CONNECTORS}
\begin{figure}[t]
    \centerline{\includegraphics[width=\linewidth]{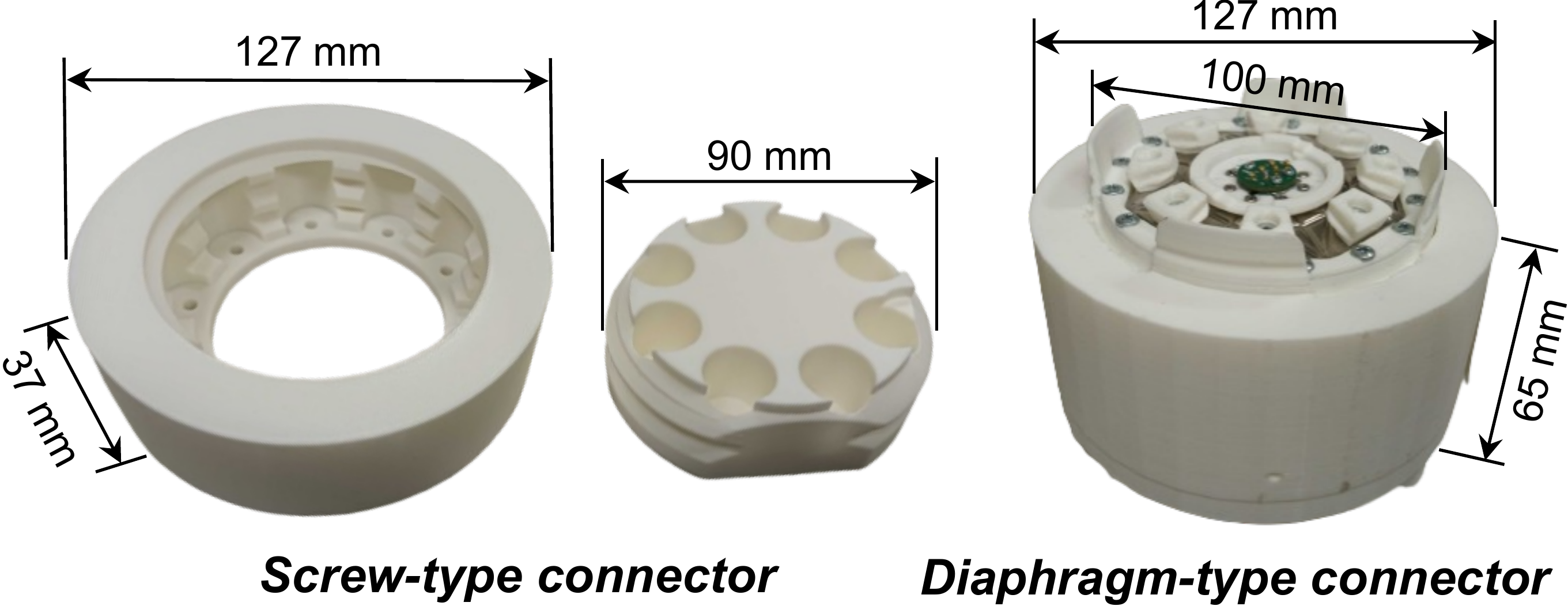}}
    \caption{\label{fig:moonbotG_connectors} Two types of connectors are used for the preliminary trials for modular limb development.}
\end{figure}

We use two types of connector for the assembly of the modular limb. They are a screw-type connector and a diaphragm-type connector (\fig{fig:moonbotG_connectors}). 

The screw-type connector, inspired by the traditional nut-and-screw mechanism, is a custom-engineered, 3D printed solution designed specifically to interconnect modular units. This mechanism consists of two interlocking components: a male part integrated into one module and a female part embedded in the adjoining module.
% --------------------------------------------------------------------------
% --------------------------------------------------------------------------
Because of the passive nature of the connector, the printing tolerance of the parts is very critical to ensure the proper attachment of the modules during operation. 
% The current 3D printed screw connector, made from PLA, faces durability issues due to the material’s susceptibility to wear and tear from friction during repeated use. PLA, while suitable for prototyping, experiences thread degradation, misalignment, and deformation from the repetitive motion of screwing and unscrewing.
% This results in a limited lifespan and reduced mechanical strength.

% The modularity of this design allows for straightforward disassembly, facilitating easy system reconfiguration, repair, and transport. This innovative, 3D-printed screw connector exemplifies advancements in modular design, offering a lightweight, durable, and cost-effective solution for the reliable interconnection of components.

% To improve durability, we propose using more robust materials such as Nylon, Polycarbonate, or metal alloys. Surface coatings or lubrication can be applied to reduce friction and wear. Reinforced thread designs or self-locking threads would enhance mechanical integrity and prevent thread stripping, thereby extending the connector’s lifespan.

We also tried incorporating a {genderless} diaphragm-type connector, which is the same type Nunziante~et al. used in~\cite{luca_SII}, and weighing 0.6\;kg. As the modules are brought together, the locking claws align with their counterparts, facilitating a secure mechanical connection. {In~\cite{diaphragm}, an improved version of the diaphragm-type connector, equipped with an error-absorption mechanism, is described in detail.}

% This precise alignment is crucial to ensure that the modules are connected in a stable and reliable manner. Although the connector is designed to connect individual robotic modules, its load-bearing capacity is limited by the design of the locking components and the size of the actuator. Any misalignment of the locking claws results in an incomplete or weak connection, compromising the integrity of the robotic system.

% \subsubsection{INTEGRATION OF 7DOF LIMB USING DIVERSE CONNECTOR SYSTEMS} % this means MoonBot G
% The 7-DOF limb consisted of a 1-DOF module and two 3-DOF modules. The 1-DOF module is equipped with diaphragm-type connectors on both ends. 
%However the screw-type connector proved to be the most robust to assemble the whole 7-DOF limb. 
% The fully assembled limb using the screw-type connectors is shown in \fig{moonbot_hardware_overview}.
%he second configuration uses screw-type connectors on both ends, while the third combines a diaphragm-type connector on one end with a screw-type connector on the other.

\subsection{WHEEL MODULE}
% --------------------------------------------------------------------------
\begin{figure}[t]
    \centerline{\includegraphics[width=.95\linewidth]{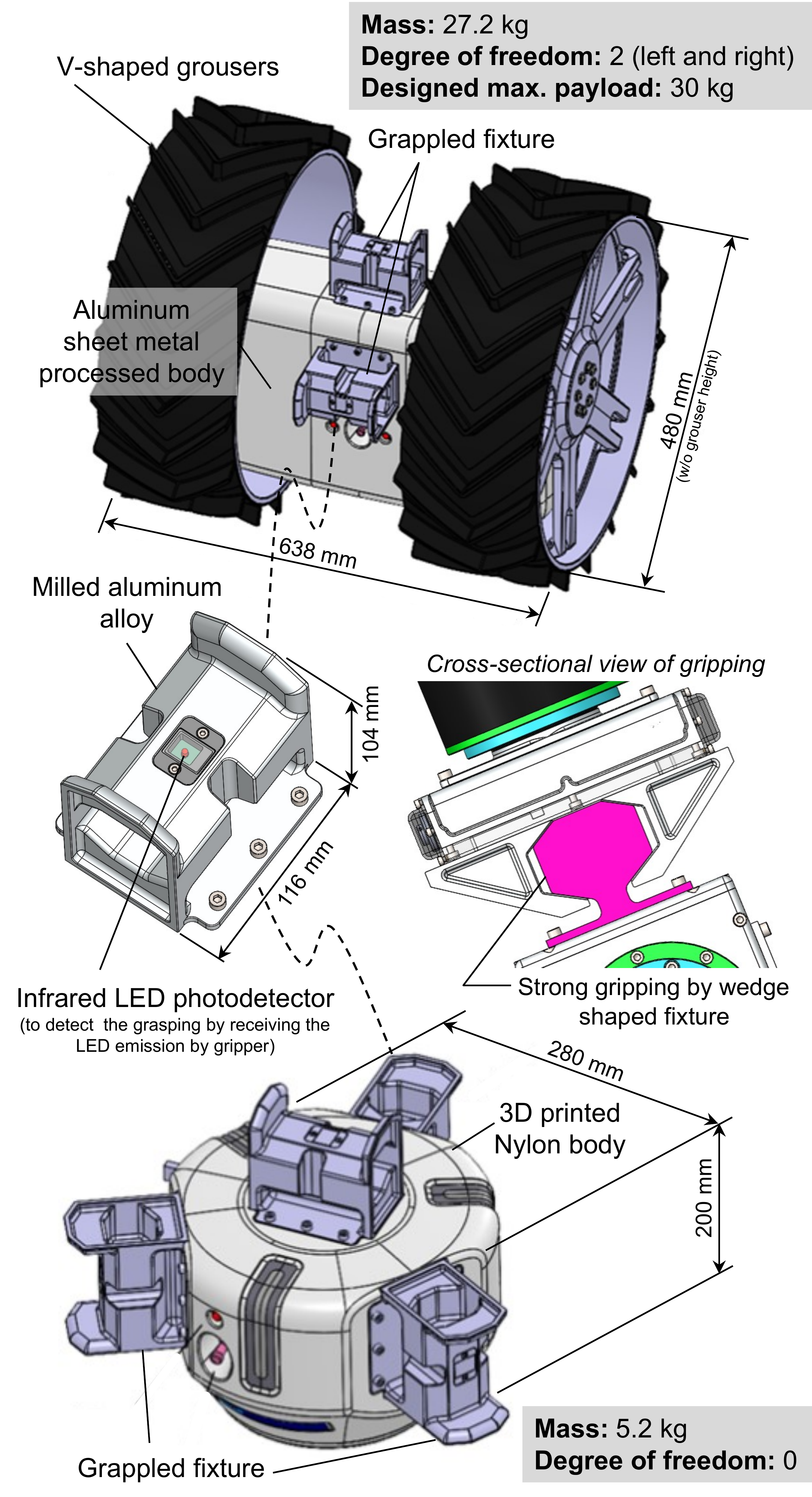}}
    \caption{Wheel and Body module design. The Wheel module is equipped with two grapple fixtures, while the Body module features four grapple fixtures. An on-board computer is mounted inside of the each system.
    \label{fig:wheel_and_body_module}}
\end{figure}
% % --------------------------------------------------------------------------
% \begin{figure}[t]
% 	\centerline{\includegraphics[width=\linewidth]{fig/body_unit.png}}
% 	\caption{\label{fig:body_unit}Body Unit Design \todo{to be improved}}
% \end{figure}
% % --------------------------------------------------------------------------
The wheel module is engineered to provide the standalone mobility and maneuverability on the terrain while providing sufficient stability to support limbs, all within a compact, lightweight design. With a width of 638\;mm and wheels measuring 480\;mm in diameter, the module attains an optimal form factor for efficient maneuverability. Despite its modest weight of 27.2\;kg, it can accommodate a target payload of 30\;kg, facilitating the effective carriage of larger items while ensuring smooth movement (see \fig{fig:wheel_and_body_module} top).

The wheel module incorporates a 2-DOF configuration with two independently actuated wheels. This arrangement facilitates precise motion control, enabling both forward and backward movements as well as skid-steering via differential control. Such a design affords exceptional agility, making the module highly effective at maneuvering through confined spaces and complex landscapes. The module operates using BLDC motors (Maxon EC45 flat), which deliver sufficient torque and speed to propel the wheels while maintaining low energy consumption and minimizing maintenance needs, hence improving overall system durability.

The module's power supply utilizes a rechargeable LiPo battery, enabling independent operation without dependence on external power sources. This feature renders it ideal for field operations in isolated or inhospitable environments where uninterrupted power availability may be impractical. The battery offers adequate capacity for prolonged operational durations, guaranteeing steady and uninterrupted performance.

The core of the wheel module's control architecture is also the LattePanda Alpha control board, which oversees actuation, communication, and sensor integration systems. This board functions as the central processing unit, managing real-time activities and ensuring accurate coordination of all components. The wheel module is additionally outfitted with several embedded sensors that augment its functionality. Motor current sensors assess the performance and efficiency of BLDC motors while employing Hall effect sensors to monitor the wheel rotation angle. An IMU is utilized to assess the orientation and posture of the module, delivering feedback for the preservation of balance and fluid mobility. Furthermore, a battery level indicator tracks power usage.

Two of the grapple fixtures are mounted on the body of the wheel module, which can be grasped by the limb, enabling MoonBot to reconfigure its morphology in various ways (see \fig{moonbot_hardware_overview}). The choice of two grapple fixtures was made to enable the sequential connection of the limb and wheel module, inspired by Souki-II, a reconfigurable rover~\cite{soukiII}. Every grapple fixture is equipped with an infrared photodetector at its top, which detects the light signal emitted by the gripper to confirm the connection between the wheel and the limb. Except for this inter-module connectivity detection, no signal and power transmission is carried out through this grappling system. Instead, all sensory data and control signal communications are secured via a Wi-Fi based wireless network connecting the modules and the operation PCs.
% The on-board PC module is equipped with Wi-Fi for communication. Control commands are conveyed by Wi-Fi, facilitating effortless wireless remote operation. Simultaneously, infrared sensors enable inter-unit communication, guaranteeing strong and dependable data sharing even under adverse conditions, such as locations with elevated dust levels or particulate matter. This dual communication mechanism improves the module's adaptability and robustness in various operational conditions.

\subsection{BODY MODULE}
The MoonBot body module is designed as a minimal core unit, offering a compact and lightweight solution for housing vital control and communication components. The module measures 200\;mm in height, 280\;mm in width, and 280\;mm in depth, and features an axial symmetric shape. Weighing 5.2\;kg, it is equipped with four grapple fixtures---three on the sides and one on the top---identical to those found on the wheel module (see \fig{fig:wheel_and_body_module} bottom).

The body module has zero degrees of freedom, serving primarily as the structural and communication hub within the robotic framework. Its power is sourced by an on-board battery rated at 7.4\;V.
% , guaranteeing autonomous operation without dependence on other energy sources. This configuration enables uninterrupted operation in diverse settings, especially in rural or isolated areas where consistent power supply may be lacking.
% The body unit utilizes a dual-channel system for communication. Control commands are transmitted wirelessly through Wi-Fi, facilitating efficient remote operation and real-time data interchange between the body and external control systems. Furthermore, infrared sensors are incorporated to enable inter-unit communication, guaranteeing the smooth transmission of essential information between neighboring modules. This dual method improves system reliability, especially in situations where wired connectivity may be hindered or impractical due to dust or physical limitations.
As the on-board computer, the XIAO ESP32C3 control board~\cite{ESP32}, a compact and high-performance microcontroller, was to provide the control and processing capabilities for the body module. However, a more powerful PC can also be mounted to support advanced processing task and handle large volumes of sensory data. 
% This board oversees the unit's communication protocols, sensor integration, and overall operations. The incorporation of this board guarantees effective data processing and synchronization with other robotic subsystems, enhancing the overall system's stability and responsiveness.
Two essential sensors are embedded in the module: an IMU, which evaluates body posture and orientation to provide real-time feedback essential for stability and alignment, and a battery level indicator, which tracks power usage and reports the remaining charge in real time.
%an IMU is to assess the body's posture and orientation, delivering real-time feedback that is crucial for maintaining system stability and alignment, and a battery level indicator monitors power consumption and provides real-time information on the remaining charge. 
% This guarantees the robotic system functions well through proactive battery management and reduces the likelihood of unforeseen power failure.

Furthermore, the body module's essential purpose is the wireless transfer of modular robot's self-status information, i.e., the identification (ID) of neighboring modules. This capability supports real-time monitoring of system integrity and enables synchronized operation among multiple robotic modules. To achieve this, the body module is equipped with a photodetector-based mechanism for receiving identifying data from adjacent limb modules.
% This facilitates precise identification and transmission of adjacent module IDs, allowing for effortless integration and synchronization within the comprehensive robotic architecture.

%%%%%%%%%%%%%%%%%%%%%%%%%%%%%%%%%%%%%%%%%%%%%%%%%%%%%%%%%%%%%%%%%%%%%%%%%%%%%%%%%%%
%%% SOFTWARE DEVELOPMENT
%%%%%%%%%%%%%%%%%%%%%%%%%%%%%%%%%%%%%%%%%%%%%%%%%%%%%%%%%%%%%%%%%%%%%%%%%%%%%%%%%%%
\section{SOFTWARE DEVELOPMENT} %: MOTION STACK FOR MOONBOT
\label{sec:software}

% A lot was learned from a pure control perspective, but software maintainability and flexibility was found to be of utmost importance for research robotics, and even more for modular robots.

% \subsection{Software introduction? software motivation?}
% \todo{better title}

% The \MS software, developed for the modular \MB platforms, was vigorously tested during this three-week field campaign. Daily hardware and software modifications by a large team revealed challenges that typical development cycles could not foresee. This field test was crucial in refining the software, demonstrating its modularity, robustness, and adaptability for deployment on research-grade robots. We consider the scope of the software field test as larger than the robot being operated in the sandbox. Those 3 weeks also tested the ability of the software to be used, updated and maintained daily by our team facing unforeseen challenges.

The literature on modular robots is sparse and software solutions are often specific to a robot~\cite{control_multi_leg_appli, paik_modrob}. Tools such as \moit provide a useful trajectory planner~\cite{moveit}, leveraging the flexibility and distributed communication of Robot Operating System 2 (\rt)~\cite{ros2}. In a sense, it can be considered ``modular'' because \moit is robot-agnostic and \rt distributed. However, \moit is incompatible with walking robots due to its limitation to a single manipulator and limited capability to perform just in time trajectory modification.

Hence, a custom and modular kinematic software called \MS~\cite{neppel2025motionStack} was employed with the goal of unifying modular multi-legged robot kinematic planning. {The developed software framework coordinates both the 7-DOF limbs presented here and reduced-DOF modules, such as the 3-DOF limbs from the early MoonBot 0 prototype \cite{moonbot_zero}.} This field test was its first intensive use, and this paper details this specific implementation and the lessons learned.

% The software works with Robot Operating System 2 (ROS2)~\cite{ros2}, facilitating easy software integration. 

\subsection{SOFTWARE REQUIREMENT}
{The fundamental software architecture of MoonBot has been integrated to satisfy the following prioritized requirements for the preliminary field demonstration.}

\subsubsection{DEEPLY MODULAR DESIGN}

The \MS is built to accommodate a wide range of robot configurations. Each computer in the system can be assigned the control of specific joints, limbs, or entire robots, enabling distributed computation and control when required. An advantage of this flexibility is its compatibility with a wide range of robots, not only \MB, but also its previous version, and future versions as demonstrated in~\cite{neppel2025motionStack}.
% The \MBH implementation reduces inter-computer data exchanges by processing locally whenever possible. This ensures that communication degradation does not propagate through the whole system.

\subsubsection{OPERATOR CONTROL}

Given that this was the first field test of the \MB, reliability took precedence, and no sensors were {actively} included in the control loop. Instead, operators controlled the robot based on direct visuals or camera feeds. %This approach emphasizes simplicity and robustness.
% real missions will require a combination of human-in-the-loop teleoperation and fully autonomous operation, depending on mission conditions.

To reduce the cognitive load on operators, especially when controlling robots with high degrees of freedom (e.g., 11 motors on the \MBM and up to 33 motors on the \MBT), the \MBS had to provide quick, reactive and intuitive controls. % Binary inputs via a keyboard allow for precise and deliberate actions, while a joystick enables smooth and analog control for dynamic operations. This hybrid control scheme balances precision and fluidity, empowering operators to handle complex tasks with reduced mental strain.

\subsubsection{SAFETY AND RELIABILITY} %in unreliable environment}
\edef\refsafety{Sec.~\thesection.\thesubsection.\thesubsubsection}

Safety is a foundational aspect of the software, Moon robots must be extremely reliable and, during critical field tests, failures are unacceptable. The system ensures the robot always converges to a safe state, with all joint velocities at zero and no significant deviation from the current state, even if software nodes crash or communication is lost. Similarly, when no input is received from the operator, the robot must always stop.
To enhance reliability, processes are designed to continue execution independently whenever possible, avoiding halts even if other processes become unresponsive. Task or state messages are the only dependencies allowed to be waited upon. Mechanisms such as handshakes or ping failures resulting in halted execution were avoided to prevent introducing brittle failure points and unnecessary complexity.

These principles ensure the system remains safe by default without compromising reliability, even in unpredictable environments.

\subsubsection{MAINTAINABILITY AND ROBUSTNESS}

Designing software that functions reliably once, under ideal conditions with expert users, is not enough for such a critical building block of the project. This experiment required the software to work every day for three weeks, as hardware fails, requirements change, new tasks need to be performed, and team members modify hardware and software daily. All of this is also part of the unpredictable environment the robot is part of.

The field test played a pivotal role in improving the \MS, ensuring it is ready for a future open source deployment for research robots.

\subsection{SOFTWARE ARCHITECTURE}
% --------------------------------------------------------------------------
\begin{figure*}[t]
	\centerline{\includegraphics[width=\linewidth]{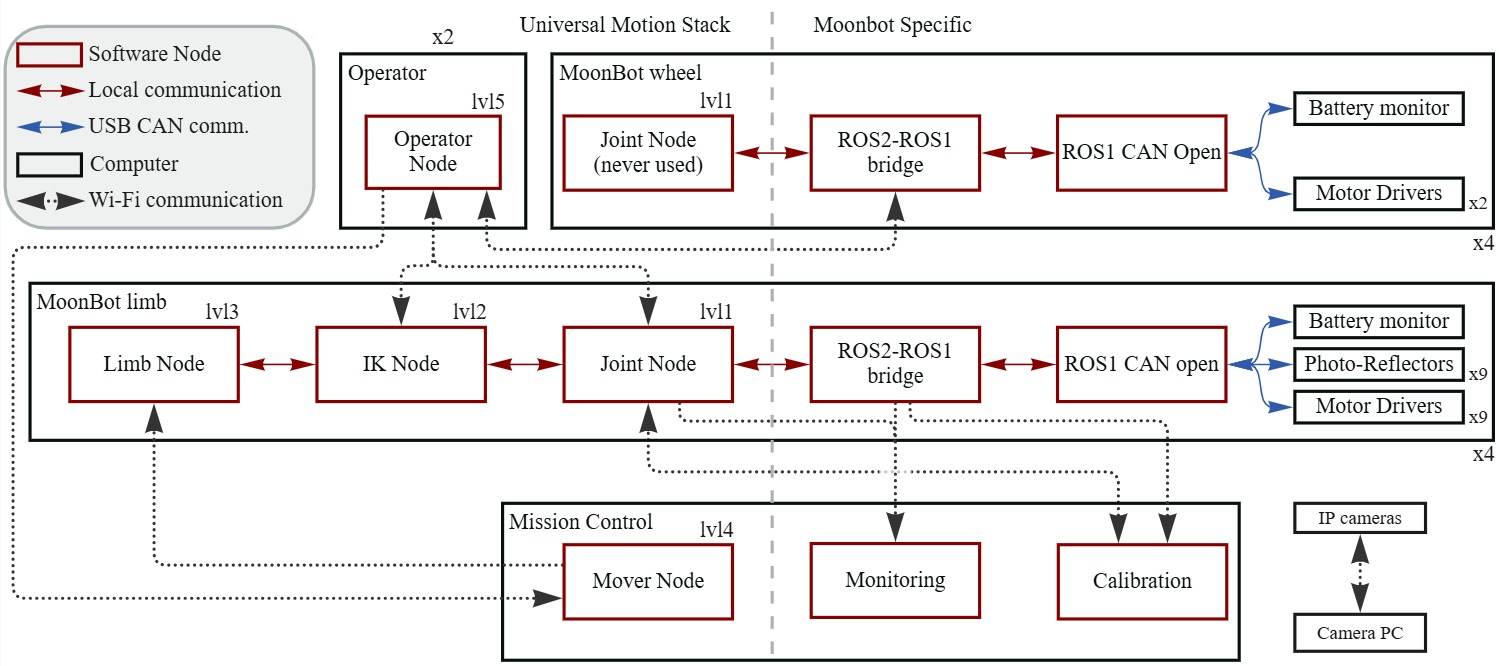}}
	\caption{\label{fig:ms_glob} Global software architecture of the \MB. Blocks represent software nodes or their host computers. Instance counts are shown at the bottom right, and hierarchy levels for \MS nodes at the top right.}
\end{figure*}
% --------------------------------------------------------------------------
\begin{figure}[t]
	\centerline{\includegraphics[width=\linewidth]{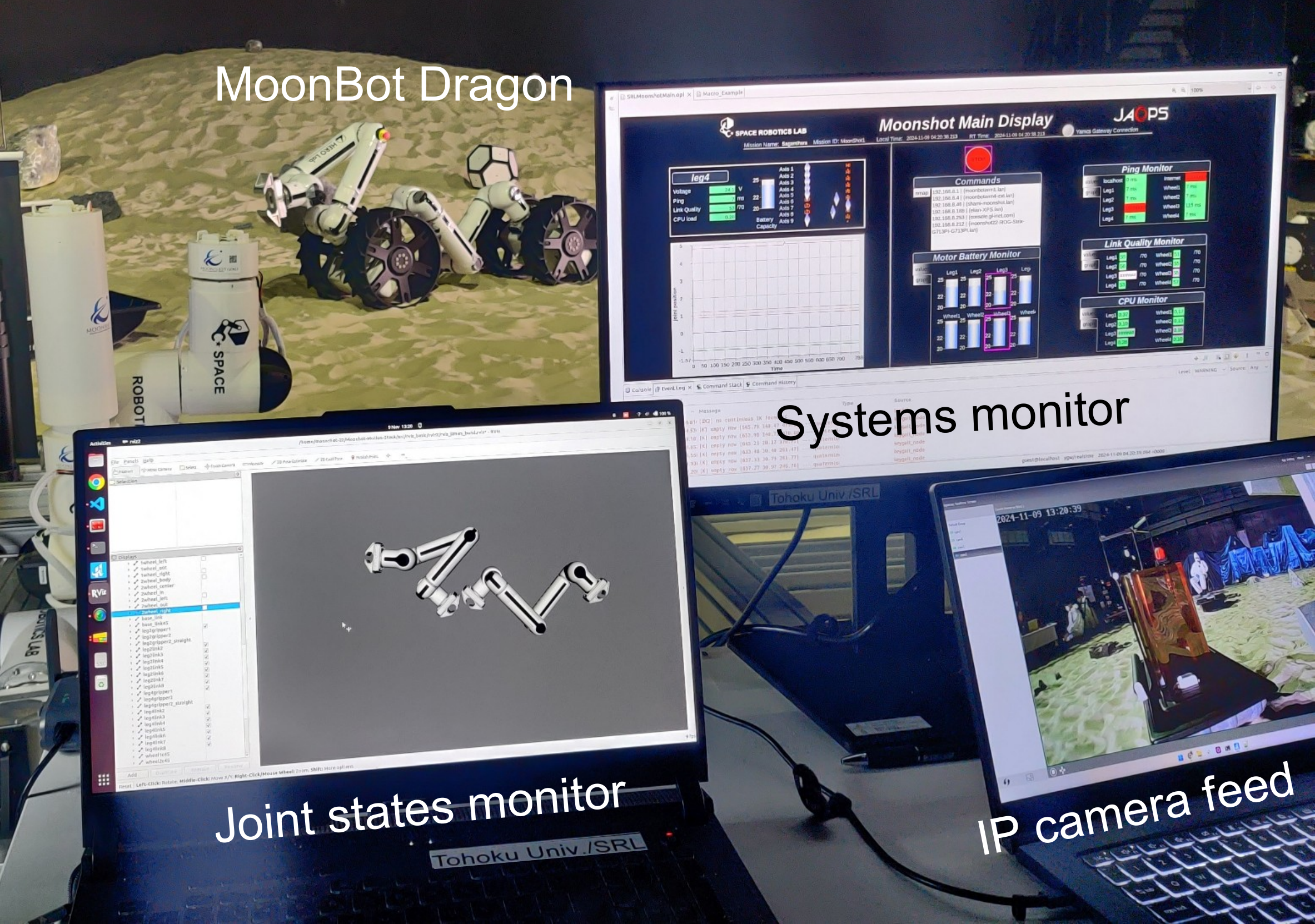}}
	\caption{\label{fig:dragon_monitoring} Mission control monitoring displays and IP camera feed in front of the \MBD during operation.}
\end{figure}
% --------------------------------------------------------------------------
The full software architecture, with all modules and devices connected to the network, is shown in \fig{fig:ms_glob}. Communication between nodes, including over Wi-Fi, is managed by \rt. For the \MB wheel and limb, a single \ro \textit{CANopen} node handles Controller Area Network (CAN) communication with up to nine motor drivers, seven photo-reflectors, and a battery monitor. The \rt--\ro bridge ensures \ro messages are seamlessly relayed to \rt, integrating them into the \MS ecosystem.

The \MS is organized into five hierarchical levels, each comprising specialized nodes:
\begin{itemize}
	\item Level 1: Joint Node -- Sends appropriate commands to individual joints.
	\item Level 2: Inverse Kinematics (IK) Node -- Computes IK solutions for joints based on end-effector target.
	\item Level 3: Limb Node -- Manages long-running trajectories for a limb.
	\item Level 4: Mover Node -- Synchronizes the movements of multiple legs.
	\item Level 5: Operator Node -- Handles operator input commands.
\end{itemize}

The \MS was not used on the wheels. Speed commands on the wheel were directly sent to the motor driver through the bridge. This goes against our safety principles as, in case of connection loss, the wheel would continue spinning. However, the wheels are designed to spin continuously and slowly; hence, this safety concern was deemed of lesser importance and the architecture was simplified, prioritizing the reduction of failure points.

Level~1-2-3 were running onboard the limb's computer to minimize latency and improve reliability. The mover node of level~4 was running on the mission control operation computer. Its role is controlling several limbs, hence it must run on the most reliable computer possible.
Two operators were active on the field, each with their independent computer running an operator node of level~5. Through keyboard or joystick, it allows for individual joint control (Level 1), IK control (Level 2) one limb and multi-limb control (Level 4) of any limb on the network.

% \todo{note: lvl3 not used because of clamped controller}

To enhance modularity and simplify integration, the camera feed was isolated from other systems. Battery-powered Internet Protocol (IP) cameras connected to an independent Wi-Fi network streamed directly to a dedicated laptop. This setup allowed cameras to be repositioned based on the mission or robot, with the laptop placed optimally for operators. The feed could also be displayed on a screen for visitors, ensuring flexibility and accessibility.

The mission control display, seen in \fig{fig:dragon_monitoring}, featured a custom monitoring interface developed using the YAMCS UI software~\cite{YAMCS_Gateway}, commonly employed in space missions.

This interface provided real-time data, including ping status, Wi-Fi link quality, CPU load, battery levels, and IP addresses for all \MB modules on the network. Additionally, a \rt-YAMCS interface was created and is planned for inclusion in a future update of the open-source YAMCS software.
In parallel, RViz visualized live joint states using the robot's mesh model, though wheels were omitted since the \MS was not used for them. This visualization allowed operators to quickly detect discrepancies between the \MS's robot state and the physical robot, ensuring accurate monitoring during operations.

The calibration node was responsible for determining joint angle offsets, as no absolute angle encoders were available. During calibration, the node executed a custom movement based on the state of a single photo-reflector to home its position. To ensure robustness, especially in the event of network issues, the calibration process was managed on a separate computer.
If network communication was interrupted, \rt would fail to transmit the next joint target, causing the---potentially dangerous---calibration movement to stop automatically.

Finally, all terminal operations were conducted via secure shell (SSH) and tmux, enabling operators and \textit{mission control} to access a shared terminal session on the robots, facilitating seamless collaboration.

\subsection{KEY FEATURES AND TECHNICAL IMPLEMENTATION}

% The inner workings of the \MS will not be detailed in this article, as the focus is on the \MB field experiment.

\subsubsection{ONE CODE-BASE FOR HETEROGENEOUS SYSTEMS}

Maintaining one single code-base was a significant challenge due to the heterogeneity of the hardware and mission scenarios.

Two primary methods were employed to adapt node behavior based on the computer they operated on, both leveraging a unique \PCID environment variable configured in each computer's operating system.

%The first method was used exclusively for the low-level CANopen node altogether with usb channel, as its behavior depended solely on the hardware rather than the \MB configuration. The Photoreflector and battery level were recovered from the usb channel isolated from interference of the CAN channel. The real-time data of serially connected motor drivers connected with photoreflewas retrieved from the register level of motor driver. During installation, the corresponding setting files, based on the \PCID, were retrieved from a small database.

The first method was used exclusively for the low-level CANopen node, as its behavior depended solely on the hardware rather than the \MB configuration. Each motor driver, photo-reflector, and battery monitor had unique identifiers on the CAN and USB bus. Those had to match exactly with the CANopen node. During installation, the corresponding setting files, based on the \PCID, were retrieved from a small database.

Also based on the \PCID, the second method involved launch-time alterations. This modified \rt node parameters, according to the computer, robot assembly, and mission. Through those, the \MS parsed the URDF {(Unified Robot Description Format)} of the assembled robot and assigned a kinematic chain to levels 1 and 2 based on the \PCID. The \PCID also determined which level of the \MS to launch, according to \fig{fig:ms_glob}.

Ultimately, through all the modules and configurations, only five different launch files were necessary. The assembled robot kinematic information is given by the URDF, and the network architecture is set through the launch file that adapts its behavior based on which computer it is running on. Other than those five files, the code base was unique, greatly simplifying development.

While the launch files could have been generated on the fly through direct use of \PY, this enhancement was deemed unnecessary for the first field test, given the limited number of configurations.

\subsubsection{MULTI-ROBOT AND OPERATOR COLLABORATION}
% --------------------------------------------------------------------------
\begin{figure}[t]
	\centering

	% Subfigure 1
	\begin{minipage}[b]{0.48\linewidth}
		\centering
		\includegraphics[width=\textwidth]{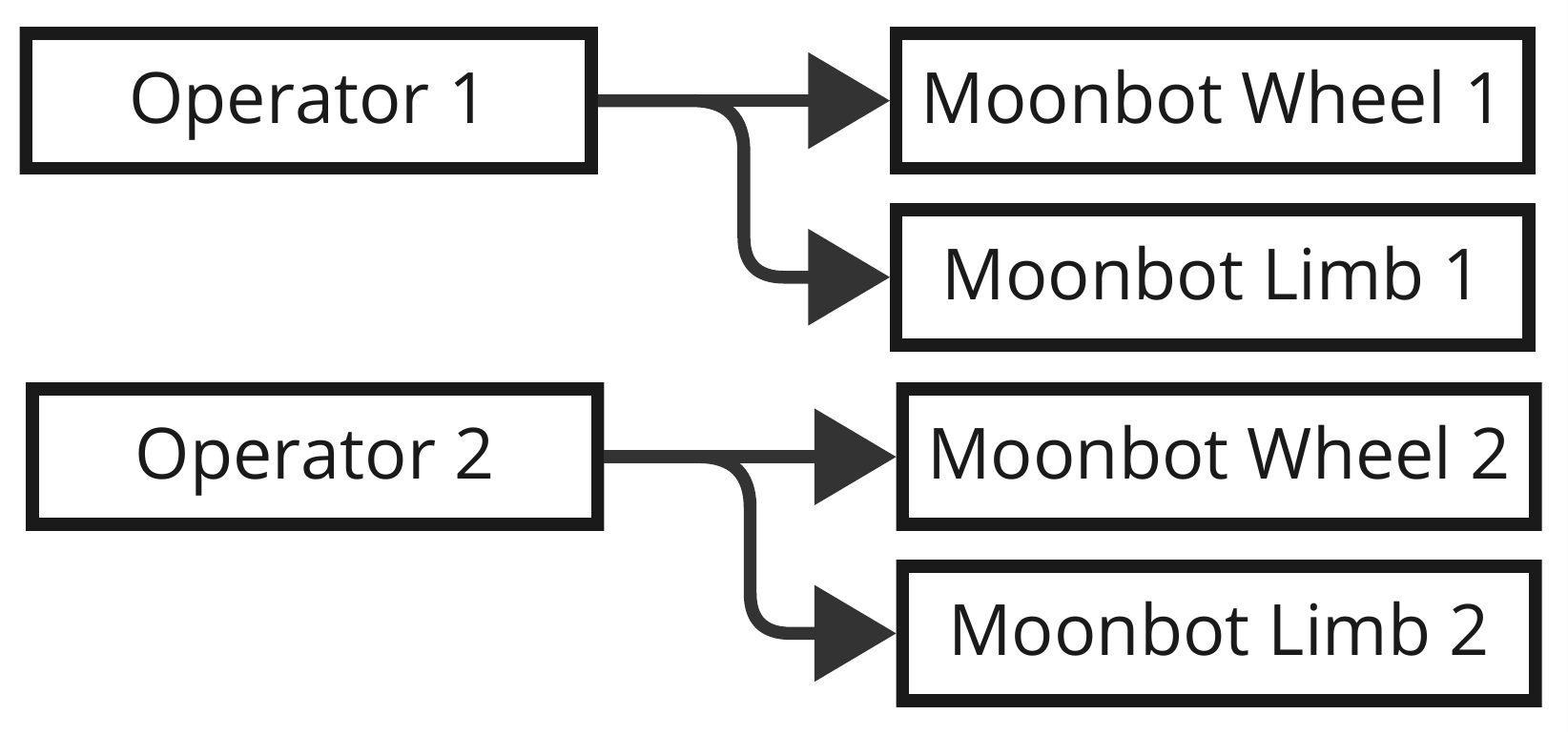}
		\subcaption{\ref{fig:software_config}.a}{Two \MBM}
	\end{minipage}
	\hfill
	% Subfigure 2
	\begin{minipage}[b]{0.48\linewidth}
		\centering
		\includegraphics[width=\textwidth]{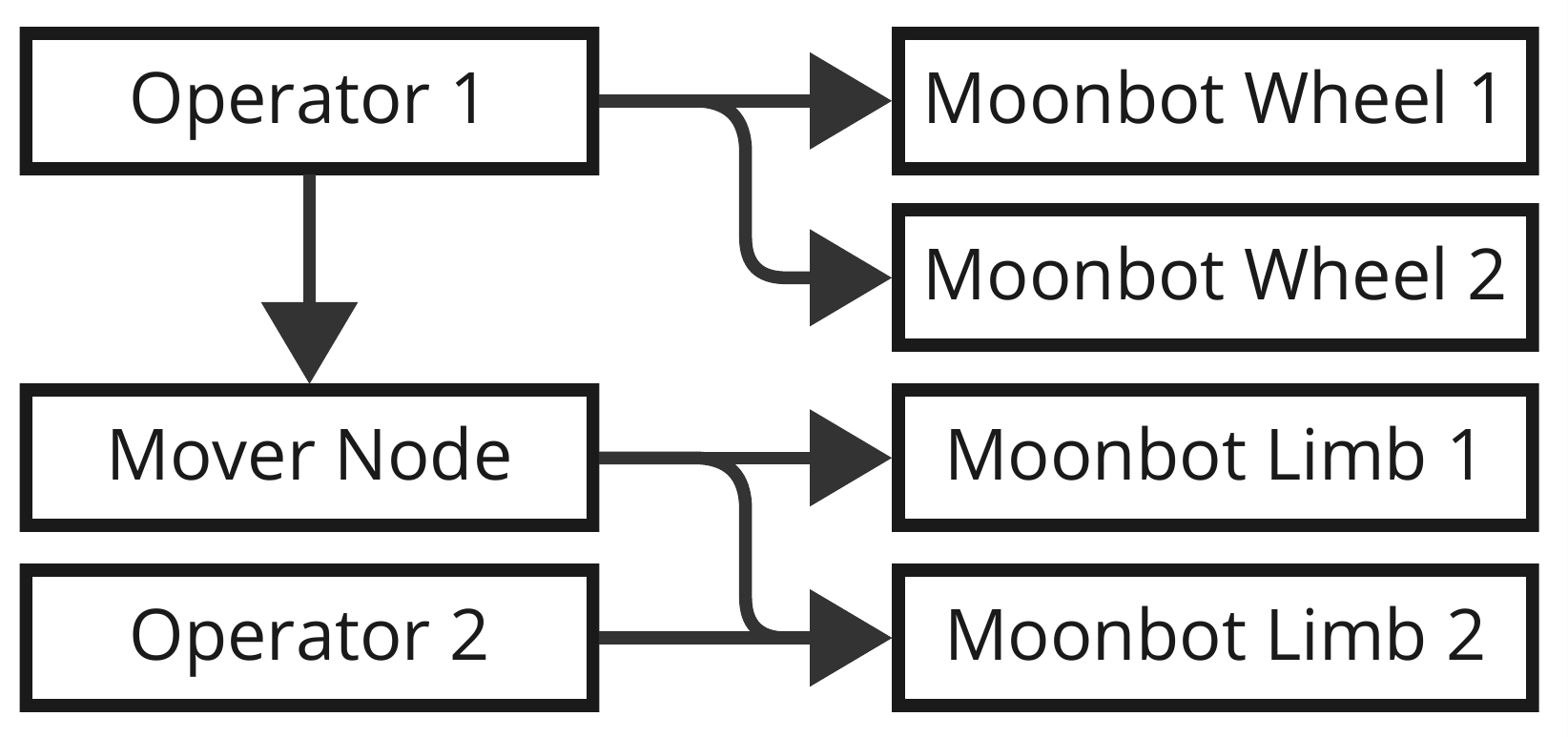}
		\subcaption{\ref{fig:software_config}.b}{\label{fig:archi_mbd} \MBD}
	\end{minipage}

	\vspace{1em} % Vertical spacing

	% Subfigure 3
	\begin{minipage}[b]{0.48\linewidth}
		\centering
		\includegraphics[width=\textwidth]{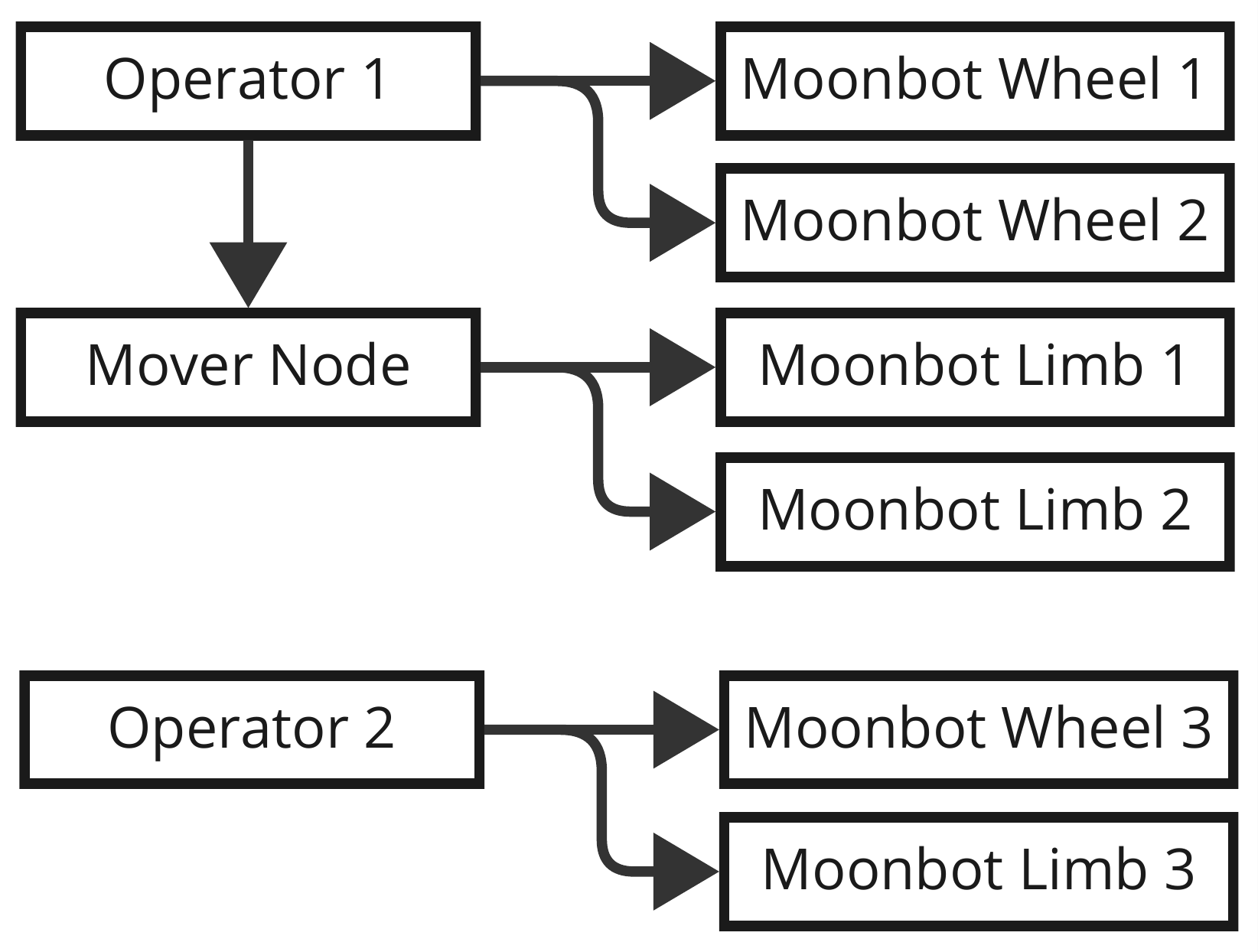}
		\subcaption{\ref{fig:software_config}.c}{\label{fig:archi_MBT_before}MoonBot Multicycle assembling}
	\end{minipage}
	\hfill
	% Subfigure 4
	\begin{minipage}[b]{0.48\linewidth}
		\centering
		\includegraphics[width=\textwidth]{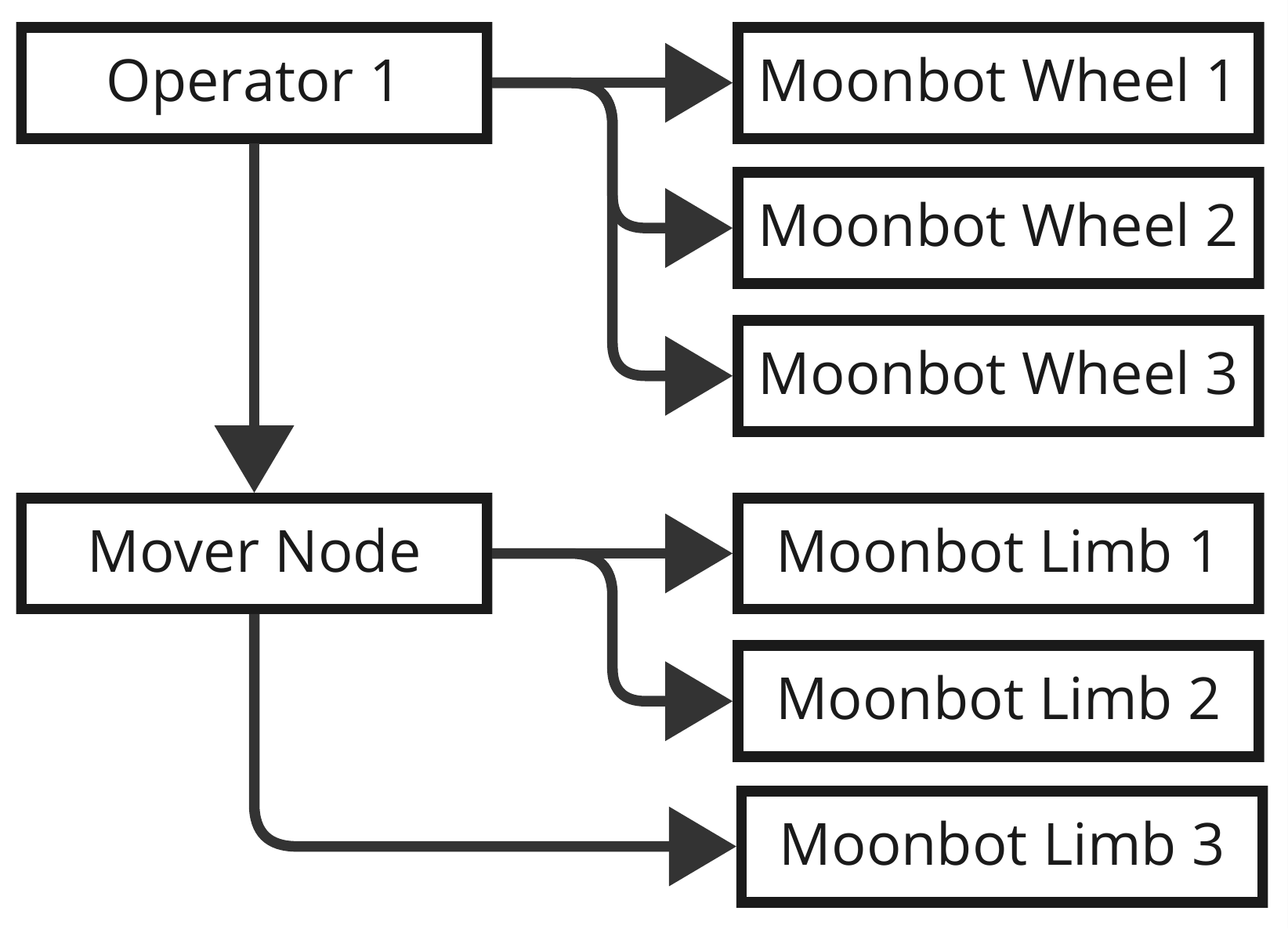}
		\subcaption{\ref{fig:software_config}.d}{\label{fig:archi_MBT_after}MoonBot Multicycle assembled}
	\end{minipage}
	\caption{\label{fig:software_config} Different software architecture used on various \MB configurations.}
\end{figure}
% --------------------------------------------------------------------------
\fig{fig:software_config} illustrates the various robot and operator architectures used during the field tests. The system’s deep modularity enabled simultaneous operation of multiple robots, each controlled by several operators, with the flexibility to control any node on the network.

This modularity was crucial not only for conducting multiple experiments in parallel, but also for enabling collaborative tasks, such as two robots working together to build the \MBT. Additionally, it allowed two operators to control a single robot. For example, on the \MBD, one operator controlled the wheels and steering, while another operated the manipulator.

\subsubsection{RESILIENT REMOTE JOINT CONTROLLER}

The operator node allows precise movement of a selected joint at a speed set by the operator---the input method being the press of a button. It employs an open-loop controller running on the operator's computer. This ``remote controller'' sends commands to a closed-loop ``local controller'' hosted onboard the robot. Although this system appears straightforward, the safety, intuitiveness and reliability requirements outlined in \refsafety{} introduce unforeseen complexity.

The primary requirements for the remote joint controller are as follows:
\begin{enumerate}
	\item The robot must move only if both the connection is established and the command is executed.
	\item The movement distance must be proportional to the integral of the speed command when the connection is established.
	\item Node pinging should be avoided.
\end{enumerate}

\begin{figure}[htb!]
	\centering
	\centerline{\includegraphics[width=\linewidth]{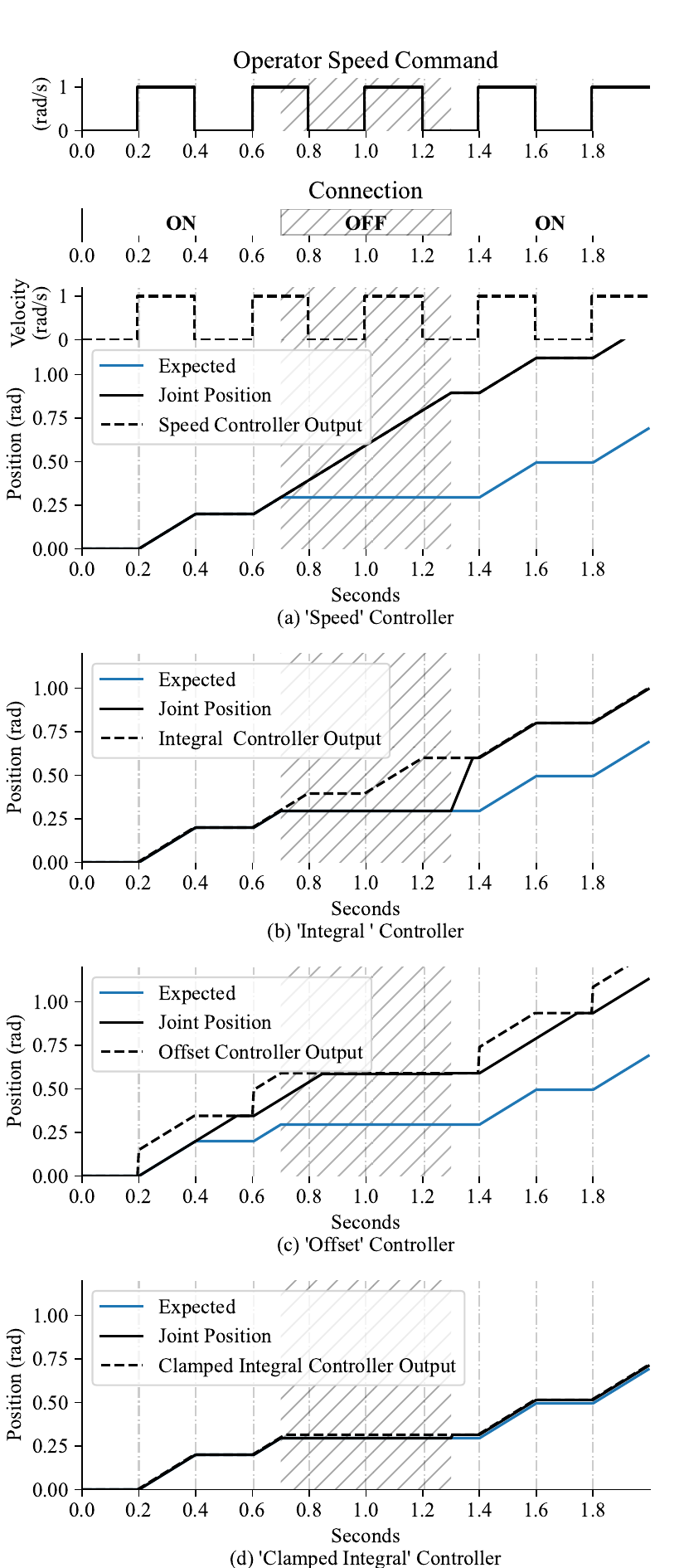}}
	\caption{\label{fig:joint_control} Simulated angular positions of several remote controller strategies, given the operator's button state and the connection status between the operator node and joint node.}
\end{figure}

\fig{fig:joint_control} illustrates the limitations of various remote control strategies under comparable operator inputs and connection statuses. Equations \eqref{eq:con1} through \eqref{eq:con4} describe the remote `speed,' `integral,' `offset,' and `clamped integral' strategies, respectively. In these equations, $u_k$ represents the remote controller's position command, $\dot{r}_k$ the velocity target, $y_k$ the joint's sensor reading, and $t_k$ the time at step $k$.

\begin{itemize}
	\item \textbf{Speed Strategy:} Sends velocity commands directly to the robot. This fails Requirement 1, as losing the connection does not stop the joint, as shown in \fig{fig:joint_control}(a).
	      \begin{equation} \label{eq:con1}
		      \dot{u}_k =  \dot{r}_k
	      \end{equation}

	\item \textbf{Integral Strategy:} Sends position commands derived by integrating the velocity target over time on the remote controller. This fails both Requirements 1 and 2. Integration continues even when the connection is lost, leading to excessive or unintended movement. This issue is evident in \fig{fig:joint_control}(b), where a noticeable jump occurs at 1.3\;s as the motor attempts to catch up with the accumulated command.
	      \begin{equation} \label{eq:con2}
		      u_k = u_{k-1} + \dot{r}_k \cdot (t_k - t_{k-1})
	      \end{equation}

	\item \textbf{Offset Strategy:} Sends position commands computed as the latest sensor value plus an offset $\delta_{{offset}}$ in the velocity direction. This fails Requirement 2, as pressing the button briefly results in a movement equal to the fixed offset, regardless of duration, as shown in \fig{fig:joint_control}(c). Additionally, fine velocity control is unachievable because the joint accelerates to its maximum speed to reach the distant target position.
	      \begin{equation} \label{eq:con3}
		      u_k = y_k + {sign}(\dot{r}_k) \cdot \delta_{{offset}}
	      \end{equation}

	\item \textbf{Clamped Integral Strategy:} Similar to the Integral Strategy, but limits the difference between the integrated position and the latest sensor reading to a threshold $\delta_e$. As shown in \fig{fig:joint_control}(d), this approach tracks the target accurately and remains resilient to connection interruptions. The resulting jump from any issue cannot be greater than $\delta_e$.
	      \begin{gather} \label{eq:con4}
		      u_k =                                                              \\
		      \max( y_k - \delta_e,\ \min (y_k + \delta_e,\
			      u_{k-1} + \dot{r}_k \cdot (t_k - t_{k-1}))) \nonumber
	      \end{gather}
\end{itemize}
% --------------------------------------------------------------------------
% --------------------------------------------------------------------------
During field tests, the ``Speed'' strategy was deemed unsuitable due to significant safety concerns and was not implemented. Both the ``Integral'' and ``Offset'' strategies were tested but posed challenges for operators, leading to slower task execution and inconsistent precision. These limitations motivated the development of the ``Clamped Integral'' strategy, which achieved great success.

\begin{figure}[t]
    \centering
    \centerline{\includegraphics[width=\linewidth]{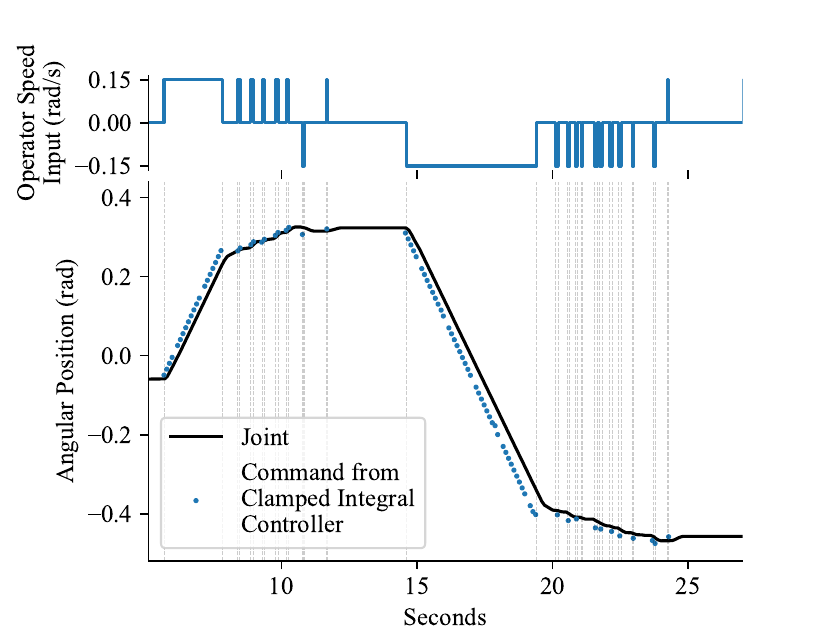}}
    \caption{\label{fig:joint_operator} Recording of the operator's speed input during a precise joint-placement task (top). Corresponding Clamped Integral controller commands and the robot joint's response (bottom).}
\end{figure}

\fig{fig:joint_operator} showcases a recording of an operator performing a precise joint control task using the Clamped Integral controller. The task was conducted with a keyboard as the input device. Following a large movement at maximum speed, the operator employed a series of brief key presses to fine-tune the joint's final position. This natural operator behavior underscores the importance of Requirement 2 and highlights why the ``Offset'' controller fails to provide the necessary precision in such scenarios.

The Clamped Integral controller allowed the operator to execute safe and precise movements effortlessly. Non-dependency on node pinging and handshakes is a major advantage for modularity; reducing the brittleness and complexity of the architecture, while easily allowing for several remote controllers to control the same joint without complication. Moreover, this clamping strategy is versatile and can be extended to trajectories more advanced than constant-speed commands. 
% Whereas, the next section will extend it to higher dimensions.
% --------------------------------------------------------------------------
% --------------------------------------------------------------------------

%%%%%%%%%%%%%%%%%%%%%%%%%%%%%%%%%%%%%%%%%%%%%%%%%%%%%%%%%%%%%%%%%%%%%%%%%%%%%%%%%%%
%%% FIELD TESTING REPORT
%%%%%%%%%%%%%%%%%%%%%%%%%%%%%%%%%%%%%%%%%%%%%%%%%%%%%%%%%%%%%%%%%%%%%%%%%%%%%%%%%%%
\section{FIELD TESTING REPORT}
% % --------------------------------------------------------------------------
% \begin{figure*}[t]
%     \centerline{\includegraphics[width=\linewidth]{./fig/field_test_overview.pdf}}
% \caption{MoonBot first field testing in a large sand field. The field was partially illuminated by a xenon lamp. MoonBot was assembled on a launch-lock palette positioned at the edge of the field and subsequently deployed to perform various tasks. Most operations were executed via remote control by human operators at the control center. \label{field_test_overview}}
% \end{figure*}
% % --------------------------------------------------------------------------
%Field Reports are descriptions of novel systems that have been in operation over an extended duration. They describe interesting and new implementations of known methods and discuss their performance in the field, or present innovative field robots and analyze their performance. Generally, Field Reports emphasize experimentation and experimental results rather than a rigorous analysis of underlying principles. 
This paper is the initial report to validate the hardware performance and demonstrate the fundamental capability of the MoonBot system to simulate the moon base construction scenario, in which the essential tasks were detailed in the introduction. 
{The primary purpose of this first field campaign was to demonstrate the feasibility of MoonBot in a partially representative lunar environment and to extract lessons learned for future system improvements.}
In this regard, the field testing campaign was designed in the dedicated analogue lunar test field to follow every milestone task settled for in our project (Section~I-\ref{Mission_Scenario}).
{Given the exploratory and proof-of-concept nature of this study, statistical significance was not the primary focus. Nevertheless, future experiments will be designed with repeated trials under controlled conditions, enabling statistical evaluation of performance metrics and providing a more rigorous quantitative basis for system validation.}

The field test was conducted in Space Exploration Field, the analogue lunar sandy field in Advanced Facility for Space Exploration~\cite{jaxafield} in Japan Aerospace Exploration Agency (JAXA) throughout three weeks. 
Performed field demonstration is overviewed in \fig{moonBaseConcept}. 
{In the test, the robots were operated through human-in-the-loop teleoperation,
% primarily via manual control,
with continuous monitoring of both sensory data and robot conditions to ensure safe and reliable operation. The batteries were manually replaced prior to each demonstration.}

% Additional testing was conducted in another lunar analogue facility. 

\subsection{ON-PALETTE ROBOT ASSEMBLY}
The self-production of a robot with basic mobility and manipulation capabilities within the MoonBot modules is the essential starting point of the mission. To validate this concept, we conducted two on-palette self-assembly demonstrations: 1) Assisted construction of a 7-DOF limb module using modular limb units and 2) Assembly of the MoonBot Minimal robot, integrating a 7-DOF limb and wheel modules. These demonstrations {highlight the feasibility of automated robotic assembly as a foundation} for modular robot deployment in future lunar operations.

\subsubsection{LIMB MODULE ASSISTED ASSEMBLY}
% \paragraph*{a) Experimental Setup}
The experiment was designed to demonstrate the assembly process of a 7-DOF modular robotic limb. In this demo, we assumed the building is processed by another limb, which can be pre-installed on the launch-lock plate to facilitate the palette deployment from the landing vehicle. The mutual connection by the multiple modules was considered for the future scope. A combination of UR16e~\cite{UR16e} robotic manipulator by Universal Robot and an adaptive three-fingered gripper by Robotiq~\cite{Robotiq-3F} was employed as the assembler limb, which is of similar scale to the MoonBot limb. The setup was arranged on a palette with dimensions of 1.5\;m by 1.2\;m.
% and a total weight of 80\;kg. 
The UR16e was strategically positioned in the top-left corner of the palette, while the modular components were placed centrally and on the left side to ensure accessibility during the assembly process. The procedure involved integrating two types of modular connectors: a screw-type connector and a diaphragm-type connector.

The modular robotic limb consisted of three components: two 3-DOF modules and one 1-DOF module. Due to its higher strength and payload capability, the screw-type connector was used in the first 3-DOF module that is attached to the palette. The second 3-DOF module consisted of a screw-type connector on one end and a diaphragm-type connector on the other end to demonstrate the compatibility with the 1-DOF module. The assembly process was divided into distinct phases to systematically evaluate the performance and precision of each connector type.

% \paragraph*{b) Semi-Autonomous Assembly Process}
% --------------------------------------------------------------------------
\begin{figure*}[th]
\centerline{\includegraphics[width=\linewidth]{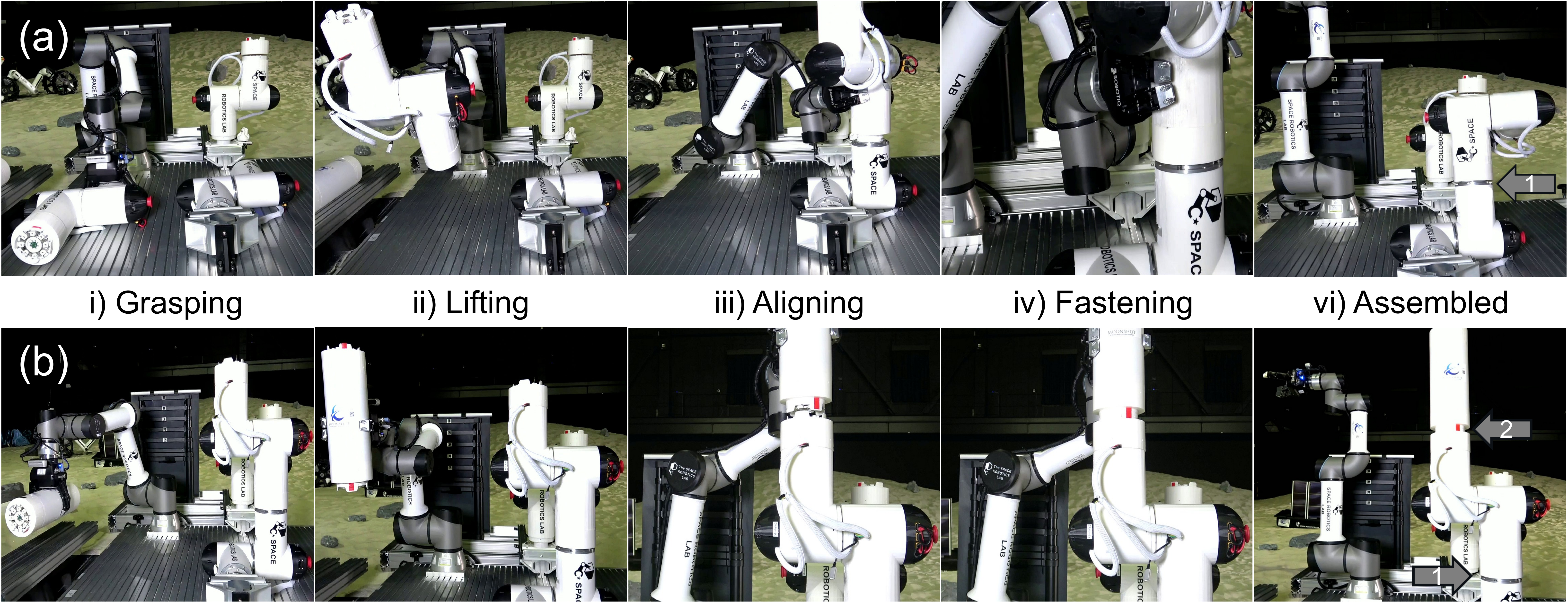}}
    \caption{Limb module assembly by another articulated robot module for both type of connectors screw connector and diaphragm connector. Grasping, lifting, aligning and attaching (a) the 3-DOF module with screw-type connector and (b) the 1-DOF module with diaphragm connector. \label{moonbotg_assembly_sequence}}
\end{figure*}
% --------------------------------------------------------------------------

% \\
% \paragraph*{\textbf{Semi-Autonomous Assembly Procedure}}
The following steps outline the assisted assembly process for the 7-DOF module (see \fig{moonbotg_assembly_sequence})

\begin{enumerate}
    \item \textbf{Initialization and Navigation:} {The UR16e robotic manipulator is programmed to operate in a semi-automated mode, following predefined trajectories with known environmental coordinates to navigate the workspace and identify assembly modules. A RealSense D435i camera provides a dual-angle view, assisting automated sequential operations and manual teleoperation. One 3-DOF module is mounted vertically on the palette with its screw-type connector facing upward for attachment, while another 3-DOF module is placed horizontally on the palette to be picked up by the UR16e manipulator.}
    
    \item \textbf{Assembly of the 3-DOF Module:} {The robotic arm executes automated motions to approach the 3-DOF module lying on the palette}, grasps and lifts it, with the screw-type connector facing downwards. Following a predefined downward trajectory, the robot applies consistent force to engage the lower module's rotating mechanism. The screw connector self-aligns as it engages, requiring only vertical orientation. Once securely connected, the robotic arm returns to pick up the 1-DOF module.

    \item \textbf{Assembly of the 1-DOF Module:} {The robotic arm executes automated motion to grasp and lift the 1-DOF module that is fitted with a diaphragm-type connector.} Unlike the screw-type connector, the diaphragm-type requires precise alignment of the locking claws. Using teleoperated IK-based control, the operator manually fine-tunes the alignment with assistance from the camera feed. Once properly aligned, the diaphragm connector engages automatically, eliminating the need for additional force. The 1-DOF module is securely mounted onto the previously connected 3-DOF module, completing the assembly of the 7-DOF limb, as shown in \fig{moonbotg_assembly_sequence} (the screw-type connector is marked as 1 and diaphragm connector marked as 2).
    
\end{enumerate}

The on-palette robotic assembly experiment validated the feasibility of utilizing a robotic arm for the on-demand construction of a modular 7-DOF robotic limb on the lunar surface, incorporating multiple connectivity systems. {The hybrid approach-combining automated pick-and-place with manual teleoperation} for precise docking ensures efficient and accurate module assembly. {Customization of the palette environment to facilitate visual feedback control, such as the addition of fiducial markers for autonomous calibration and visual servoing, could enhance the precision of autonomous operations.}
% --------------------------------------------------------------------------
\begin{figure*}[!thb]
    \centering
    \centerline{\includegraphics[width=\linewidth]{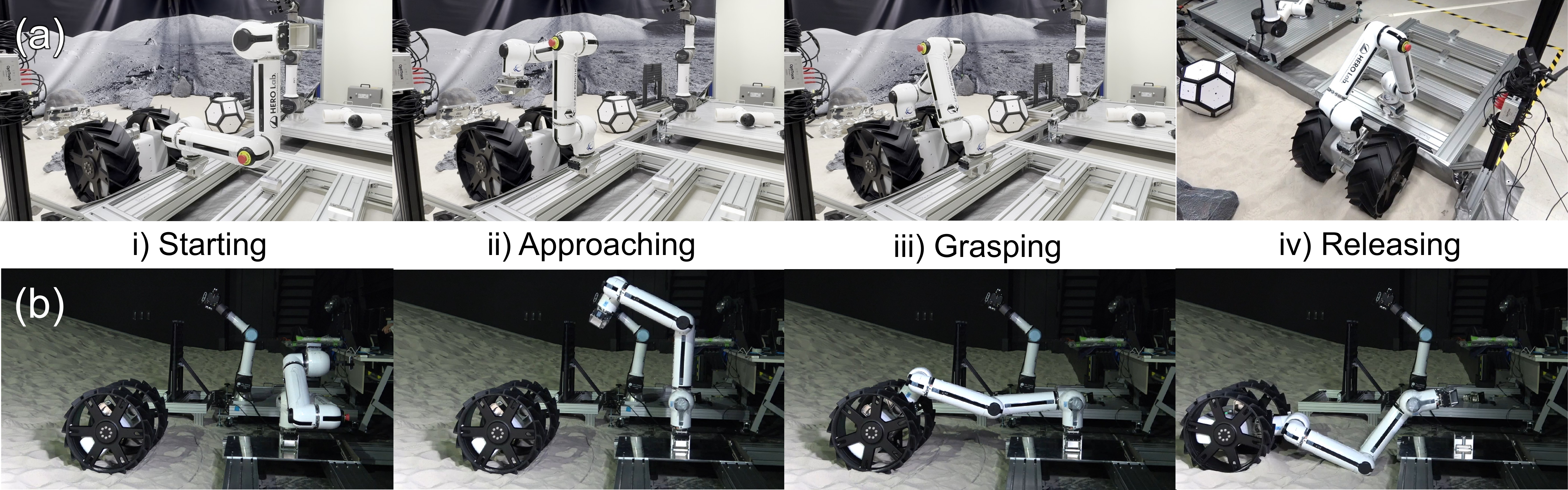}}
    \caption{Assembly process of the \MBM configuration. The Limb module is initialized in its zero position and then approaches the wheel's grapple fixture using IK. Releasing the opposing edge gripper from the palette prepares the assembled \MBM robot for field deployment. (a) In-lab testing was conducted to obtain ground truth for evaluation. (b) Corresponding demonstration was also conducted in the lunar analogue test field.}
    \label{fig:assembly_process}
\end{figure*}
% --------------------------------------------------------------------------
\subsubsection{MINIMAL MOBILE ROBOT ASSEMBLY}
The \MBM configuration was assembled by autonomously grasping and connecting the wheel module with the \MB limb module. Clamped Integral Controller was always used for teleoperation. 
% This process illustrates the \MB design's adaptability and modular assembly capacity in building functional robotic units for a range of tasks.
% \paragraph*{\textbf{Assembly Procedure}}
The following steps outline the assembly process for the \MBM configuration (see \fig{fig:assembly_process}):

\begin{enumerate}
    \item \textbf{Initialization:} The limb module was positioned at its calibrated zero position to ensure alignment and readiness for grasping.
    
    \item \textbf{Approaching and grasping operation:} Using \MS's Joint Node (level 1) and IK Node (level 2), the limb was teleoperated by an operator to approach the wheel's grapple fixture. The end-effector gripper securely grasped the wheel module. 
    
    \item \textbf{Releasing:} After grasping, the other gripper released its hold, and the assembled Minimal robot starts traveling.
\end{enumerate}
% --------------------------------------------------------------------------
\begin{figure}[thbp]
    \centering
    \centerline{\includegraphics[width=.9\linewidth]{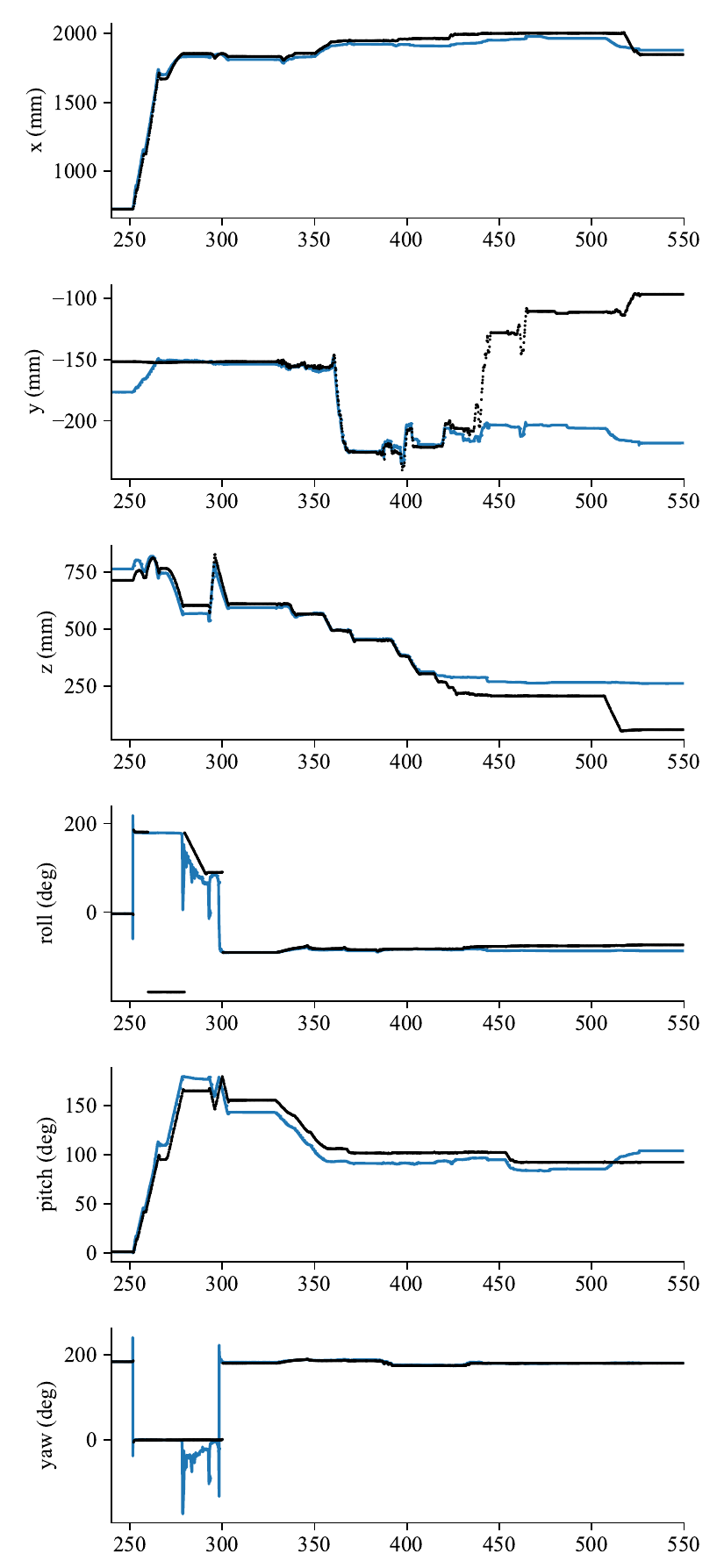}}
    \caption{Comparison of expected and measured tip poses during assembly: \mocap (blue) v.s. Forward Kinematics of \MS (black). Note the larger deviations in the $y$-direction during translational motion.}
    \label{fig:data_comparison}
\end{figure}
% --------------------------------------------------------------------------
% \paragraph*{\textbf{Ground Truth Data Collection}}
To confirm the limb's end-effector's pose during assembly, in the lab testing, the ground truth data was collected using the OptiTrack \mocap device, consisting of four Flex~3 and four Flex~13 cameras synchronized via an OptiHub 2 system~\cite{optitrack_website}. While the control system relied solely on the \MS's internal IK solver, this data was only used for comparison. Throughout the assembly process, the correctness of the tip posture was evaluated using the \mocap data as a reference.
As seen \fig{fig:data_comparison}, the comparison of the observed and anticipated poses showed great accuracy in the majority of orientations. A significant disparity, however, was noted along the $y$-direction, where deviations were greater than those in the $x$ and $z$ directions. Calibration offsets between the internal frame of the robot and the \mocap system are probably the cause of this disparity. The significance of careful system calibration is shown by the similar variances seen in previous research that used OptiTrack for robotic posture estimation~\cite{optitrack_stability}.

% \paragraph*{\textbf{Discussion: Insights for Modular Assembly}}
The outcomes of the \MBM configuration assembly show that autonomous operations are feasible for modular robot assembly, and ground truth validation is essential for measuring process correctness. To guarantee accurate ground truth comparisons, the \mocap system's calibration has to be improved, as seen by the observed $y$-direction disparity. This testing highlights the ability of modular robots to precisely validate using external ground truth systems while autonomously reconfiguring for certain tasks. Full autonomous operation is implemented in future work.

\subsection{IN-FIELD DEMONSTRATIONS}
To prove the concept of our mission scenario, MoonBots were deployed in the sand field designed to simulate analog lunar terrain. Tohoku Silica Sand, designated as ``Tohoku-Keisa'' No.~5~\cite{tohokukeisa}, is prepared to fill a 20\;m by 20\;m field to a depth of 0.3\;m. This sand is characterized by a homogeneous particle size distribution ranging from 0.1 to 0.6\;mm, a property that differs from that of lunar regolith. Nonetheless, we employed this test site in terms of safety and repeatability, focusing on evaluating the general capabilities of MoonBot rather than on a detailed qualification of its terramechanics performance in the actual fine lunar particles. The field was partially illuminated by a xenon lamp, whose light flux has a spectral distribution similar to natural sunlight but with lower irradiation intensity.
% This type of sand is rich in silicon dioxide (SiO2), exhibits excellent wear and acid resistance, and is a whitish natural silica sand. The sand is derived from a mixture of frog-eye clay and silica sand, from which quartz sand is separated by washing, drying, and sieving. This process ensures that the sand contains very few impurities and maintains a stable granular size. The grain-size distribution ranges from 0.106 to 0.6 mm. The particles are round and highly fluid, facilitating consistent conditions during testing. As a hard quartz sand, the particles have no cracks and exhibit low crushability, which is crucial for obtaining consistent results in the wheel-friction and wear tests.
\subsubsection{LOCOMOTION ON THE SAND}
% --------------------------------------------------------------------------
\begin{figure*}[thb]
    \centerline{\includegraphics[width=\linewidth]{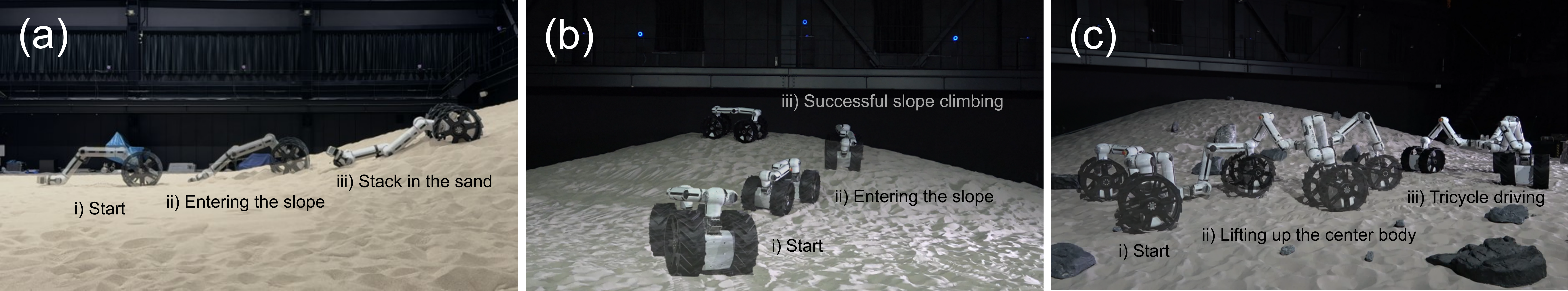}}
    \caption{Preliminary experiment on several configuration of MoonBot's traversability in the sand. Robot motion paths were visualized using fused sequential snapshots. (a) Minimal configuration utilizes its one limb as a tail for body stabilization. Ground pushing helped to increase the traction for slope climbing ($<$ 20$^\circ$), though deep sinking eventually led to stack of the wheel. (b) Vehicle configuration showed improved slope traversability, resulting in the successful climbing to the hill's summit. (c) Multicycle configuration has more wheels, which enhanced traction and thus showed significant potential for transporting heavier payloads. The enhanced performance in Vehicle and Multicycle configurations is attributed to the full weight of the middle limb(s) increasing net ground reactions and the distributed pressure across the additional wheels reducing the risk of immobilization.  \label{field_test_traversability}}
\end{figure*}
% --------------------------------------------------------------------------
\begin{figure*}[thb]
    \centerline{\includegraphics[width=\linewidth]{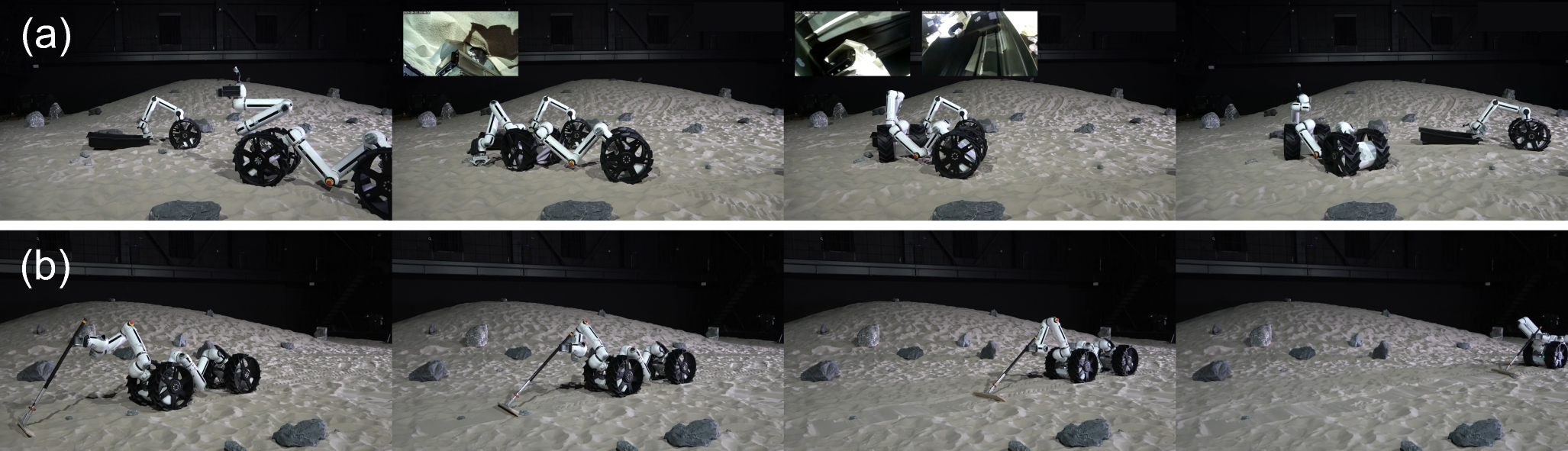}}
    \caption{Fundamental civil engineering demonstration by MoonBot teleoperated by human operator. (a) Rock collection to clear the surface by a collaboration of two robots. A Dragon was employed used its front limb module as a mobile manipulator to pick up rocks, which are then deposited into the sled and carried by a Minimal. (b) In a subsequent task, a Dragon employs a raking tool to level the terrain. A wireless camera mounted on the limb's wrist facilitates remote operation (real-time camera view are shown in selected snapshots). \label{field_test_civil_engineering}}
\end{figure*}
% --------------------------------------------------------------------------
\begin{figure*}[thb]
    \centerline{\includegraphics[width=\linewidth]{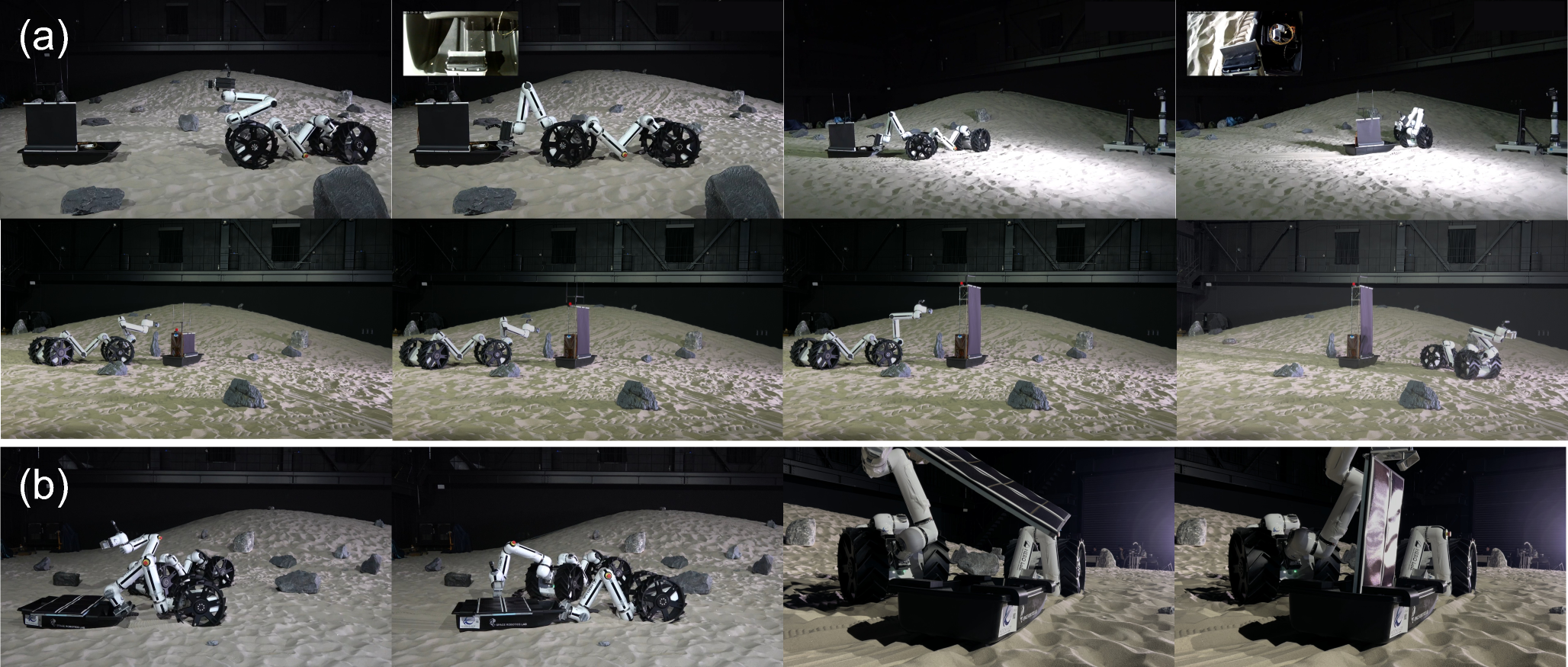}}
    \caption{Object transportation and manipulation demonstration by MoonBot for deployment of essential infrastructural components. (a) An extendible tower module ($>$30\;kg), representing a solar power generator and local communication station, was transported using a Dragon configuration. Upon reaching the designated illuminated area, the tower module was released and subsequently began its extension. (b) A panel-type solar power system was assembled by two robots. A lightweight panel module was carried by a Minimal, while a Dragon manipulated it into an upright position. These demonstrations were teleoperated by a human operator relying on a wireless camera mounted on the end-effector. \label{field_test_infra}}
\end{figure*}
% --------------------------------------------------------------------------
\begin{figure*}[thb]
    \centerline{\includegraphics[width=\linewidth]{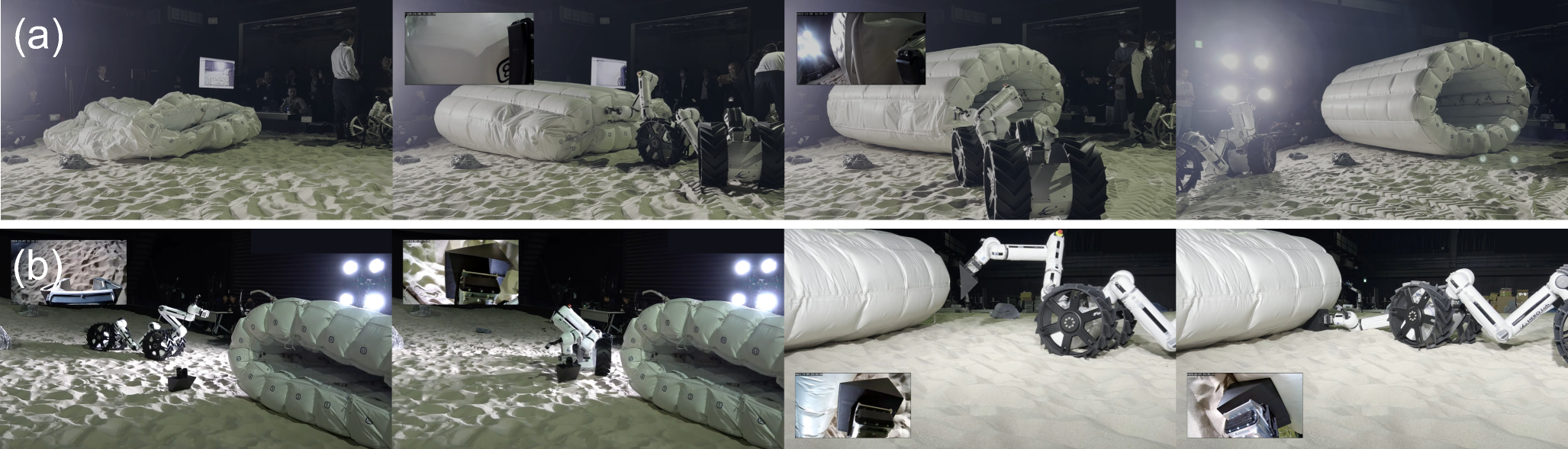}}
    \caption{Inflatable module deployment assistance demonstration by MoonBot. (a) MoonBot in Dragon configuration monitored the inflation of the module on a cell-by-cell basis using a hand-eye camera. By applying additional pressure to each cell with an external robot, the system verified the absence of gas leaks through internal pressure sensors integrated into each cell. (b) Once the inflatable module had been fully deployed and well-positioned, the robot placed stopper objects to further enhance its stability. \label{field_test_hidas}}
\end{figure*}
% --------------------------------------------------------------------------
Traversability on the granular surface is essential for the robot on the moon. MoonBot must adapt its locomotion mode to suit varying environmental conditions (see \fig{field_test_traversability}). To this end, a preliminary experimental evaluation of MoonBot was conducted to assess the traversability of its different locomotion modes in sandy terrain.

In Minimal configuration, the robot utilizes one of its limbs as a tail to stabilize its body; This limb can also be used for steering, while the wheel module independently controls the rotation speeds of its two wheels for skid steering. Through the test, we observed pressing against the ground enhances traction in case only wheel rotation was insufficient for traversing sandy terrain---a condition frequently encountered in deep sand areas or during slope climbing. However, on a 20$^\circ$ incline, the robot became immobilized due to excessive sinking at the midpoint of its trajectory. 

In Vehicle configuration, the robot demonstrated significantly improved traversability even on steeply inclined surfaces, successfully reaching the hill's summit. This enhancement is attributed to the middle limb, which bridges two wheel modules and presses them firmly against the ground to increase traction. Additionally, the increased number of wheels helped to mitigate the concentration of the pressure to the ground, preventing excessive digging into the sand. Notably, the control of Vehicle configuration is much simpler than that of Minimal. Versatile locomotion was confirmed just by rotating all four wheels at a constant speed, with the two wrist joints of the bridging limb actuated for steering. 

% {In Multicycle configuration, the robot utilizes an even greater number of wheels and limbs, effectively enhancing the net ground reaction force while distributing pressure across individual wheels and utilizing the limbs' degrees of freedom as active suspensions to negotiate irregular terrain. Consequently, in the current MoonBot design, this configuration yields the highest traction performance and minimizes the risk of wheel entrapment in sand, making it particularly effective for transporting a heavier object. The payload capacity scales with the number of connected Minimal modules. Advantages of parallel connection include the ability to transport objects without direct contact with the sand, while maintaining a larger stability polygon, which is critical for carrying large-scale payloads. It is also noteworthy that the use of a central Body module, which can accommodate additional limbs, enables advanced loco-manipulation tasks (see the right illustrations in \fig{MoonBot_concept}). Our currently developed MoonBot limbs can support a Body module (5\;kg) and a Limb module (21\;kg) under lunar gravity with three limbs (assuming that each limb owns six-times heavier payload capacity on the Moon).} 
% % It is meaningful to mention that this morphology represent a way to lift and carry a heavier payload by multiple Minimals. 

{
In the Multicycle configuration, the robot connects multiple limbs in parallel around a shared structure. This design is particularly attractive because the number of limbs can scale with the payload, enabling larger assemblies to support heavier objects. Unlike serial configurations (Vehicle, Dragon)---where a stuck wheel can immobilize the entire system---the Multicycle benefits from higher DOF and redundancy: a wheel can be lifted or reoriented to free itself from sand, or used as an active damper, reducing the risk of entrapment and rollover. The parallel layout also enlarges the stability polygon and keeps the mounted structure farther from the dusty ground---beneficial not only for protecting mission-critical payloads but also for maintaining line-of-sight in communications and improving onboard perception. Moreover, the shared structure can accommodate additional modules, payloads, actuators, and limbs, opening possibilities for advanced loco-manipulation missions as illustrated in~\fig{MoonBot_concept}.
Under Earth gravity, each Minimal module in the Multicycle provides at least 2\;kg of payload capacity---the maximum load that an arm can lift in its worst-case fully extended configuration. Under one-sixth lunar gravity, this scales to more than 12\;kg per Minimal, making it feasible for three Minimals (~\fig{moonbot_hardware_overview}) to support a structure composed of a body (5\;kg) and a 7-DOF limb (21\;kg).
}

\subsubsection{MOON BASE CONSTRUCTION TASKS}
In the field test, a variety of tasks were demonstrated by MoonBot. This preliminary ground testing was conducted not only to validate MoonBot's performance but also to serve as a proof of concept for a modular robotic construction mission on the moon. The lessons learned have been incorporated into subsequent iterations of the robot's hardware and software development, with a focus on enhancing autonomy and facilitating teleoperation. As presented in the introduction, the demonstrated tasks are categorized as follows: 1) Infrastructural hardware transportation and deployment, 2) Fundamental civil engineering operations, and 3) Robotic assistance in deploying pressurized modules.

\subsubsection*{a) Civil Preparation on the Sand}
The lunar polar region---our targeted area for base construction---features uneven terrain with scattered boulders and gravel. Consequently, after on-palette self-assembly of the robot, terrain leveling and obstacle removal become essential preparatory steps to ensure the stable and safe deployment of infrastructural components. To address these challenges, we conducted preliminary field demonstrations using MoonBot, focusing on fundamental civil engineering tasks. The robots were teleoperated by a human operator using wireless camera footage streamed from an on-robot camera mounted on the wrist of a limb, thereby enhancing remote operation. 

The experiments demonstrated both rock clearing and sandy terrain leveling (\fig{field_test_civil_engineering}).
For rock clearing, two robots collaborated to remove surface obstacles. In this phase, a Dragon configuration was employed for rock collection, during which the robot successfully picked up rocks and loaded them onto a sled. Subsequently, a Minimal robot was used to transport the sled by dragging it across the terrain. The experiments confirmed that the Dragon provided stable locomotion and manipulation, supported by its massive mobile body, while IK-based control of the front limb enabled the pick-and-place operations. In contrast, the Minimal exhibited sufficient traction performance for relatively lightweight loads and superior maneuverability through skid-steered pivot turning, a capability not available in the Dragon configuration.
After the removal of major obstacles, Dragon was operated to rake the sandy terrain. In this test, a simple raking tool was attached to Dragon’s front manipulator arm, and the robot was controlled to adjust both the tool orientation and applied force while maneuvering via wheel rotation. 
% --------------------------------------------------------------------------
\begin{figure*}[thb]
    \centerline{\includegraphics[width=\linewidth]{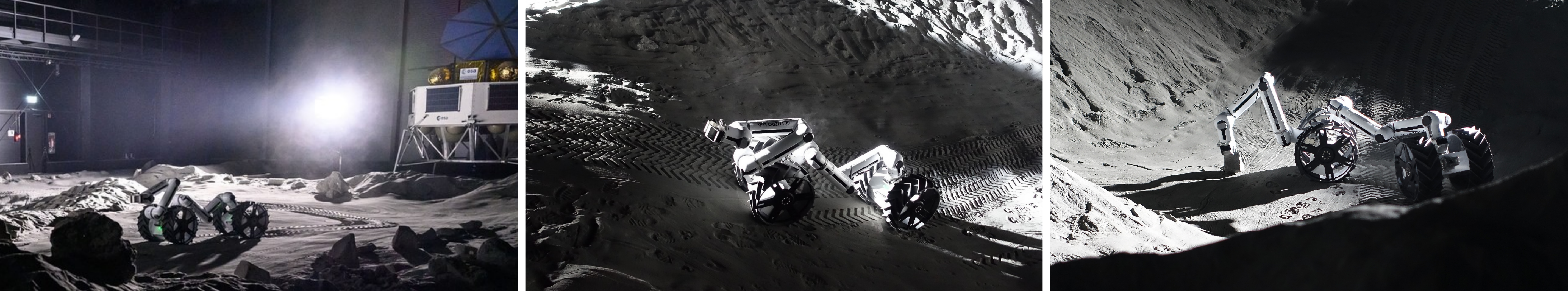}}
    \caption{{Additional field testing on a terrain fully covered with regolith simulant. The MoonBot was deployed in a lunar dust environment and teleoperated to traverse the 20\;m × 35\;m field (left). MoonBot in the Dragon configuration was able to climb a crater with a 20$^\circ$ inclination (middle). The Dragon configuration interacting with a medium-sized rock at the bottom of the crater (right). Unlike silica sand, fine regolith particles were easily lofted by wheel rotation, adhering to and contaminating the robot's surface.}
    \label{moonbot_dragon_luna}}
\end{figure*}
% --------------------------------------------------------------------------
\subsubsection*{b) Infrastructural Component Transportation and Deployment}
% \subsubsection{INFRASTRUCTURAL COMPONENT TRANSPORTATION AND DEPLOYMENT}
The ability of the robot to transport and manipulate components is crucial for deploying essential infrastructure. To demonstrate this capability, we designed two mission-oriented experiments focusing on the deployment of tower-type and panel-type structures (\fig{field_test_infra}).
At the lunar poles, sunlight is predominantly horizontal, necessitating the vertical arrangement of solar cells. To achieve this, an extendible tower-type structure was considered. Such mechanisms have been widely utilized in space missions, including the Extendible Optical Bench (EOB)---an extendible mast developed for orbital radio telescopes to achieve a longer focal length~\cite{EOB}. This type of structure has been employed in several in-orbit missions due to its high stiffness~\cite{extendibleMast1,extendibleMast2}. We adopted a similar extensible mast for this demonstration, which can also serve as a local communication station. 

For the experiment, a mock-up of the extendible tower structure was prepared. The mast was equipped with pole antennas at the top, and a rolled screen was attached along its side to deploy in synchrony with mast extension. The developed extendible tower module weighs over 30\;kg, which exceeds the payload capacity of MoonBot; consequently, a sled-based transportation method was selected. During the demonstration, a Dragon robot successfully dragged the sled carrying the mast over a distance of more than 10\;m to a designated illuminated area. Once positioned, the mast was released and began to extend. 

In addition to the tower-type structure, we also evaluated a panel-type structure. While panel-based assembly is highly adaptable for various configurations, this demonstration assumed constructing a small-scale solar power station. The test involved two robots: a Minimal robot transported the panel module, while a Dragon robot manipulated and positioned it in an upright orientation. These demonstrations were teleoperated by a human operator using a wireless camera mounted on the gripper for visual guidance.

\subsubsection*{c) Assistance of Inflatable Module Deployment}
% \subsubsection{ASSISTANCE OF INFLATABLE MODULE DEPLOYMENT}
In this project, we consider the use of an inflatable module known as HIDAS (Homeostatic Inflatable Decentralized Autonomous Structures)~\cite{Kimura_HIDAS}. HIDAS consists of multiple decentralized, pressurizable cells arranged in a cylindrical configuration. By controlling the internal pressure of these cells, the module can traverse sandy surfaces by rotating its entire body, effectively functioning as part of a modular robotic system~\cite{HIDAS_mobility}. To ensure the successful deployment of HIDAS, it is crucial to monitor the inflation process with the assistance of robots. Notably, HIDAS achieves spatial extendability through interconnected modular units---a design principle similar to that of MoonBot. To facilitate connections between inflatable modules, the module rotates and adjusts its position on the terrain after inflation. To further enhance this adjustment, external robotic assistance is anticipated; expected tasks include placing guide objects and inserting stoppers to prevent unintended rotation caused by local terrain inclinations.

In the demonstration (\fig{field_test_hidas}), a half-scale HIDAS prototype (fully inflated diameter: 2\;m) was deployed in the field. A MoonBot in Dragon configuration was selected to assist with the deployment process. 
Initially, the robot inspected the HIDAS cells to confirm the absence of severe membrane damage, using a hand-mounted camera. During inflation, the robot monitored the process by applying pressure to each cell, each equipped with an independent internal pressure sensor; an observed increase in internal air pressure in response to the applied external force confirmed the absence of gas leaks.
Once both inflation and positioning were complete, MoonBot carried on an additional stabilization task by inserting a stopper. The stopper was designed with a handle compatible with MoonBot's gripper fixture, thereby enabling teleoperated pick-and-place execution. The robot successfully positioned the stopper underneath the side of HIDAS to prevent unintended movement.
{
\subsection{ADDITIONAL TEST IN LUNAR REGOLITH SIMULANT}
An additional field test was conducted in LUNA~\cite{casini2020lunar}, an analog test facility for future human and robotic missions to the Moon, jointly operated by the German Aerospace Center (DLR) and the European Space Agency (ESA), to evaluate the basic durability and traversability of the MoonBot in a more realistic lunar dust environment (see \fig{moonbot_dragon_luna}). The MoonBot in the Dragon configuration demonstrated significant capability to traverse a granular surface covered with lunar mare regolith simulant (EAC-1A~\cite{engelschion2020eac}) and to descend and ascend crater-like terrain with inclinations of approximately 10$^\circ$--20$^\circ$. The robot was teleoperated from the control room by human operators and exhibited no functional issues due to dust throughout the three-day testing period.}

\section{LESSONS LEARNED AND FUTURE DIRECTIONS}
The field experiment, conducted in a simulated lunar environment, successfully demonstrated MoonBot's general capability to execute a range of milestone tasks critical to lunar base construction. Furthermore, throughout this testing campaign, numerous valuable insights were gained, which are guiding further refinements and enhancements to the robot system.

\subsection{HARDWARE}
\subsubsection{DAMAGE ASSESSMENT FOR ROBUST DESIGN}
%Sand Intrusion
In total, the MoonBot field test lasted three weeks. Prolonged exposure of the MoonBot hardware to sand will cause abrasion, clogging of mechanisms, electrical shorts, and sensory interference. This impairs performance and causes hardware failure. 
During the test, MoonBot robotic modules were stowed in a clean maintenance room between testing runs. This minimized exposure to the sand field to only during the operation of the MoonBot. Further study is required on the long-term effects of exposure to the simulated lunar environment on the MoonBot hardware. Even with sand exposure limited to hours at a time, the rough environment posed a significant challenge to the reliable operation of the robot system. To extend the service life of the MoonBot, only the functional ends of the modular hardware should make contact with the sandy terrain. However, contact was inevitable due to swaying, imbalance, or human error during teleoperation. This caused surface-level damage to the outer surface of the MoonBot as the extremities were dragged over the sand and other surfaces. 
% In extreme cases, incorrect operation caused joints, hatches, and other openings to come into contact with sand. This caused the gradual intrusion of sand into the MoonBot hardware through imperfections and other small openings.

% A systematic triage approach was used 
During the test, when robot modules lost functionality, to identify, resolve, and prevent further hardware issues, a quick overall evaluation of the robot's condition was performed. This evaluation comprised a visual inspection for physical damage, auditory checks for abnormal noise, and a review of telemetry data. Issues were prioritized based on their impact on mission-critical functions in the following order: safety-related issues, mobility impairment, power issues, communication system faults, and sensor malfunctions. One notable case involved a partial failure of the wrist joint of a Limb module. Visual inspection revealed no external damage or unusual odor; however, the joint's motor operated quieter than normal. Although joint states were still present in the telemetry, the Limb failed to follow target trajectories. Consequently, the use of this Limb was suspended, it was replaced with another Limb module. Disassembly of the faulty Limb module revealed that a substantial volume of sand had entered through the hatch opening and propagated through the internal components during operation. Sand grains presented the joint links and infiltrated the harmonic drive, ultimately jamming the gear teeth. Coarse sand grains blocked the teeth from properly fitting, while finer grains statically clung to internal surfaces. 
% \subsubsection{IMPACT ON ROBOT FUNCTIONALITY}
% \todo{To be updated}

Furthermore, the sand used in the test field does not replicate some issues caused by the material properties and range of grain sizes of real lunar regolith. The lunar surface is electrostatically charged from exposure to solar radiation, causing wheels and other surfaces in contact with the regolith to tribocharge~\cite{wheel_charging}. The buildup of this static charge risks sudden electrostatic discharge (ESD) into sensitive electronics and instruments, at worst causing total part failure. Aside from ESD, charge buildup will cause dust to statically cling to the robot's surface, impacting the thermal dissipation of the robot and increasing the risk of particle ingress as more material and charge build up over time. This underscores the importance of proper sealing and proactive measures to prevent sand and dust ingress.

These experiences yielded several key lessons. First, fault tolerance through modularity is highly advantageous; if a module becomes nonfunctional, an identical replacement can restore system performance without requiring extensive repairs. Second, enhanced sand-proofing is essential for lunar deployment. The initial MoonBot prototype---designed primarily as a ground testing platform---featured multiple hatches and access points to facilitate maintenance. However, to prevent sand intrusion and improve resistance to dust contamination, the next version will minimize such openings, thereby enhancing overall durability in harsh environments. Moreover, several mitigation strategies can be employed: improved sealing of the openings and joints to prevent the ingress of foreign material, secondary external coverings like sleeves or gaskets for redundant protection, the selection of abrasion-resistant materials or the application of anti-static coatings, and active cleaning methods like wipers or brushing to maintain hardware condition.
% After the conclusion of the on-site experiment, the MoonBot hardware was thoroughly cleaned with brushes and air compressors with particular care paid to the end effectors, joints, and any recesses and pockets that would accumulate debris. Despite this, hardware disassembly done after returning the hardware to our laboratory showed that internal compartments still had a substantial buildup of dust and debris. 
% In-lab experiments in the hardware were temporarily suspended for two Limbs that lost joint functionality to deeply clean the motor housings and electric compartments.  

%Manuscripts describing robotic systems should clearly describe the principles underlying them in addition to the design and performance. Manuscripts should advance the current state of understanding and make clear why the advance matters. Authors are encouraged to report on what was learned in doing the work, rather than merely on what was done.

\subsubsection{CONNECTOR ANALYSIS FOR ADVANCED OPERATION}
% The testing highlighted notable differences in the characteristics and performance of the two connector types. The screw-type connector proved to be robust and versatile, offering the advantage of self-alignment during the screwing process. Its ability to tolerate minor misalignments, coupled with a mechanical locking mechanism requiring downward force, made it particularly suitable for autonomous assembly operations. However, repeated use of 3D-printed screw connectors exposed durability limitations, such as thread degradation and misalignment due to material wear over time. To address these shortcomings, future iterations may benefit from employing more durable materials, such as polycarbonate or metal alloys.

% In contrast, the diaphragm-type connector exhibited the advantage of automatic locking once aligned and activated, thereby eliminating the need for additional mechanical force during the connection process. That said, this connector required precise alignment of the locking claws, a task that posed challenges for fully autonomous operations. Nonetheless, teleoperation supported by a dual-camera system effectively resolved these alignment challenges, enabling successful docking. 
% While the diaphragm connector demonstrated reliable performance under controlled conditions, its reliance on precise alignment limits its applicability in unstructured or dynamic environments, making it less viable for fully autonomous assembly.
Connector design is a key for a modular robotic system. In this study, we have used three prototypes: {a parallel jaw gripper, primarily used in the MoonBot system; a screw-type connector; and a diaphragm-type connector, both applied to the modular limb. The parallel jaw gripper and screw-type connector demonstrated superior robustness and ease of use in semi-autonomous operations, while the diaphragm-type connector excelled due to its genderless design. To fully exploit modularity, genderless connectors are preferable~\cite{higen2014,toriiICM,diaphragm}; however, they generally exhibit lower connection stability compared to gendered grippers, making this an important engineering challenge. Another crucial point learned is the trade-off between allowing flexibility to compensate for alignment errors and maintaining high connection strength. Because connection robustness is critical for supporting a reconfigured robot body, our current conclusion is that
% which allows more flexible and scenarios requiring precise alignment. 
% These findings emphasized the importance of 
we will improve the design that facilitates visual feedback---such as integrating a 2D marker for visual servoing during autonomous docking---rather than relying solely on mechanical flexibility in the connector system.}

% To improve durability, we propose using more robust materials such as Nylon, Polycarbonate, or metal alloys. Surface coatings or lubrication can be applied to reduce friction and wear. 

Electrical connection capability is essential for power transmission between modules, enabling a module with lower battery levels to be charged by other connected modules. {It is important to note that next versions of MoonBot incorporate power connectivity between the gripper and the grapple fixture. This power-transmittable handle will also be mounted on the solar power station, enabling the robot to recharge itself simply by grasping it.}
% In the field test, human operators manually replaced batteries every before the demonstration, in the actual mission on the moon, this power-transmittable grapple fixture will be connected on the solar power station

Subsequent field tests should investigate troubleshooting scenarios anticipated on the lunar surface. Maintenance tasks among MoonBot modules will be a primary focus. For instance, replacing a Limb module that has experienced partial or total failure may involve one MoonBot serving as the ``patient'' and another as the ``doctor.'' In such cases, it is essential that the connectors in the failure module remain detachable even when locked. In this paper, only diaphragm-type connector possesses such capability, unlike the screw-type and the gripper-type connectors. 

\subsection{SOFTWARE}
% This field test was crucial in refining the software, demonstrating its modularity, robustness, and adaptability for deployment on research-grade robots.

\subsubsection{SAFETY AND RELIABILITY FOR TELEOPERATION}
Safety is paramount for a robot operating in close proximity to human operators. For autonomous space robots, safety is replaced by reliability.
% , as failures during a mission can be catastrophic. 
Considering this, we adhere to stringent safety requirements:
In the absence of commands, the robot always converges to a safe state, ensuring all joint velocities are zero, and no significant deviation from the current state. 
Commands and trajectories had to be computed ``just in time'', as low-level nodes were not allowed to execute pre-planned motion independently.
This greatly constrains the robot's performances.
However, given the robot's slow speed, thus semi-static nature, the absence of low-level trajectory planning was not a critical limitation.

One of the key challenges was detecting failures and reacting to ensure the safety requirements were met. Traditional health checks between nodes risked introducing more failure points than they resolved. Instead, the clamped controller provided a significant solution: by design, it halts movement when errors occur, inherently maintaining safety without requiring explicit failure detection. % This approach simplified the system dramatically, enabling seamless modularity where any node could safely command any sub-node, agnostic of the command source. This also facilitated multi-operator scenarios.
Ultimately, the ``just in time'', reliability-first system architecture proved to be remarkably effective. Over three weeks of testing, the software never led to failure. This was achieved despite daily updates, multiple simultaneous operators and robots, a variety of robot modules, and a wide range of configurations.

% \subsubsection{VALUE OF OPERATOR FEEDBACK}

Operator feedback played a critical role in identifying and resolving issues. Tested operators could get a ``feel'' of the robot, thus developing systems appropriate to its performances. 
Notably, it drove the development of the clamped controller, which addressed responsiveness and safety concerns. Without this early feedback, the robot would have been harder to control, hampering future development. {Human presence is critical for robot operation in real missions. In human-in-the-loop teleoperation, operators dynamically adjust the level of autonomy based on the mission condition. For example, they may operate almost manually during mission-critical tasks monitoring sensor feedback, or alternatively provide only high-level commands while relying on the robot's autonomous capabilities. Such ``supervised autonomy'' not only ensures practical and reliable operation but also provides a safety margin to handle unexpected situations during complex tasks.}

Vision-enhanced autonomous software is also demanded. Robustness under the extreme lunar illumination conditions---characterized by a high dynamic range due to the absence of an atmosphere---can be improved using machine learning approaches~\cite{camille_SII}.

Through the test, we also concluded that providing a third-person view via an external camera, in addition to the first-person view from a hand-eye camera, effectively enhances positioning accuracy. In future testing, we will focus on collaboration vision sharing among multiple robots.

\subsubsection{SOFTWARE PHILOSOPHY FOR MODULARITY}

Modularity was a central focus in the system’s design and implementation. However, it is crucial to recognize that each failure point is amplified by the number of modules, and interdependencies introduce the risk of cascading failures. A key strategy for ensuring high reliability was maintaining homogeneity. Increasing the number of nodes and processes inherently adds potential failure points, but an even greater risk arises from the proliferation of code paths. Every piece of software tailored to a specific mission or robot becomes an additional point of failure, requiring continuous monitoring and maintenance. Standardizing both hardware and software elements is therefore essential to improving system robustness and operational efficiency.

However, creating robot-agnostic systems is inherently challenging, particularly under tight time constraints. A clear example of this difficulty was observed in the launch system, where this principle was not consistently followed. The launch process varied across different MoonBot configurations, making it the most brittle part of the system. While these launch-related issues were identified and addressed before mission execution, a similar failure occurring mid-task could have led to significant disruptions. This highlights the critical importance of standardization and robustness in mission-critical systems.

Counterintuitively, while the hardware is modular, the software must prioritize homogeneity to ensure reliability and maintainability. Standardized software reduces complexity, minimizes failure points, and enhances system robustness, ultimately improving mission success rates.

%%%%%%%%%%%%%%%%%%%%%%%%%%%%%%%%%%%%
% \subsection{RECOMMENDATIONS}
% \todo{To be updated}
% New scenarios, tasks, special modules
% More field tests need to be made with the current hardware to verify the fitness of the MoonBot platform in different Lunar base construction tasks and to continue identifying critical issues to improve with subsequent MoonBot versions. The next tests with the current MoonBot hardware will focus on automating MoonBot operation at every control level. 

% Currently, operation of the Minimal, Vehicle, Dragon, and Multicycle modes is fully teleoperated, while semi-autonomous assembly of the Modular Limb was achieved using the UR16 manipulator. A

% As new scenarios and problems are identified, more modular modes, specialized tools, and connectors must be designed and verified to ensure compatibility between all different modules within the MoonBot ecosystem. 

% Weatherproofing the hardware
%%%%%%%%%%%%%%%%%%%

\section{CONCLUSION}
We have demonstrated how modular, reconfigurable robots can function as unmanned constructors for lunar base development. The proposed robot, MoonBot, was designed based on the concept of functionality-based modularity, an approach particularly well-suited for space missions with strict launch mass constraints.

{In the preliminary field tests, first MoonBot prototype successfully executed essential milestone tasks for lunar base construction. As next step, we plan to conduct larger-scale experiments with repeated trials and statistical evaluations to provide robust, quantitative validation of the principles of modularity, heterogeneity, and reconfigurability. These efforts will establish a stronger theoretical foundation and advance the development of the future versions of MoonBot.}

The project is continued by the rapid cycle of improvements and mission-oriented testing. In the short term, our primary focus is on enhancing autonomy in task execution. While the current demonstrations relied on human teleoperation, except for on-palette assembly, considering the roundtrip communication delay and the sheer distance to the Lunar surface, achieving {increased} automation of the MoonBot system is essential, with human intervention limited to high-level task assignment and goal setting for the robotic team. 
 
For this, vision-based autonomous navigation and manipulation will be essential to improve task accuracy and time efficiency. Additionally, as the number of deployed modules increases, higher-level autonomy {such as the task allocation planner} will be necessary for task allocation and progress management, ensuring optimal system-wide efficiency. {In the long term, further hardware developments are necessary, including qualification testing to ensure readiness for actual flight missions, as well as the design and implementation of an improved connector that optimally balances misalignment compensation with connection strength. These efforts are essential to enhance the overall reliability and performance of the system for future operational scenarios.}

In {overall} conclusion, we believe that modular and on-demand reconfigurable robots, as exemplified by MoonBot, will play a crucial role in future lunar development. The insights gained from these preliminary field tests will contribute to refining robotic systems for real-world lunar base construction missions. 

\section*{ACKNOWLEDGEMENT}
The authors would like to thank the Hyper-Environmental Robots Laboratory (HERO Lab.) in Hakusan Corporation, Tokyo, Japan, for their invaluable contributions to the development of MoonBot, and gratefully acknowledge Asteria Artefacts, ISP, and JAOPS for their support in the the development of the simulation software and assistance with robot operations during the field experiments. We also extend our gratitude to NIPPI Corporation for providing the opportunity to use the extendible mast in the field experiment. We would like to acknowledge Prof. Shinichi Kimura and the members of his laboratory at Tokyo University of Science for the valuable discussion and their operation of HIDAS in the collaborative experiments with MoonBot.

We are also grateful to the staff of the Advanced Facility for Space Exploration at JAXA Sagamihara and the staff of the DLR-ESA LUNA Analog Facility in Cologne for their kind support in utilizing their facilities for field testing. Finally, heartfelt appreciation to the members of SRL—Danish Ai, Matteo Brugnera, Elena Carulla Ruiz, Marcus Dyhr, Masazumi Imai, Takeaki Komine, Kamil Kozinski, Nette Levijoki, Jakob Madestam, Yudai Matsuura, Lukas Meier zu Biesen, Jakopo Mensch, Yoshimasa Muneishi, Tharit Sinsunthorn, Kazuki Takada and James Hurrell— for their continuous assistance in ensuring the success of the field test campaigns.

\bibliography{./IEEEabrv,bib.bib}

\vfill

\begin{IEEEbiography}[{\includegraphics[width=1in,height=1.25in,clip,keepaspectratio]{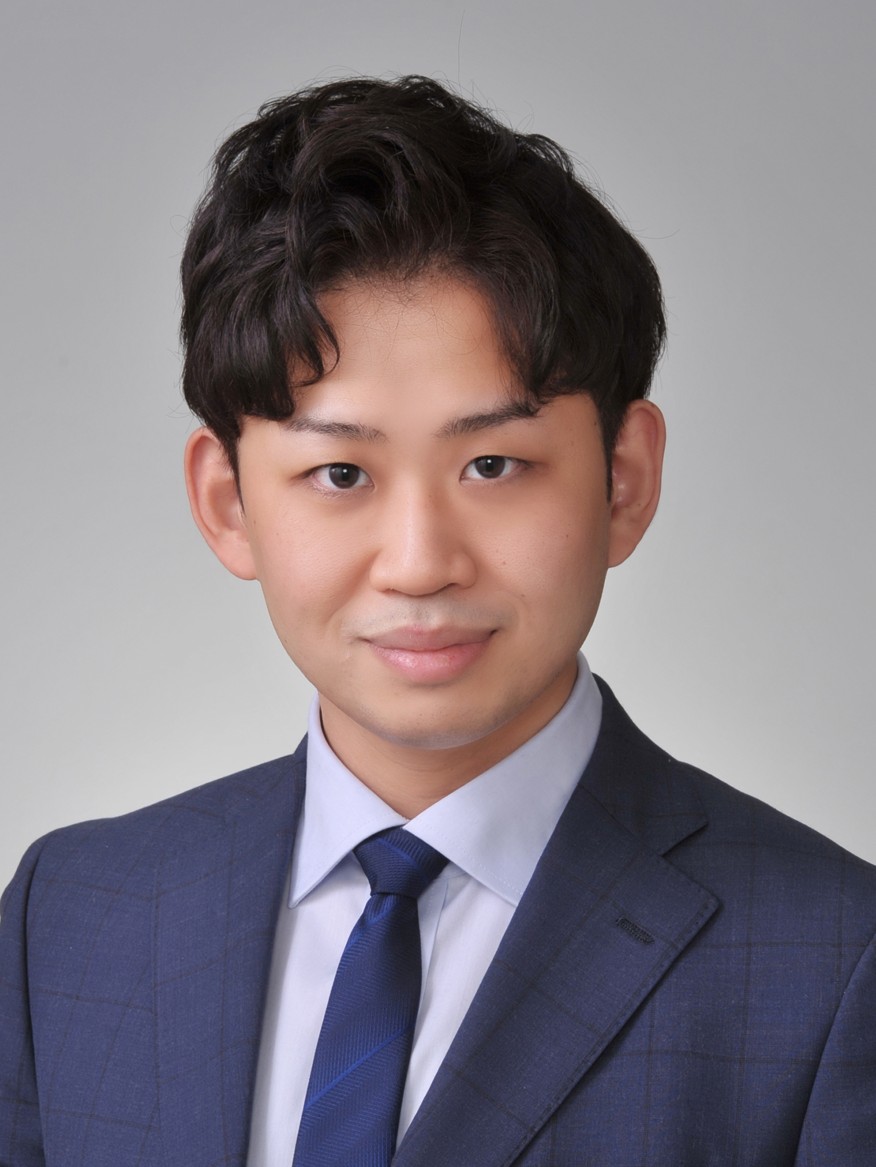}}]
{KENTARO UNO}~(Member, IEEE)~received B.S., M.S., and Ph.D. degrees in engineering at the Space Robotics Laboratory in 2016, 2018, and 2021, respectively, from Tohoku University, Japan. He studied as a Research Intern with ETH Zürich, Switzerland, in 2020. Since 2021, he has been working as an Assistant Professor with the Department of Aerospace Engineering, Tohoku University. His research interests include mobile service robotics in extreme environments, such as space and unstructured terrain. Representative keywords of his research expertise are planetary exploration robots, legged climbing robots, robotic mechanisms, and orbital servicing robotics.
\end{IEEEbiography}

\begin{IEEEbiography}[{\includegraphics[width=1in,height=1.25in,clip,keepaspectratio]{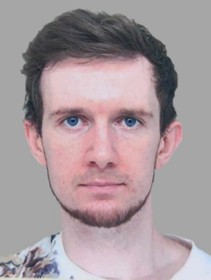}}]
{ELIAN NEPPEL}~(Member, IEEE)~received his B.S. degree in engineering in 2019 at École Centrale de Lyon (ECL), Lyon, France. He then followed a double-degree program, with 2 more years at ECL, and 2 years at Tohoku University, Japan. This earned him in 2023 a Diplôme d'Ingénieur and a first M.S. degree in engineering from ECL, along with a second M.S. degree in aerospace engineering from Tohoku University. He is currently pursuing a Ph.D. at Tohoku University’s Space Robotics Laboratory, where he focuses on motion planning, network and distributed control systems for modular robots as part of the Moonshot project. His previous work includes research on gait generation, path planning, and control systems for the SCAR-E (Space Capable Asteroid Robotic–Explorer) hexapod robot prototype.
\end{IEEEbiography}

\begin{IEEEbiography}[{\includegraphics[width=1in,height=1.25in,clip,keepaspectratio]{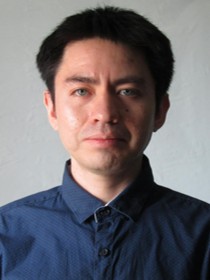}}]
{GUSTAVO H. DIAZ}~received his B.S. degree in electrical engineering from the University of Chile, Santiago, Chile, in 2018 and the M.S. degree in aerospace engineering from Tohoku University, Sendai, Japan, in 2024. During his M.S. degree, he worked on mobile robots testing localization and mapping algorithms and on autonomous assembly tasks with robot manipulators using reinforcement learning and vision systems. He is currently a Ph.D. student at the Space Robotics Lab at Tohoku University. His research includes hardware and software development of modular robots for space applications, such as self-assembly and construction tasks with a focus on learning frameworks.
\end{IEEEbiography}

\begin{IEEEbiography}[{\includegraphics[width=1in,height=1.25in,clip,keepaspectratio]{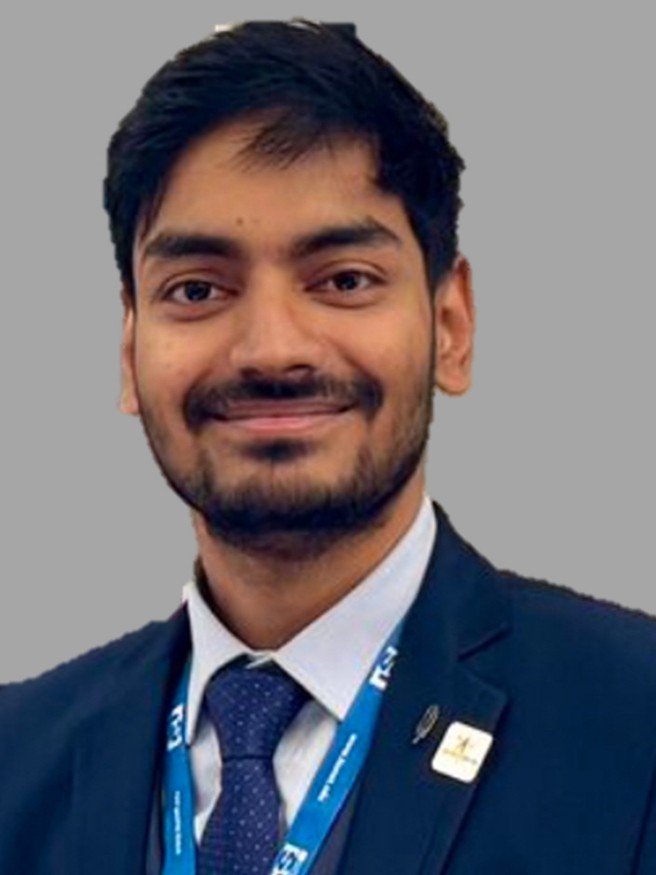}}]
{ASHUTOSH MISHRA}~(Member, IEEE)~earned his B.S degree in electronics from the University of Delhi, India, in 2021, followed by a M.S degree in electronics from Pune University, India, in 2023, where he graduated with distinction. Currently, he is a Ph.D at the Space Robotics Lab, Tohoku University, Japan, as a MEXT Scholar. His research is focused on developing autonomous, intelligent self-evolving robotic systems for lunar exploration and construction. Ashutosh's academic and research interests span space robotics, AI for autonomous systems, embedded systems, and neural networks. He has contributed to projects aimed at advancing lunar exploration technologies and is committed to pushing the boundaries of robotics for space applications.
\end{IEEEbiography}

\begin{IEEEbiography}[{\includegraphics[width=1in,height=1.25in,clip,keepaspectratio]{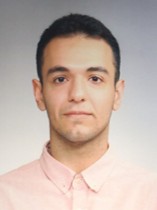}}]
{SHAMISTAN KARIMOV}~received the B.S. degree in mechatronics from Azerbaijan State Oil and Industry University, Baku, Azerbaijan, in 2022. He is currently pursuing the M.S. degree in aerospace engineering at Tohoku University, Sendai, Japan, where he is a student of the Space Robotics Lab. He is also part of the Moonshot project team, focusing on the development of modular robots for lunar exploration. His current research interests include self-assembly and reconfiguration of modular robots, with an emphasis on control systems, visual feedback control, and sensor fusion to enable autonomous adaptation to varying mission requirements on the Moon.
\end{IEEEbiography}

\begin{IEEEbiography}[{\includegraphics[width=1in,height=1.25in,clip,keepaspectratio]{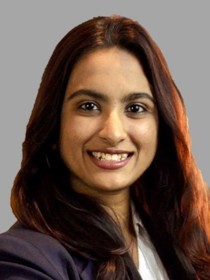}}]
{A. SEJAL JAIN}~(Associate Member, IEEE)~received the B.Tech degree in electronics and  communication engineering from PES University, Bangalore, India, in 2022, and M.S degree in aerospace engineering at the Space Robotics Laboratory, Tohoku University, Japan, in 2025. She is an alumna of the International Space University’s Space Studies Program (SSP22). Her research focuses on the development of wireless power transmission technologies for lunar
surface robots, specifically the design of receiver subsystems that enable efficient wireless power transfer to robots operating on the lunar surface. Additionally, she works in the hardware development of modular robots for self-assembly, with applications in lunar outpost construction tasks and avionics systems for orbital robotics.
\end{IEEEbiography}

\begin{IEEEbiography}
[{\includegraphics[width=1in,height=1.25in,clip, keepaspectratio]{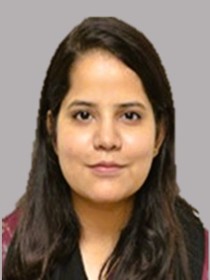}}]
{AYESHA HABIB}~received her B.S. degree in electrical engineering from GIKI University, Swabi, Pakistan, in 2021. After completing her undergraduate studies, she worked as an embedded firmware engineer at NRTC, a leading telecommunication company in Pakistan, where she gained hands-on experience in embedded systems. Currently, she is pursuing her M.S. degree in aerospace engineering with a specialization in Space Robotics at Tohoku University, Japan. Her research interests focus on the integration of robotic perception and learning-based techniques into autonomous systems. Specifically, she is working on robotic manipulators and hybrid modular robots designed to operate in unstructured environments.
\end{IEEEbiography}

\begin{IEEEbiography}
[{\includegraphics[width=1in,height=1.25in,clip, keepaspectratio]{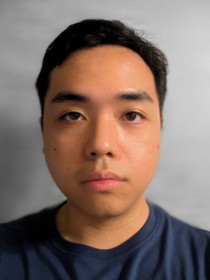}}]
{PASCAL PAMA}~received his B.S. in engineering from the Institute of Science Tokyo (2021) and M.S. in aerospace engineering from Tohoku University (2024). He is currently a Ph.D. student at Tohoku University, specializing in space robotics and advanced control systems. His research focuses on developing universal controllers for modular robots using transformer-based methods and vision-language-action models. Pascal’s work aims to create adaptive control systems that can operate across various hardware configurations without retraining or reprogramming, particularly for lunar base construction applications. He explores robot-aware control strategies to improve transferability and efficiency, aiming to advance modular robotics and autonomous systems for space exploration.
\end{IEEEbiography}

\begin{IEEEbiography}
[{\includegraphics[width=1in,height=1.25in,clip,keepaspectratio]{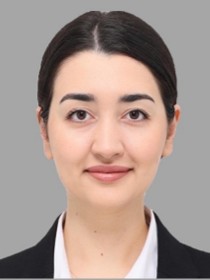}}]
{HAZAL GOZBASI}~received her B.S. degree in electrical and electronic engineering from Fırat University, Turkey, in 2018, completing exchange programs at the University of Maribor, Slovenia, and Shenyang University, China. Currently, she works as a Technical Assistant at Tohoku University, Japan, contributing to the Moonshot project, which focuses on developing robotic systems for Lunar Exploration and Human Outpost Construction. Her research interests include robotics for extreme environments, focusing on lunar exploration, robotic mechanisms, motion planning, and the simulation environments of modular robotic systems to optimize their performance. Earlier in her career, she gained experience in technical sales and project management across various industries.
\end{IEEEbiography}

\begin{IEEEbiography}[{\includegraphics[width=1in,height=1.25in,clip,keepaspectratio]{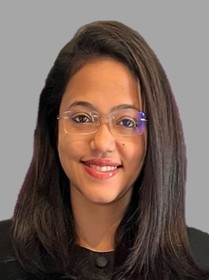}}]
{SHREYA SANTRA}~(Member, IEEE)~received the B.Tech degree in electrical and electronics engineering from National Institute of Technology, Jamshedpur, India, in 2014 and the M.S. degree in space and engineering systems from Skolkovo Institute of Science and Technology, Moscow, Russia in 2018. She received the Ph.D. degree in aerospace engineering at the Space Robotics Laboratory in Tohoku University, Sendai, Japan in 2021. During 2020-2021, she worked as a Visiting Researcher at DLR Institute of Communications and Navigation, Oberpfaffenhofen, Germany. Since 2021, she has been working as an Assistant Professor with the Department of Aerospace Engineering, Tohoku University. Her research interests include coordination, control, perception, and planning for multi-robot systems integrated with advanced algorithms to operate in unstructured challenging environments.
\end{IEEEbiography}

\begin{IEEEbiography}[{\includegraphics[width=1in,height=1.25in,clip,keepaspectratio]{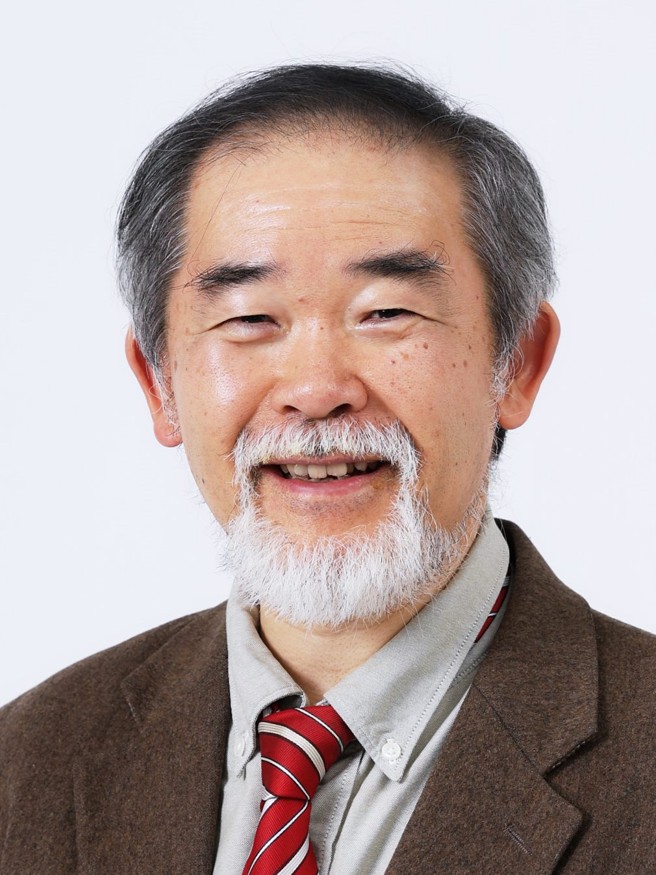}}]
{KAZUYA YOSHIDA}~(Life Member, IEEE)~is a Full Professor and Head of the Space Robotics Laboratory at Tohoku University. He received his B.E., M.S., and Dr.Eng. degrees in Mechanical Engineering from the Tokyo Institute of Technology. After serving as an Assistant Professor at the Tokyo Institute of Technology and as a Visiting Scientist at MIT, he was appointed Associate Professor in 1995 and Full Professor in 2003 in the Department of Aerospace Engineering at Tohoku University. His research focuses on the dynamics and control of space robotic systems, including free-flying robots and planetary rovers. Professor Yoshida has contributed to missions such as ETS-VII, the Hayabusa asteroid sample-return mission, the Google Lunar XPRIZE Hakuto rover, and the development of microsatellites for scientific applications. He is also active in international education as a non-resident faculty member of the International Space University located in Strasbourg, France.
\end{IEEEbiography}

\vfill\pagebreak

\end{document}